%% file: thesis.tex
\documentclass[10pt]{ClemsonThesis}

\usepackage{amsfonts}
\usepackage{amsmath}
\usepackage{pdfpages}
\title{Multiple View Neural Regression of a Facial Shape Model}
\department{School of Computing}
\documentType{Dissertation}
\major{Computer Science}
\degree{Doctor of Philosophy}
\graduationMonth{August}
\graduationYear{2025}
\author{Xiang Li}
\committeeChair{Dr. Eric Patterson}
\committeeMemberOne{Dr. Daljit Singh Dhillon}
\committeeMemberThree{Dr. Joseph T. Kider Jr.}
\committeeMemberTwo{Dr. Matias Volonte}

\hypersetup{
    colorlinks,
    linkcolor={black},
    citecolor={black},
    filecolor={black},
    urlcolor={black},
    pdftitle={\theTitle},
    pdfauthor={\theAuthor},
    pdfsubject={\theDocumentType},
    pdfkeywords={Clemson University, \theDepartment, \theDocumentType, \theMajor, \theDegree},
    pdfstartpage={1},
}

\begin{document}
    \frontmatter 

    \addtotoc{Title Page}{\maketitle}          
    \doublespacing                             
    \setcounter{page}{2}                       
    \addtotoc{Plain Language Abstract}{\input{plainabstract}}    
    \addtotoc{Abstract}{\input{abstract.tex}}  

    %
    %
    \addtotoc{Dedication}{\input{dedication.tex}}

    %
    %
    \addtotoc{Acknowledgments}{\input{acknowledgments.tex}}

    \singlespacing                             
    \tableofcontents \clearpage                

    %
    %
    \addtotoc{List of Figures}{\listoffigures} 

    %
    %

    \mainmatter 
    \doublespacing 

    %
    %
    
    \input{introduction.tex} \clearpage

    
    \input{backgroundMaterial.tex} \clearpage

    \input{relatedWork.tex} \clearpage

    \input{meshAcquisition.tex} \clearpage

    \input{acquisitionSystem.tex}
\clearpage

    \input{camera.tex} \clearpage

    \input{deepnet.tex} \clearpage

    \input{conclusions.tex} \clearpage

    %
    %

    \singlespacing                             

    %
    %
    %
    %
    %
    %
    %
    \bibliographystyle{plain}
    \addtotoc{Bibliography}{\bibliography{bibliography}}
\end{document}

%% file: plainabstract.tex
\chapter*{Plain Language Abstract}
This dissertation focuses on how to make realistic 3D face models more quickly and with less manual work. These models are used in things like animated films, video games, and virtual reality. Currently, creating high-quality face models usually takes a lot of time, effort, and expensive equipment.

To improve this, I built a custom device called VarIS. It takes pictures of a person’s face from different angles. These images help build detailed 3D models of the face. Although VarIS works well, it still takes time to process everything by hand.

To make the process more efficient, I also designed a pipeline that uses artificial intelligence to predict 3D face shapes directly from images. Instead of using photos of real people, I trained the system with computer-generated images that simulate real faces. I created these training images using software that I developed called Visage Craft.

This work also shows that using accurate camera settings helps the AI system to do a better job matching facial features. In the end, this project offers a faster way to make clean, usable 3D face models with very little manual work. It could be useful for anyone working on digital animation, gaming, or virtual characters.

%% file: abstract.tex
\chapter*{Abstract}
Creating re-topologized 3D facial meshes is a critical step in high-quality facial animation pipelines, yet it remains a labor-intensive and time-consuming task. Traditional approaches typically rely on multiview stereo reconstruction and specialized photometric environments to acquire accurate geometric and reflectance data under controlled conditions. This dissertation presents work toward more efficient capture of production-ready meshes including (1) developmental aspects of VarIS, a custom-designed light sphere capable of capturing high-resolution stereo geometry and reflectance maps—including diffuse, specular, and normal components under programmable illumination; (2) a study of the effects of camera parameters on automatic 2D and 3D landmarking methods, (3) methods for using synthetic data to train neural face regression, and (4) techniques proposed to improve neural multi-view regression of face shape. 

While VarIS enables photorealistic face capture, its operational cost and the need for manual processing of its acquired data highlight the need for a more scalable solution.  To address this, a deep learning–based framework is proposed, enabling direct prediction of re-topologized facial meshes from synthetic multiview images. Training data was generated using Visage Craft, an in-house rendering system built upon a physically based Appearance 3D Morphable Model (A3DMM). The method infers dense mesh geometry in a standardized format ready to rig and animate.  
Results demonstrate that integrating precise camera intrinsics and extrinsics during training markedly improves landmark accuracy and geometric consistency, and incorporating 3D landmarks themselves in the regularization of the network also improves results. The final system presents a robust, data-driven alternative to conventional face analysis/synthesis workflows, capable of producing facial meshes with minimal human supervision. 

%% file: dedication.tex
\chapter*{Dedication}

\begin{center}
  {\large In memory of my mother, Huizhen Wei}

  {\large \mbox{1965--2017}}

  {\large Your blessings carried me through my PhD journey \\ I love you}
\end{center}

%% file: acknowledgments.tex
\chapter*{Acknowledgments}
I would like to express my deepest gratitude to my advisor, Dr. Eric Patterson, for his unwavering support, insightful guidance, and thoughtful mentorship throughout my doctoral research. I am also sincerely thankful to my committee members, Dr. Daljit Singh Dhillon, Dr. Matias Volonte, and Dr. Joseph T. Kider Jr., for their valuable feedback and encouragement at every stage of this dissertation.

%% file: introduction.tex
\chapter{Introduction}
Facial analysis and synthesis have gained significance in computer graphics research for a variety of applications. Substantial progress has been achieved over the past few decades building on more complex models as compute power, computer vision, and computer graphics methods have improved. These advancements have found practical applications across various domains, notably in biometric systems utilized for mobile-device security and communication, but also in film production and game development among other areas. Facial analysis involves several core tasks: detecting human faces in images, accurately identifying landmark points around key facial features, and aligning different faces to a shared reference structure for further evaluation. In parallel, facial modeling or generation focuses on creating digital representations that faithfully capture the geometric complexity of the human face. These virtual avatars aim to preserve anatomical accuracy and fine surface detail while supporting visually convincing and expressive renderings.

High-quality facial analysis and synthesis in 2D and 3D environments require a clean background. However, real-world images often exhibit various backgrounds, facial expressions, and occlusions, posing challenges to completing these tasks. Researchers have dedicated considerable efforts over several decades to automating these processes. Nevertheless, in many cases, the involvement of human labor remains indispensable, particularly when high quality is desired. This chapter offers an overview of facial analysis and acquisition technologies and their challenges, explores the motivation behind harnessing state-of-the-art deep neural networks to enhance the quality and efficiency of 3D mesh generation, and outlines the structure of this dissertation.

\section{Overview of Facial Analysis and Acquisition}
Even with significant advances in digital facial modeling, building realistic, animatable human faces is still a complicated task that demands both technical precision and significant time. Traditional pipelines tend to follow a multi-step process: starting with face detection (like Viola-Jones), followed by landmark tracking using models such as Active Shape Model (ASM) and Active Appearance Model (AAM), then moving into 3D geometry reconstruction with techniques like multiview stereo. After that, the geometry usually needs to be retopologized for animation or production. Achieving lifelike results also requires well-controlled capture setups—carefully managed lighting, for example—to record fine surface details and reflectance accurately.

In recent years, deep learning has helped automate parts of this process. Convolutional neural networks, in particular, have outperformed older statistical models in many cases. However, these systems still rely on large, labeled datasets, which are usually collected under highly controlled conditions. That makes it harder to scale or apply them broadly. So, while neural models have reduced the need for manual work in some areas, creating clean, animation-ready facial meshes still calls for skilled human input, especially in the final stages like retopology. The field continues to wrestle with how to balance automation and artistic control. So far, neither traditional workflows nor deep learning solutions have fully eliminated the need for curated data and manual refinement when the goal is photorealistic synthesis.

\section{Motivation}
Recent advances in deep learning have significantly advanced digital facial synthesis, addressing long-standing challenges in creating realistic, animatable digital twins while mitigating the uncanny valley effect. Traditional workflows depended heavily on statistical methods and labor-intensive manual processes, including landmark detection, controlled lighting environments, multi-view stereo reconstruction, and manual retopology. However, convolutional neural networks (CNNs) have now surpassed these traditional techniques in accuracy and robustness, effectively managing diverse real-world conditions such as pose variations, occlusions, and illumination changes by leveraging comprehensive and varied datasets (e.g., 300W, COFW, WFLW).

Furthermore, innovative architectures like Autoencoders (AEs), U-Net,  and ResNet enable an end-to-end, data-driven approach that directly generates production-ready, retopologized 3D facial meshes from images. Unlike conventional methods that required sequential, multi-stage processing, deep neural networks efficiently learn detailed geometry, textures, and reflectance properties simultaneously within a unified framework. These advances also achieve real-time performance through GPU acceleration, significantly reducing processing times from hours to milliseconds per frame, thus enabling dynamic, high-fidelity avatar creation directly from video inputs.

Despite the transformative improvements in efficiency and realism, deep learning-based methods remain reliant on extensive datasets captured under controlled conditions, underscoring the continuing importance of foundational computer graphics techniques. The synthesis of traditional graphics principles with deep learning thus represents a balanced paradigm shift, ensuring both artistic quality and practical scalability across diverse applications in film, gaming, and real-time digital avatar creation.


The following chapters present the background, motivation, and technical contributions that underpin the development of facial shape modeling using both traditional statistical approaches and deep neural network frameworks. Chapter 2 begins with a review of foundational concepts in facial analysis, geometric modeling, animation, and photorealistic rendering — establishing a solid basis for the research that follows. Chapter 3 continues with a detailed survey of related work, covering advances in Appearance 3D Morphable Models (A3DMMs), Light Stage capture systems, and deep learning methodologies for 3D facial synthesis.

Chapters 4 and 5 introduce core technical contributions of this dissertation. Chapter 4 explores traditional 3D mesh acquisition workflows, focusing on the challenges inherent in geometry capture, mesh retopology, and the integration of these assets into production pipelines. Chapter 5 presents two key tools developed as part of this research: the Variable Illumination Sphere (VarIS), a hardware system for capturing high-fidelity facial data under controlled lighting, and Visage Craft, a software platform for generating synthetic face meshes using physically based rendering. Together, these tools enable the generation of diverse and photorealistic datasets for training data-driven facial synthesis systems.

Chapter 6 evaluates the impact of camera intrinsic and extrinsic parameters on facial landmark localization accuracy. Through controlled experiments, this chapter demonstrates the importance of precise camera modeling in enhancing 3D synthesis quality and ensuring alignment consistency across views—an essential factor in multi-view deep learning architectures.

Chapter 7 presents the main contribution of this dissertation: a multi-view neural regression framework for generating retopologized, animation-ready 3D face meshes. Two neural network architectures are developed and compared using synthetic training data, with an emphasis on achieving geometric accuracy, topological consistency, and practical applicability for production environments.

The dissertation concludes by summarizing key findings, reflecting on their significance within the fields of computer graphics and computer vision, and proposing directions for future research in realistic facial modeling, dataset generation, and neural avatar construction.

\section{Contributions}
In summary the contributions of the work described here include:  (1) developmental aspects of VarIS, a lower-cost, custom-designed light sphere capable of capturing high-resolution stereo geometry and reflectance maps—including diffuse, specular, and normal components under programmable illumination; (2) a study of the effects of camera parameters on automatic 2D and 3D landmarking methods, (3) methods for using synthetic data to train neural face regression, and (4) techniques proposed to improve neural multi-view regression of face shape that include incorporation of visibility point projection and regularization using landmarks.

%% file: backgroundMaterial.tex
\chapter{Background Material}
Creating a visually accurate 3D representation of human faces has remained a prominent and ongoing research focus within the field of computer graphics for several decades.\cite{debevec2000acquiring}. The achievements made in facial analysis and recreation have found widespread applications across diverse domains, encompassing but not limited to the realms of film production, digital and virtual reality games, medicine, forensic sciences, and anthropology. The process of obtaining realistic human face representations involves a series of intricate steps, including facial recognition, 3D data acquisition, geometry modeling, retopologizing, and texturing. Each of these subprocesses poses unique challenges due to the complex geometry and physical properties of human faces, coupled with inherent limitations in computational resources. Consequently, each step represents a significant research area with numerous unresolved obstacles. This chapter aims to provide a comprehensive overview of the process of acquiring photorealistic rerendered images of human faces. Furthermore, it will delve into the historical significance of milestone papers within each sub-area,  thereby tracing their pivotal role in shaping the landscape of computer graphics in facial analysis and generation.


\section{Facial Analysis}
In the domain of film production, the process of facial analysis holds great significance, particularly in the context of facial recreation. Prior to advancements in computer graphics techniques, the duplication of actors' faces for makeup effects in film production was labor-intensive, requiring hours of meticulous work to create silicone and plaster prosthetic molds.  
However, computer graphics techniques have revolutionized the facial synthesis process, enabling faster and more streamlined workflows. Instead of relying solely on traditional methods, images have become instrumental in generating 3D facial representations. This process commences with facial analysis, which involves detecting and locating facial features (e.g. landmarks) within images or video frames.  These key points play a pivotal role in numerous applications, ranging from facial recognition and expression classification to 3D facial synthesis and emotion processing. Consequently, identifying and extracting facial regions within complex images have become a crucial preliminary step in generating photorealistic human avatars. As a fast and accurate face detection technique, the Viola-Jones algorithm has been widely used in many  applications \cite{viola2001rapid}. In this paper, Viola and Jones detected human faces in images with Haar filters to extract face features and AdaBoost \cite{freund1997decision} to optimize classifiers. The algorithm used \emph{integral image} representation for rapid face feature evaluation and cascade-structure classifier for fast rejecting non-facial regions. Upon successfully locating faces within image frames, a critical undertaking in any 3D facial synthesis process is the identification of point-to-point correspondences on key points across multiple image pairs\cite{xiang2021, kim2017single}. These correspondences hold significant importance as they serve to abstract fundamental facial features such as the jawline, eyebrow contours, eye regions, nose shape, nostril positions, and mouth characteristics. By aligning and correlating the keypoints across different images, a more complete and detailed 3D facial model can be constructed, capturing the nuances and intricacies of the subject's facial geometry.

Establishing corresponding key points between stereo images involves assessing the similarity of pixel intensity values\cite{kim2017single}, and the acquisition of pixel-level features in images has long been an area of concentrated investigation in computer vision.  Local (window-based) algorithms \cite{Scharstein2001} are commonly used to match features between image pairs. Such algorithms conduct matching operations by comparing a pixel encompassed within a window from one image against a set of pixels' windows in the other image. The similarity between patches is frequently assessed through template matching algorithms, including methodologies such as cross-correlation or the sum of squared difference method (SSD) \cite{anandan1989computational}. Alternatively, in a more general manner, feature descriptors are used to abstract and measure distinctive features in one image, which are then compared to corresponding descriptors in other images. Various feature descriptors, including Scale-Invariant Feature Transform (SIFT) \cite{lowe2004distinctive}, Histogram of Oriented Gradients (HOG) \cite{dalal2005histograms}, and Harris corners\cite{harris1988combined}, provide robust representations of image features for matching and analysis.

\section{Face Modeling and Animation}
The generation of highly detailed face geometry combined with accurate reflectance information is a longstanding objective in the field of computer graphics\cite{debevec2000acquiring}. Traditional methods for acquiring human face geometry, such as the use of silicone and plaster prosthetic molds, suffer from significant drawbacks in terms of labor intensity and time consumption. Consequently, researchers in computer graphics have long focused on the challenge of representing human facial geometry and color information within a computational framework. This objective has also been a prominent research topic in the field of computer vision. Various approaches have been proposed to represent 3D objects in computer graphics, including Polygonal Meshes, Implicit Surfaces, Point Clouds, and Volumetric Representations. Each representation offers unique advantages: some excel at capturing intricate geometric details, while others are more suitable for rendering or animation purposes. This section will provide a concise overview of the historical development of face representation techniques in both 2D and 3D contexts and methods for facial animation.

\subsection{Representing a Face in 2D}
Throughout the decades of research, the Principle Component Analysis (PCA) has played a crucial role in facial modeling. PCA is a dimension reduction algorithm. Due to its linearity, PCA can present a large data set with variations on a set of orthonormal bases. As a complex geometry, human has a high degree of flexibility, which makes PCA a suitable tool to extract features of a face by eliminating less important information. In 1991, Turk and Pentland presented one of the first successful facial recognition methods using Principle Component Analysis in 2D, also called the Eigenface method\cite{turk1991eigenfaces, turk1991face}. They pre-trained a set of grey-scale images of human faces to get the eigenvectors, which defined the face space in a lower dimension. A new face image could be represented by projecting it onto the space spanned by the eigenvectors. This Eigenface method is still used in some systems for facial recognition and location. The Viola-Jones algorithm and Eigenface method provided promising results. However, both of them had one drawback: they trained the models with 2D images, which failed to recognize face features (e.g. location of eyes, face outlines, etc.) \cite{Egger2019}. Many researchers were spending efforts on training 2D shape face models with explicit representation for synthesis and recognition. As a modification of the Eigenface model, Jones and Poggio computed eigenvectors on pixelwise image correspondences on 2D images \cite{jones710791}. Jones and Poggio named the one-to-one correspondence-based model Morphable Model (MM) \cite{jones710791}. Unlike Jones' paper, Cootes et al. introduced an Active Shape Model (ASM), which used landmarks to manually annotate faces in images and train a statistical face model \cite{COOTES199538, cootes2000introduction}. The landmarks brought one-to-one correspondence across all training faces. Applying PCA to those landmarks allowed ASM to describe meaningful features of faces during feeding time. However, the ASM statistical deformable model only located face shapes in 2D images without texture information. In 1998, Cootes et al. proposed an Active Appearance Model (AAM) by integrating the texture information into the ASM while training \cite{cootes1998active}.  AAM combined the shape eigenvectors and color eigenvectors into one joint model. The combined parameters would be learned during the fitting stage. 

\subsection{Representing a Face in 3D}
In parallel with the advancements in 2D image processing techniques for faces, considerable research efforts have been directed toward the representation of faces in three-dimensional space. Arguably speaking, face geometry representation can be categorized into two distinct approaches. Firstly, geometrical models capture the surface position of facial structures in world space through the use of shape primitives such as points and polygons. While geometrical models excel at preserving intricate details of individual faces, they encounter challenges in effectively representing the general shape. Ever since Parke \cite{Park10.1145/800193.569955} proposed the first polygon-based human avatar, geometrical models have been adopted for production as an initial acquisition of human face data. Conversely, statistical models provide a retopologized representation of the entire distribution of human faces. Egger et al. \cite{egger20203d} offer valuable insights into the statistical acquisition of raw 3D face shape data, which can be broadly classified into two approaches: Geometry Methods and Photometric Methods. Geometry methods involve extracting 3D points by establishing correspondences between surface points in paired images. Various techniques, such as comparing pixel intensity or brightness across images and analyzing projected light patterns, are employed to achieve this correspondence. In contrast, photometric methods focus on analyzing the reflectance properties of objects in captured images, enabling the estimation of surface normals by examining variations in lighting conditions.

\subsubsection{3D Face Representation}
Obtaining three-dimensional vertices is arguably the most important stage in 3D facial synthesis \cite{egger20203d}. Parke at the University of Utah first proposed computer graphics techniques for representing, constructing, rendering, and animating human faces in 3D space \cite{Park10.1145/800193.569955}. Parke represented faces by several connected polygons. Those polygons reduced geometry complexities, which allowed rendering faces with Gouraud's smooth shading algorithms \cite{gouraud1971continuous}, reducing rendering time and interpolating facial expressions between frames. Parke's breakthrough research on 3D facial synthesis and animation brought various possibilities into modeling human faces in 3D, especially in 3D statistical models. One of the still commonly used statistical models, 3D Morphable Model (3DMM), was introduced by Blanz and Vetter in 1999 \cite{blanz1999morphable}. As an extension of Jones' paper \cite{jones710791}, the original 3DMM trained the model on 200 3D meshes with texture information. Many types of research have been tried to fit 3DMM for image-based synthesis \cite{cootes1998active, cootes2000introduction,blanz1999morphable, Pascal5279762}. Those synthesis approaches attempted to apply parameters from 3DMM to describe faces in images by minimizing the energies between models and images. This analysis-by-synthesis approach has shaped research trends for the past 20 years \cite{egger20203d}. Note that the concept of 3D morphable models was first introduced by Blanz and Vetter and was a significant milestone in computer graphics. It is a known challenge in the computer graphics and computer vision research community that the ability to acquire a large facial dataset for 3DMMs is quite limited.  It wasn't until a decade later that the first 3D morphable model, the Basel Face Model, became publicly available for use \cite{Pascal5279762}. For past decades, the field of building 3DMMs for public usage has been in substantial progress by many researchers \cite{booth2018large, Brunton2014_Wavelets, FLAME:SiggraphAsia2017, COMA:ECCV18, li2020learning, 3d-morphable-models}. Their dedicated efforts in developing and expanding 3DMM have resulted in a wider range of applications beyond just human faces \cite{allen2003space, anguelov2005scape, loper2015smpl, dai2018data, khamis2015learning, sun2020cafm, zuffi2018lions, shelton2000morphable}.
\subsubsection{Geometry Methods for Facial Capturing}
According to Egger et al.'s categorization, the acquisition of facial correspondence data plays a crucial role in most 3-dimensional morphable models \cite{egger20203d}. Among the prevalent techniques employed for geometry acquisition, multi-view stereo emerges as a prominent approach. Active multi-view stereo involves the identification of identical projected patterns within corresponding pixels across multiple images through the utilization of active illumination patterns \cite{levoy2000digital, intel-realsense-depth-and-tracking-cameras_2022, Geng:11}. Some helpful devices, such as the Cyberware facial scanner, have been used in movie production. The Cyberware facial scanner employs low-intensity laser beams to illuminate the face surface, and a pair of stereo cameras find the same patterns to create object geometry\cite{zhangCyberware}. Similarly, the product 3DMd has adopted the active multi-view stereo methodology, employing infrared light patterns projected onto the face to acquire facial geometry\cite{3dmd_2024}. In the absence of a need for illumination compared to active stereo counterparts, passive multi-view stereo simplifies setups that offer more practicality and accessibility for a broader range of applications. Beeler et al. proposed a facial synthesis pipeline in the paper ``High-Quality Single-Shot Capture of Facial Geometry'', which utilized the passive multi-view images to obtain depth maps \cite{beeler2010high}. Those depth maps are derived by finding the dense correspondences through epipolar lines of two images, allowing for an improved and precise estimation of the facial geometry. 

\subsubsection{Image-Based Photogrammetry}
Recreating a realistic and convincing facial appearance with merely plausible geometry is insufficient, and textures are needed. Image-Based Photogrammetry is also able to get color information beside the facial geometry. The process of obtaining the 2D texture information for the original 3DMM paper \cite{blanz1999morphable} involved a back-projection of the RGB color from scanning cameras. Notwithstanding, the presence of self-shadowing, environmental lighting, or occlusion can introduce artifacts to the final textures. One predominantly used solution to mitigate texture artifacts in textures is capturing the reflectance field of a human face under novel lighting conditions through a light stage introduced by Debevec et al. \cite{debevec2000acquiring}. The light stage provided an array of LED lights that emit light from various angles, intensities, and colors onto a face while simultaneously capturing images with multiple cameras. By computing linear combinations of these images, the dome could provide any incident field of illuminations of a subject. Moreover, the incorporation of polarised spherical illumination to the light stage facilitated the acquisition of high-resolution normals, diffuse albedos, and specular textures \cite{Ma_spherical}.

\subsection{Face Animation}
Rendering photo-realistic images of the human face necessitates a high-resolution polygonal mesh that accurately represents the performer's facial geometry in order to achieve convincing visual results. However, resource constraints and rendering time considerations require a balance between mesh complexity and efficiency. In visual effects and game production, quad-based meshes are commonly employed, although their creation often involves labor-intensive processes \cite{ensticeBenjamin}.Early attempts at generating facial meshes, pioneered by Parke, involved capturing images of an assistant with painted polygons on their face, which were then back-projected to create symmetrical 3D meshes \cite{Park10.1145/800193.569955}. In modern computer graphics animation, commercial software tools like Maya and ZBrush are frequently utilized to create retopologized human facial meshes \cite{patterson2018landmark}. Manual attachment of polygons to 3D facial objects, such as point clouds or triangulated meshes, is performed to align with the underlying muscle flows. Despite the availability of mature 3D tools, the retopology process can still be time-consuming, often taking several hours per individual. To address this challenge, automatic mesh retopologization methods have gained attention. One common approach involves establishing correspondences between a clean mesh and digitalized facial data, which includes facial alignment and landmark detection. These correspondences assist in automating the mesh warping process. Software tools such as Wrap3 leverage these correspondences to attach the clean mesh to the surface of the digitalized scan. Researchers have also explored mapping 3D correspondences directly to 2D images. For instance, Blanz et al. projected landmarks onto a cylinder and flattened it into two 2D textures.  This approach enables the use of image registration optimized by optical flow to establish correspondences  \cite{blanz1999morphable}. Patel et al. utilized thin-plate splines to establish correspondences between manually labeled landmarks on the 3D mesh and the UV space \cite{tps1989}. Another widely utilized technique, known as the Active Appearance Model, is employed to locate landmarks in the UV space for individual faces. These correspondences between the scanned 3D data and the 2D UV space can be leveraged for facial alignment, often employing Iterative Closest Point (ICP) algorithms \cite{ICT2007} to estimate rotation and translation.

Neutral faces alone are insufficient to capture the complexity and range of human facial expressions. Realistic facial rendering requires the incorporation of expressive capabilities. In the production pipeline, facial rigging plays a vital role in animating facial meshes. Riggers enhance facial meshes by introducing joints and control points, enabling animators to manipulate these controls to generate desired facial expressions. The ultimate objective is to create an animatable retopologized facial mesh that can faithfully mimic major facial movements and deformations \cite{patterson2018landmark}. Parke \cite{Park10.1145/800193.569955} employed a collection of facial expressions with consistent topology from a performer, aiming to minimize manual labor. The cosine interpolation scheme was used to manipulate points on the mesh, enabling facial animation. Another approach involves the utilization of the Facial Action Coding System (FACS) as a basis for representing facial expressions \cite{Ekman1978FacialAC}. FACS breaks down expressions into distinct Action Units (AUs), each representing a specific muscle action that contributes to the overall facial posture \cite{Parkebook2008}. By combining different AUs from the FACS repertoire, a wide range of human facial expressions can be effectively portrayed. Interpolation techniques that interpolate meshes between different expressions guided by FACS can be employed to generate realistic human facial expressions.

\section{Photorealistic Facial Rendering}
The primary objective in rendering 3D facial models lies in the generation of photorealistic images that closely emulate actual photographs, achieved through the utilization of a retopologized and animatable facial mesh. Once the mesh is appropriately prepared, the subsequent step involves the creation of image sequences depicting desirable facial expressions. As discussed in Computer Facial Animation \cite{Parkebook2008}, this image synthesis process entails three key tasks: (1) transforming the 3D model into the camera coordinate system and projecting it onto 2D images, (2) determining visible patches, and (3) computing color information for each visible patch. To effectuate the projection of 3D models onto 2D images, model-view-projection matrices are employed, where a model matrix M encompasses transformation T, rotation R, and scale, facilitating object placement in world space. Subsequently, the camera matrix repositions all objects into the eye coordinate system relative to the camera's position. Once all objects reside in camera space, they undergo transformation into clip coordinates. In photorealistic rendering, perspective projection is commonly employed to simulate real camera behavior. In the traditional computer graphics pipeline, z-buffer testing is utilized to determine hidden surfaces. The final step involves selecting shading models for rendering. In computer graphics, a 2D image is represented as an array of pixels, and determining the color of each pixel is crucial for achieving photorealistic rendering. Retopologized facial meshes often consist of quadratic polygons, which are further subdivided into triangles through tessellation in the computer shading pipeline. In the early days of facial rendering, each vertex of a triangle was represented by a single color of skin, obtained by back-projection of RGB color from individual scanning cameras \cite{blanz1999morphable}. Texture mapping is employed to apply albedo images back to 3D models, and pixel colors within a triangle can be interpolated using barycentric coordinates \cite{Blinn1976Texture}.  Gouraud's smooth shading averages normal vectors at each vertex of the triangle shared by different patches to compute lighting, resulting in a smooth visual effect suitable for various lighting modes \cite{gouraud1971continuous}. To approximate the lighting effect in 3D scenes, Phong introduced an algorithm that consists of ambient, diffuse, and specular components \cite{phong1975illumination}. Ambient lights simulate the impact of indirect or global illumination, diffuse lights model the scattering of light on rough surfaces following the cosine law, and specular lights represent shiny highlights reflected from a surface. Phong's approach has been widely used for decades in most 3D production software, providing a basic lighting model. In contrast, Physically Based Rendering (PBR) enables the generation of more accurate and realistic images by evaluating the physical properties of 3D scenes. PBR relies on a model that describes how light interacts with material surfaces. The bidirectional reflectance distribution function (BRDF) is commonly used to approximate how much each incoming ray will be reflected to the outgoing ray on a rough surface. In 1967, Torrance and Sparrow proposed a statistical microfacet model to understand light reflection from roughened surfaces\cite{torrance1967theory}. The model assumed that roughened surfaces consist of a collection of small, mirror-like microfacets and derived a BRDF based on a normal distribution function, a geometry attenuation factor (shadowing-masking function), and the Fresnel reflectance term.


%% file: relatedWork.tex
\chapter{Related Work}
In the statistical domain of 3D facial synthesis from images, the most common framework is called analysis-by-synthesis. The task of analysis-by-synthesis is minimizing the energy between the observed images and synthesizing an estimated 3D face \cite{egger20203d}. There are multiple modalities of representing 3D faces, though, 3DMMs have been widely used for image-based reconstruction. The fundamental concept underlying the construction of a statistic 3D Morphable Model involves expressing new faces with a linear combination of a set of eigenvectors derived from a pool of facial meshes with fixed connectivity through principal component analysis. The parameters associated with new faces can be determined by projecting them onto the eigenvectors. Minimizing the difference between 3DMM and observed images could be easily done by adjusting the parameters to be the best fit. However, recent attention to facial synthesis methods, particularly with the application of deep convolutional neural networks, has yielded notable improvements. This chapter focuses on how deep neural networks demonstrated potential capability for learning a mapping to 3D face meshes directly from facial photos.

\section{Appearance 3D Morphable Model}
Active Shape Model (ASM) \cite{COOTES199538} and Active Appearance Model (AAM \cite{cootes1998active}) initially employed Principal Component Analysis (PCA) to model and reconstruct shapes from 2D images. Extending this foundational approach, the 3D Morphable Model (3DMM) \cite{3d-morphable-models} was afterward developed to facilitate the reconstruction of three-dimensional facial shapes. The 3DMM represents a statistical, parametric framework that encapsulates both the geometry and texture information of human faces. By leveraging PCA, 3DMM enables the generation and manipulation of realistic face models, supporting a range of tasks including facial synthesis, 3D synthesis from single or multiple images, expression transfer, and robust face recognition across varying poses and illumination conditions.

\subsection{Principal Component Analysis} \label{sec:pca}
\begin{figure}[!htb]
    \centering
    \includegraphics[width=0.5\linewidth]{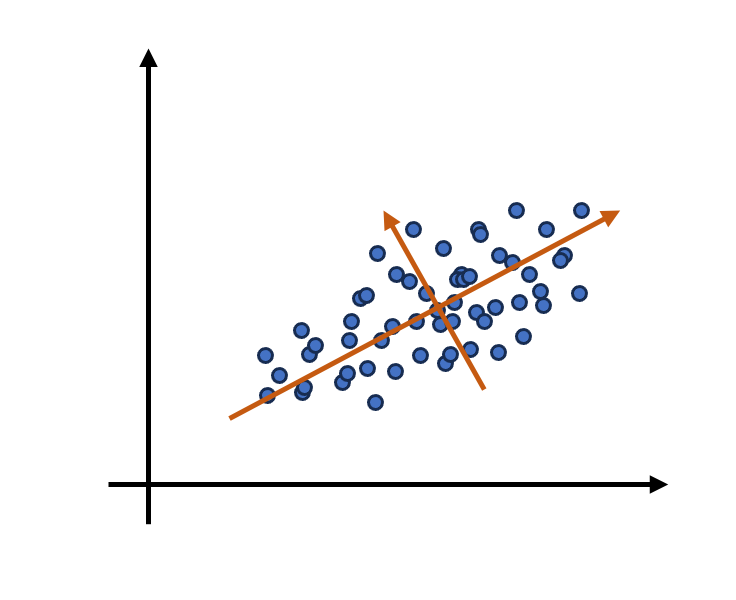}
    \caption{Illustration of Principal Component Analysis}
    \label{fig:pcs}
\end{figure}

The core concept of parametric models involves utilizing Principal Component Analysis (PCA) on a given training dataset to identify orthogonal vectors that span a new representation space. These orthogonal vectors are mathematically determined to efficiently reduce the dimensionality of the original data while retaining the dataset's essential features. PCA reduces the representation of high-dimensional data using fewer components, removing redundant information and extracting primary variations present in the dataset. Figure \ref{fig:pcs} illustrates how PCA calculates a new set of orthogonal vectors (shown in red) to represent the original data points (blue). The orthogonal vectors depict the principal directions along which the data exhibits maximum variance.

Given a set of data points $x_i$, the mean-subtracted dataset is computed by:
\begin{align}  
    X &= x_i - \overline{x}, \quad \overline{x} = \frac{\sum^{n}_{i=1}x_i}{n}
\end{align}

A covariance matrix C then can be computed by:

\begin{align}  
    C &= \frac{X^TX}{n-1}
\end{align}
Here, C is d \(\times\) d symmetric matrix.The covariance matrix can then be decomposed through eigenvalue decomposition as follows:

\begin{align}  
   Cv_i &= \lambda_i v_i
\end{align}
where $v_i$ are the principal components, also known as eigenvectors, and $\lambda_i$ are eigenvalues which describe the variance encaptured by the corresponding eigenverors and $\lambda_i$ are eigenvalues that represent the variance captured by the corresponding eigenvectors. If the eigenvectors are rearranged in descending order according to their eigenvalues: $\lambda_1 \geq \lambda_2 \geq \lambda_3 \geq \cdots \geq \lambda_n  $ These eigenvectors form the projection matrix
\begin{align}  
   P &= [v_1, v_2, v_3, \cdots, v_n]
\end{align}
The first k eigenvectors, corresponding to the largest eigenvalues, represent the primary components of the training dataset. Any data point can be approximated by a parameter set $b$ by projecting it into the eigenvector space:

\begin{align}  
   b &= P^T(x_i - \overline{x})
\end{align}

\subsection{3D Morphable Model}
Adopting Principal Component Analysis (PCA), a 3D Morphable Model (3DMM) efficiently represents the 3D shape and appearance of human faces using a relatively small number of parameters. To construct a 3DMM, a collection of face meshes with corresponding texture maps is carefully selected. These training meshes must adhere to a consistent topology, meaning each mesh contains the same number of vertices, with corresponding vertices on different meshes sharing the same semantic meaning (e.g., corners of the eyes, tip of the nose). Each face can thus be represented by a shape vector $S_i$ a mean shape vector $\overline{S}$ a texture vector $T_i$  and a mean texture vector $\overline{T}$, defined as:

\begin{align}  
   S_i &= [x_1, y_1, z_1, \cdots, x_n, y_n, z_n]\\
   T_i &= [r_1, g_1, b_1, \cdots, r_n, g_n, b_n]
\end{align}
where $x_i, y_i,$ and $z_i$ epresent the 3D coordinates of the mesh vertices, and $r_i, g_i,$ and $b_i$ represent the RGB color values of the corresponding points on the texture map.
Following the PCA decomposition method outlined in Section \ref{sec:pca}, two eigenvector matrices $P_s$ and $P_t$ corresponding to shape and texture can be derived. Therefore, any new face can be synthesized using linear combinations of these eigenvectors:

\begin{align}  
   S &= \overline{S} + P_sb_s\\
   T &=  \overline{T} + P_tb_t
\end{align}

\subsection{PCA Applications
}
Active Shape Models (ASM), introduced by Cootes et al. \cite{COOTES199538}, have served as foundational techniques for modeling deformable objects, especially human faces. ASM captures shape variations by learning statistical distributions of landmark points from a training dataset. Later, Cootes et al. \cite{cootes1998active} incorporated appearance information into the ASM framework, developing the Active Appearance Model (AAM), an enhanced version that simultaneously models both shape and appearance. Blanz and Vetter \cite{blanz1999morphable} advanced the concept of appearance models from 2D into 3D space by introducing the 3D Morphable Model (3DMM). This model represents facial geometry and appearance through PCA-based low-dimensional linear models trained on collections of 3D facial scans. Several publicly available 3D Morphable Models, including the Basel Face Model (BFM) \cite{bfm09} and FLAME \cite{FLAME:SiggraphAsia2017}, have since emerged, improving upon earlier models with enhanced expression accuracy, stronger generalization across diverse identities, and increased fidelity of reconstructed facial details.

\section{Light Stage: Facial Light Field Acquisition}
3D Morphable Models (3DMMs) are statistical representations derived from datasets containing scanned human subjects. However, generating faces from linear combinations of eigenvectors tends to omit high geometric frequencies and subtle reflectance properties, leading to decreased visual realism. In current applications such as films and video games, digital characters lacking convincing human appearance or realistic dynamic behavior are often dismissed, limiting their usefulness and potentially undermining research that relies on them.

A significant advancement in realistic digital character creation has been the development of the Light Stage system at the Institute of Creative Technology \cite{debevec2000acquiring}. A typical Light Stage is composed of a dome or spherical structure densely equipped with individually controllable light sources, enabling precise simulation of complex illumination conditions. Multiple cameras positioned around the Light Stage simultaneously capture both the geometry of subjects at the center and their detailed reflectance characteristics. Using this technique, the Light Stage not only effectively captures accurate 3D geometry but also produces highly realistic appearance and texture maps, greatly enhancing the realism of digital characters.

\subsection{Multiview Stereo Reconstruction}
\begin{figure}[!htb]
    \centering
    \includegraphics[width=0.8\linewidth]{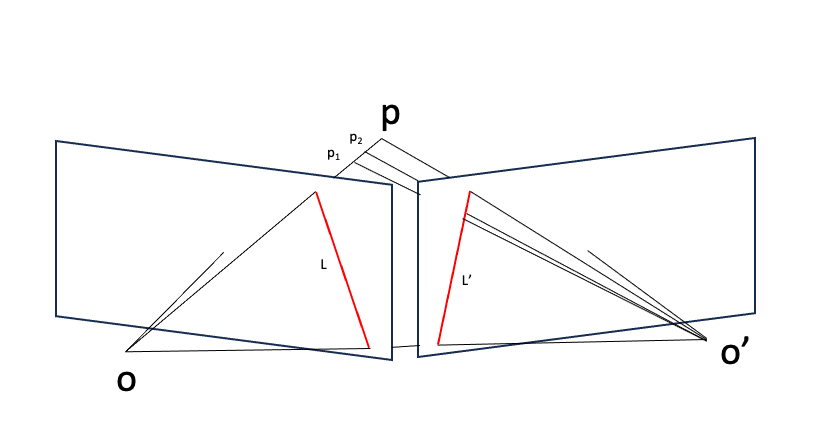}
    \caption{Epipolar Geometry}
    \label{fig:epipolar}
\end{figure}
One practical application of the Light Stage is the synthesis of 3D geometry from multiple images captured from slightly different viewpoints around a subject. This approach, known as Multi-view Stereo (MVS) Reconstruction, allows a detailed 3D model to be derived directly from these 2D images. The fundamental principal of MVS relies on identifying corresponding points across image pairs. Specifically, when two images of a subject are captured from different angles, a pixel from one image can be associated with its corresponding pixel in the other image, allowing reconstruction of the original point's 3D position through triangulation. Figure \ref{fig:epipolar} illustrates an example of stereo reconstruction: any point along the ray OP in the left image has a corresponding point along the line L' in the right image. To determine the 3D coordinates of a pixel in the left image, it is sufficient to search for its matching point along the corresponding line L' in the right image. These corresponding lines, denoted as L and L', are epipolar lines, while the plane defined by the points P, O, and O' is known as the Epipolar Plane. Using these geometric relationships, it is possible to reconstruct accurate 3D surface points from pairs of stereo images.

\subsubsection{Image Rectification}
\begin{figure}[!htb]
    \centering
    \includegraphics[width=0.8\linewidth]{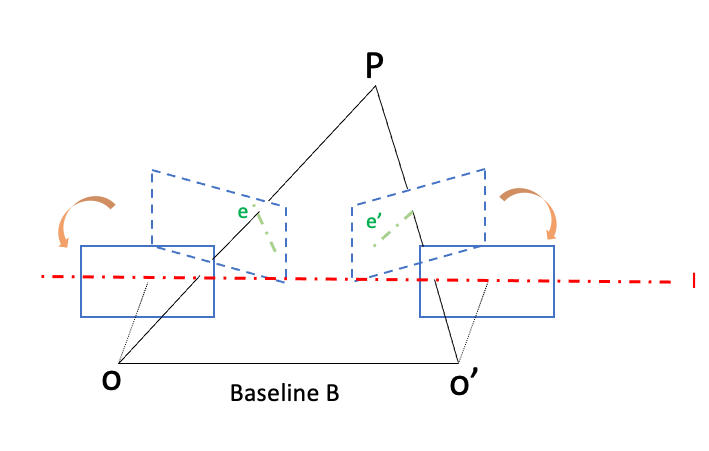}
    \caption{Epipolar Line Alignment through Image Rectification}
    \label{fig:imgrec}
\end{figure}

Without image rectification, searching for pixel correspondences involves matching points along arbitrary epipolar lines. This procedure is computationally demanding and inefficient because epipolar lines are typically skewed or tilted. Image rectification is used to address this inefficiency. Rectification is the process by which pixels in stereo images are rearranged so that all epipolar lines become horizontal and parallel. As a result, the complexity of the search for correspondences is significantly reduced to a straightforward one-dimensional horizontal search within the same image row.
Figure \ref{fig:imgrec} illustrates how epipolar lines change after applying the image rectification process. Before rectification, the epipolar lines e and e' are tilted, complicating the correspondence search. After rectification, the images are rotated and translated such that the epipolar lines become horizontally aligned. Therefore, correspondence searches become more computationally efficient by limiting them to aligned epipolar lines. The distance between the two camera centers O and O' is referred to as the baseline B.

A classic algorithm for image rectification is Hartley's method \cite{hartley1999theory}. Suppose two stereo images have corresponding camera matrices defined as $P_1 = K_1[R_1 | T_1]$ for the left camera and $P_2 = K_2[R_2 | T_2]$, where $P_i$ denotes the projection matrix,  $K_i$ the intrinsic matrix, and $R_i$ and $T_i$ are rotation and translation matrices, respectively, combining to form the extrinsic camera matrix. The camera centers $C_i$ can then be computed by $C_i = -R_i^TT_i$ where i = 1, 2. Initially, a new coordinate system for rectification is computed from the stereo camera system, defined as follows:

\begin{align}  
    r_x &= \frac{C_2 - C_1}{|C_2 - C1|} \\
    r_y &=  \frac{r_{1,3} - r_x}{|r_{1,3} - r_x|} \quad \quad \text{(\(r_{1,3}\) is the 3rd row of\(R_1\))}\\
    r_z &= r_x \times r_y
\end{align}

Once the new rectified coordinate system is defined, the rectification rotation matrix $R_{rec}$ and intrinsic matrix $K_{rec}$ are computed as follows:
\begin{align}  
    R_{rec} &= \begin{bmatrix} r_x^T\\r_y^T\\r_z^T\end{bmatrix}\\
    K_{rec} &=\frac{K_1 + K_2}{2}
\end{align}

Finally, pixels in the original images can be mapped to their new rectified positions using homography transformations $H_1$ and $H_2$ defined as:

\begin{align}  
    H_1 &= K_{rec}R_{rec}R_1^TK_1^{-1}\\
    H_2 &= K_{rec}R_{rec}R_1^TK_2^{-1}
\end{align}

Given an original pixel position x, the corresponding pixel coordinates in the rectified image x'
  are computed through:

\begin{align}  
    x^{'} = Hx
\end{align}

This transformation simplifies the process of finding pixel correspondences across stereo image pairs by aligning the epipolar lines horizontally, thus significantly reducing computational complexity.

\subsubsection{Pixel Pairs \& Depth Map}
Following stereo image rectification, corresponding pixels between two stereo images can be efficiently identified by searching along the aligned epipolar lines. A widely utilized metric for measuring pixel similarity is the Normalized Cross-Correlation (NCC). For each pixel pair, neighboring pixels are typically extracted to form a comparison window of size N \(\times\) N. Given two corresponding windows from the stereo images, $W_l$ (left image) and $W_r$(right image), the NCC is defined as follows:

\begin{align}  
    NCC(W_l, W_r) &= \frac{\sum_{x,y}(W_l(x, y) -\overline{W_l})(W_r(x, y) -\overline{W_r})}{\sqrt{\sum_{x,y}|(W_l(x, y) -\overline{W_l})|^2}\sqrt{\sum_{x,y}|(W_r(x, y) -\overline{W_r})|^2}}
\end{align}

Where $\overline{W_l}$ and $\overline{W_r}$ represent the mean pixel intensity values within each window.

\begin{figure}[!htb]
    \centering
    \includegraphics[width=0.5\linewidth]{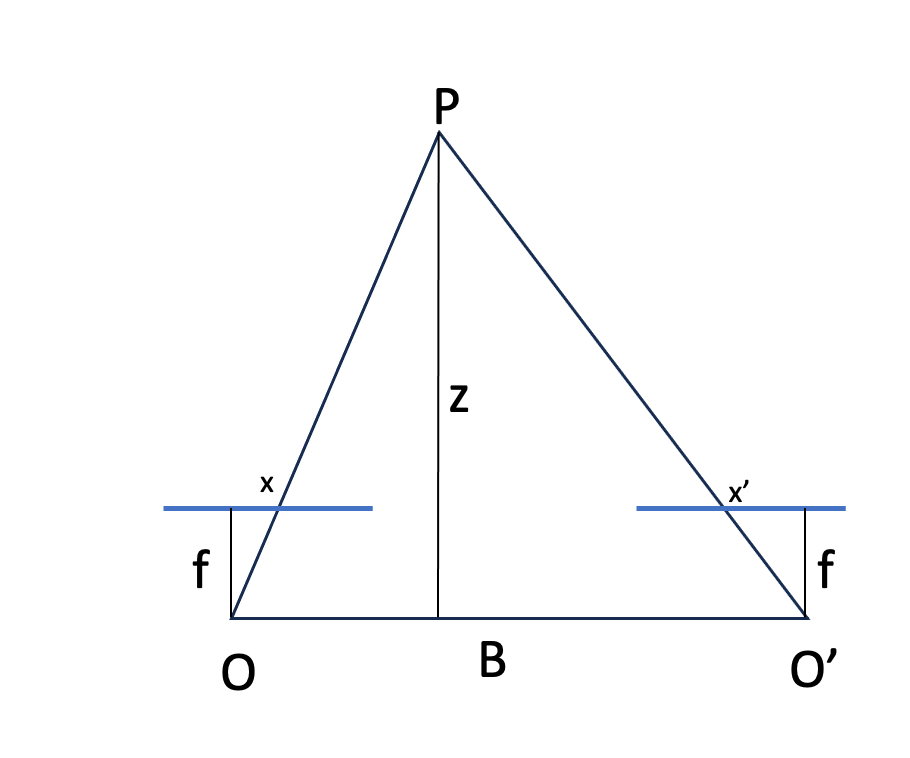}
    \caption{Disparity map generation from stereo images}
    \label{fig:depthmap}
\end{figure}

As illustrated in Figure \ref{fig:depthmap}, once pixel correspondences between the stereo images are determined, a disparity map—representing horizontal pixel-coordinate differences—can be computed. This disparity map is subsequently used to derive depth information in 3D space.

The disparity d and the depth z are calculated as follows:

\begin{align}  
    d &= x - x'\\
    z &= \frac{Bf}{d}
\end{align}
Where B denotes the baseline distance between camera centers, f s the camera's focal length, and x and x' represent the corresponding pixel coordinates along the x-axis in the left and right images, respectively.

\subsection{Textures Acquisition}
\begin{figure}[!htb]
    \centering
    \includegraphics[width=0.8\linewidth]{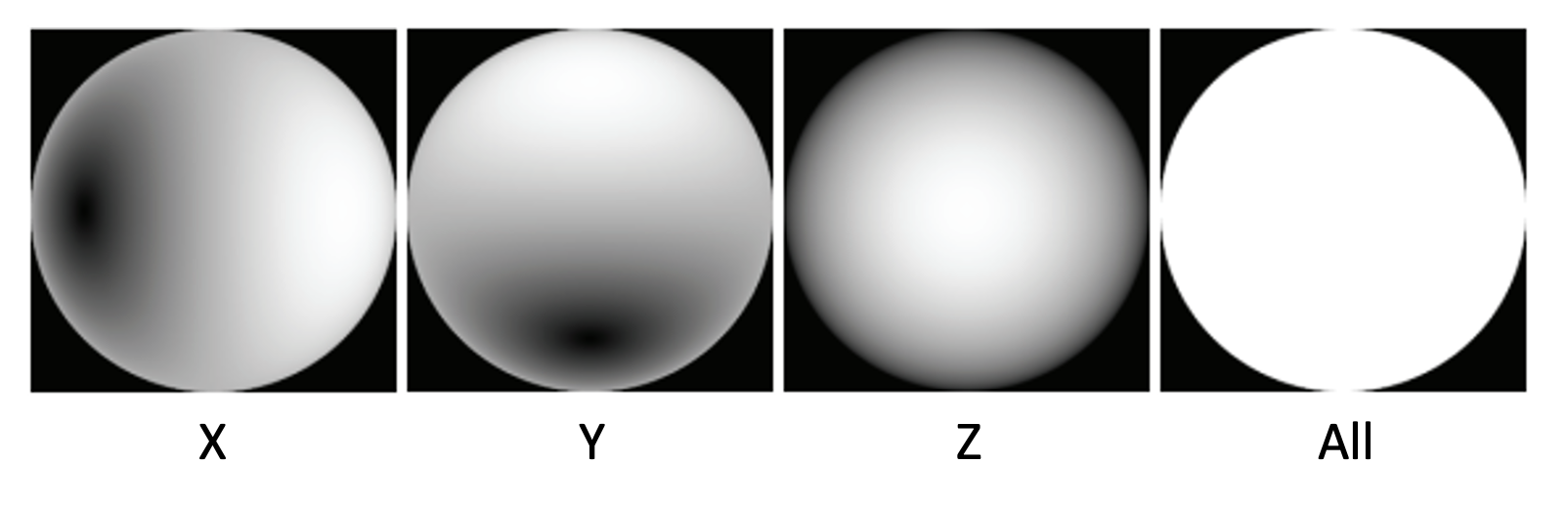}
    \caption{Gradient lighting patterns along X, Y, and Z axes. The "All" condition refers to uniform illumination from all directions.}
    \label{fig:gradientLight}
\end{figure}

The Light Stage not only enables the acquisition of high-resolution geometry but is also capable of capturing normal, diffuse, and specular reflectance maps. The system consists of hundreds of LED light sources arranged spherically around the subject. By systematically varying the direction, polarization, and intensity of illumination, the Light Stage captures the subject’s appearance under a rich set of lighting conditions. Surface normals can be accurately recovered from images captured under spherical gradient illumination patterns. Typically, three gradient lighting conditions aligned with the global X, Y, and Z axes are used. Figure \ref{fig:gradientLight} shows an example of these gradient patterns, along with a full-illumination (ambient) image.

The surface normal at each pixel can be estimated by first computing the gradient components as ratios of the directional gradient images to the ambient image:

\begin{align}  
    dx &= \frac{img_{dX}}{img_{dA}}, dy = \frac{img_{dY}}{img_{dA}}, dz = \frac{img_{dZ}}{img_{dA}}\\
    normal &= \frac{[dx, dy, dz]}{||[dx, dy, dz]||}
\end{align}

\begin{figure}[!htb]
    \centering
    \includegraphics[width=0.5\linewidth]{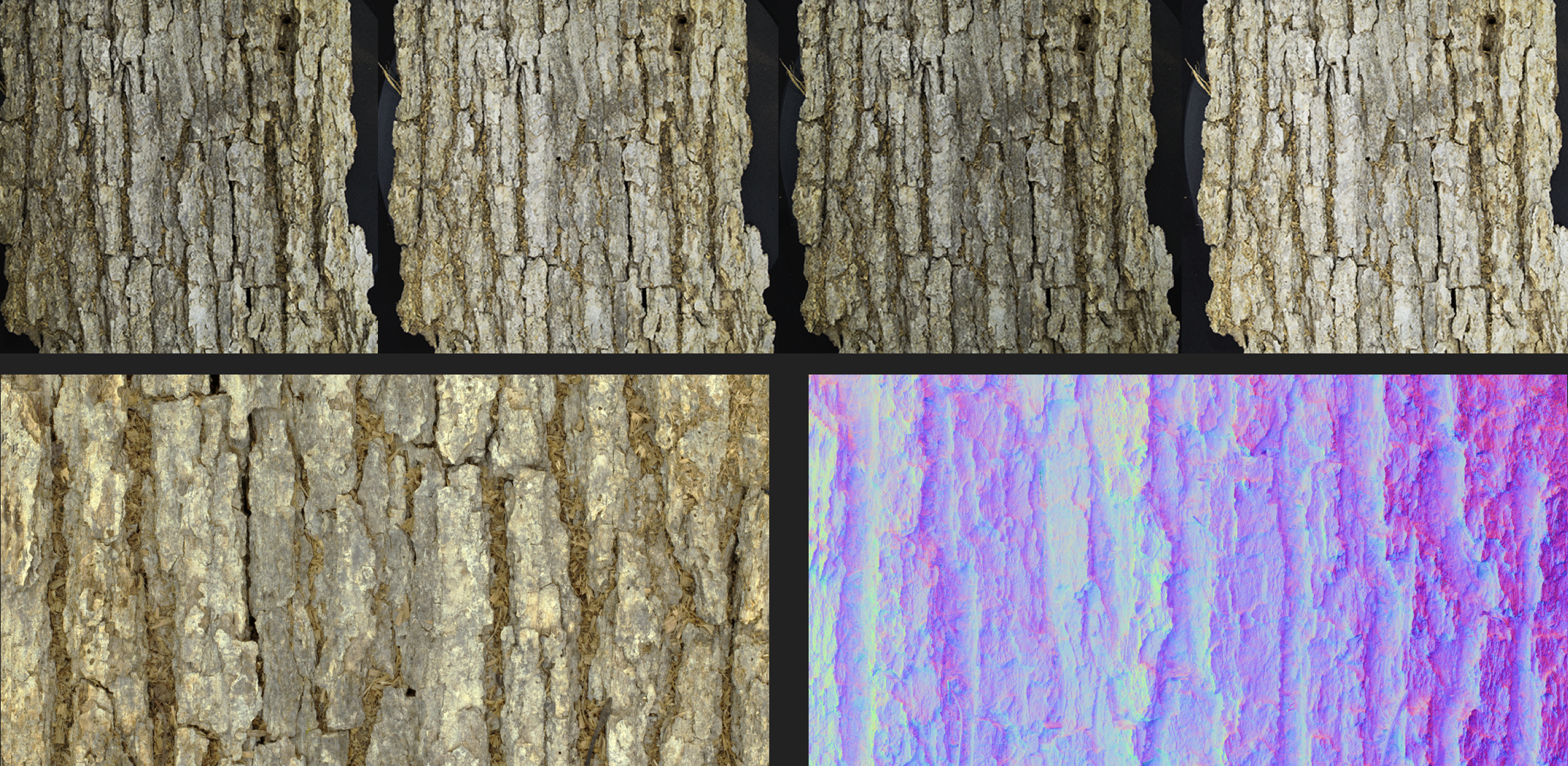}
    \caption{Surface normals recovered from gradient lighting patterns.}
    \label{fig:normal}
\end{figure}

This produces a unit normal vector n per pixel, encoding the surface orientation. Figure \ref{fig:normal} shows a normal map computed using this method.

A diffuse map (also called an albedo map) represents the object's intrinsic color under purely diffuse (subsurface) reflection. To separate diffuse and specular reflections, polarizers are mounted on both the lights and camera lenses. Two images are captured per lighting condition:
\begin{itemize}
    \item Parallel polarization image $I_{\parallel}$: the camera polarizer is aligned with the light source polarizers; this image contains both diffuse and specular reflections.
    \item Cross polarization image $I_{\bot}$: the camera polarizer is orthogonal to the light polarizers; this image contains primarily diffuse reflection, with specular reflection largely eliminated.
\end{itemize}

\begin{figure}[!htb]
    \centering
    \includegraphics[width=0.5\linewidth]{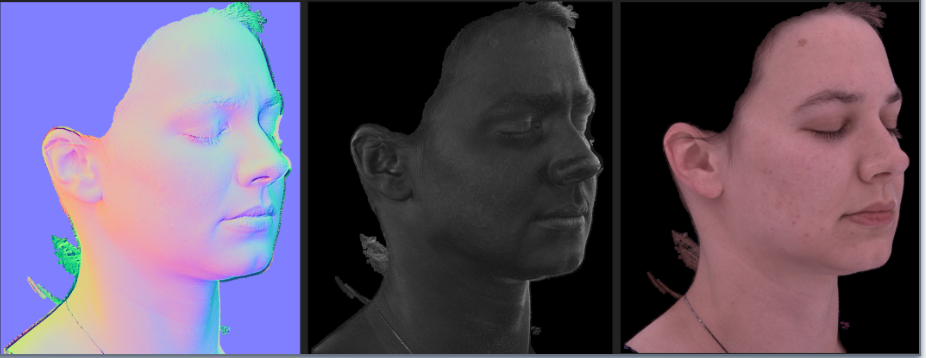}
    \caption{Normal, Specular and Diffuse maps from active illumination patterns with polarizations}
    \label{fig:maps}
\end{figure}

From these two images, the diffuse and specular components can be computed as:

\begin{align}  
    I_{diffuse} &= I_{\bot}\\
    I_{specular} &= I_{\parallel} - I_{\bot}
\end{align}
Figure \ref{fig:maps} shows an example of the computed normal, specular, and diffuse albedo maps from a single viewpoint, derived from the captured gradient illumination data.

\subsection{Related Work}
One significant advancement toward creating realistic digital characters was the development of Light Stage systems by Debevec at the Institute of Creative Technology \cite{debevec2000acquiring}. The Light Stage employs programmable spherical illumination combined with multiple high-resolution cameras, enabling the precise capture of detailed facial reflectance and geometry. To create accurate facial 3D meshes, Beeler et al. introduced a passive, single-shot multiview stereo vision system capable of efficiently recovering detailed facial geometry without the need for active illumination. This innovation became instrumental in the creation of Disney's renowned facial capture system, Medusa, which has been extensively used in numerous live-action films to achieve convincing digital-human performances. Further advancing these techniques, Ma et al. \cite{Ma_spherical} first introduced the use of gradient-based spherical illumination patterns to simultaneously extract high-resolution normal maps, specular reflectance, and diffuse albedo textures from human faces. These pioneering works collectively established critical methodologies that now underpin many state-of-the-art facial capture and rendering systems in contemporary visual effects production.

\section{Deep Learning in Facial Synthesis}

\begin{figure}[!ht]
    \centering
    \includegraphics[width=0.8\linewidth]{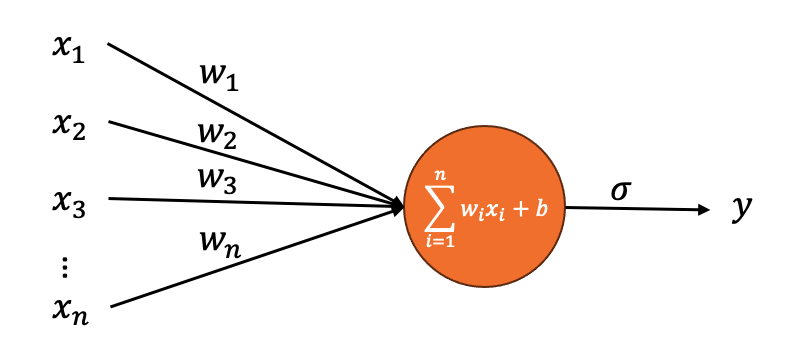}
    \caption{A Single Neuron}
    \label{fig:singleNeuron}
\end{figure}

In 2018, the research conducted by Johnston and Chazal wrote a survey and they observed a significant shift of interest towards deep-learning techniques, motivated by their potential to enhance performance and facilitate the execution of 3D alignment and construction tasks\cite{johnston2018review}. The origins of deep learning can be traced back to 1943, when McCulloch and Pitts pioneered the development of a rudimentary neural network employing electrical circuits\cite{mcculloch1943logical}. This early endeavor served as an inspiration for subsequent advancements in neural network research. Subsequently, Frank Rosenblatt introduced the fundamental machine learning algorithm, known as the Perceptron, a precursor to modern deep learning methods\cite{rosenblatt1958perceptron}. The Perceptron exemplified a model that sought to emulate the functionality of neurons within the human brain, establishing a pivotal foundation for the evolution of deep learning. Figure \ref{fig:singleNeuron} represents a single neural network neuron.

\( [x_1, x_2, x_3, ..., x_n] \in \mathbb{R}^n \) denotes a \(n\) dimension vector containing the input parameters; \( [w_1, w_2, w_3, ..., w_n] \in \mathbb{R}^n \) are the corresponding adjustable weights that will be tuning during the training process; \(b\) is the bias for this neuron; \(\sigma\) is the activation function; \(y\) is the output. With \(n\) dimensional inputs, the output \(y\) is calculated as 

\begin{align}  
    & z = \sum_{i=1}^{n} w_ix_i + b = w_1x_1 + w_2x_2 + w_3x_3 + ... + w_nx_n + b  \label{eq:weights_inputs_multi}\\
    & y = \sigma(z) \label{eq:apply_activation}
\end{align}

A frequently utilized approach to train deep learning models is through the implementation of the backpropagation algorithm \cite{robbins1951stochastic}. This method intricately utilizes a gradient descent optimization strategy, systematically adjusting the model's parameters for improved performance. Specifically, for instance, the gradient \(\partial w_i\) can be calculated by deriving the partial derivative of \(y\) with respect to the variable \(w_i\). In this case, the partial derivative \( \frac{\partial y}{\partial w_i} \) can be computed as:

\begin{align} 
    dw_i = \frac{\partial y}{\partial w_i} =  \frac{\partial y}{\partial z} \frac{\partial z}{\partial w_i}
\end{align}

The weight \(w_i\) can be updated as \(w_i = w_i - \alpha dw_i\) where \(\alpha\) is the learning rate.

\begin{figure}[!ht]
    \centering
    \includegraphics[width=0.8\linewidth]{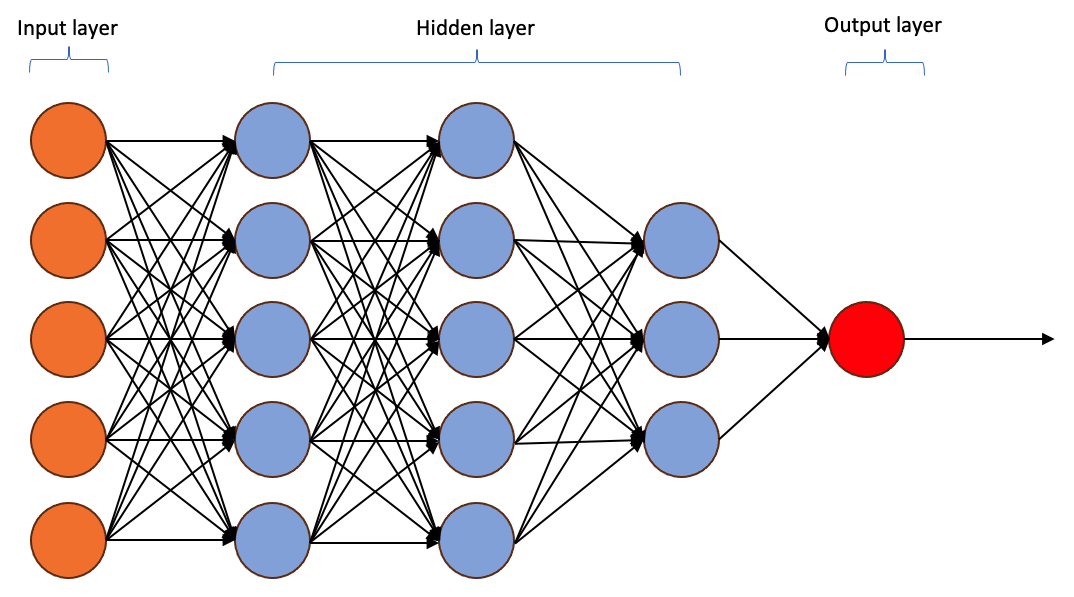}
    \caption{A Simple Neural Network}
    \label{fig:simpleNN}
\end{figure}

In a neural network, a layer comprises a vertical arrangement of individual neurons, with multiple layers collectively forming a neural network. Figure \ref{fig:simpleNN} illustrates a basic neural network architecture. Generally, a neural network can be divided into three main components: the input, hidden, and output layers. Determining the appropriate number of neurons within a layer and the number of hidden layers is a task that depends on the specific project and its requirements.

\subsection{Convolutional Neural Network}

As illustrated in Figure \ref{fig:singleNeuron}, a neural network's input is represented as an n-dimensional vector \( [x_1, x_2, x_3, ..., x_n] \). In the domain of computer vision, particularly image regression tasks. However, conventional neural networks face scalability constraints. For instance, processing a \( 100 \times 100 \times 3 \) image requires training a substantial number of weights in the input layer, totaling 30,000. To address this challenge and reduce the number of training parameters significantly, Convolutional Neural Network (CNN) has been developed, specifically designed to handle images.

\subsubsection{Convolutional Layers}
At the heart of a Convolutional Neural Network (CNN) lies the fundamental operation of convolutional layers, which employ filters, also known as kernels, to perform convolutions over input images. These convolutional layers yield a collection of feature maps, each of which encapsulates distinct features extracted from the input images, encompassing elements like edges, components, or even intricate object structures.

\begin{figure}[!ht]
    \centering
    \includegraphics[width=0.8\linewidth]{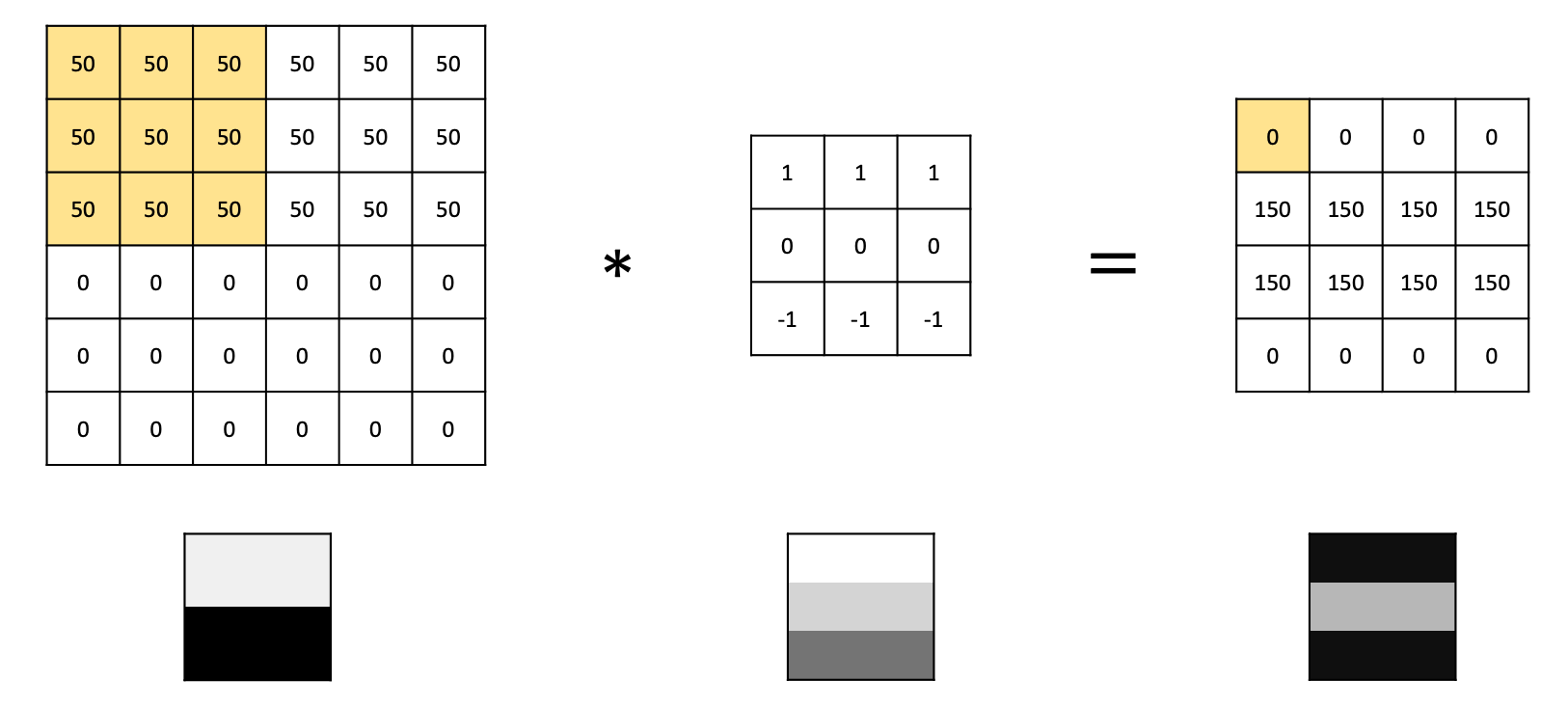}
    \caption{Convolution Operation on an Image with \(3 \times 3\) filter}
    \label{fig:ConvolOp}
\end{figure}

\begin{figure}[!ht]
    \centering
    \includegraphics[width=1\linewidth]{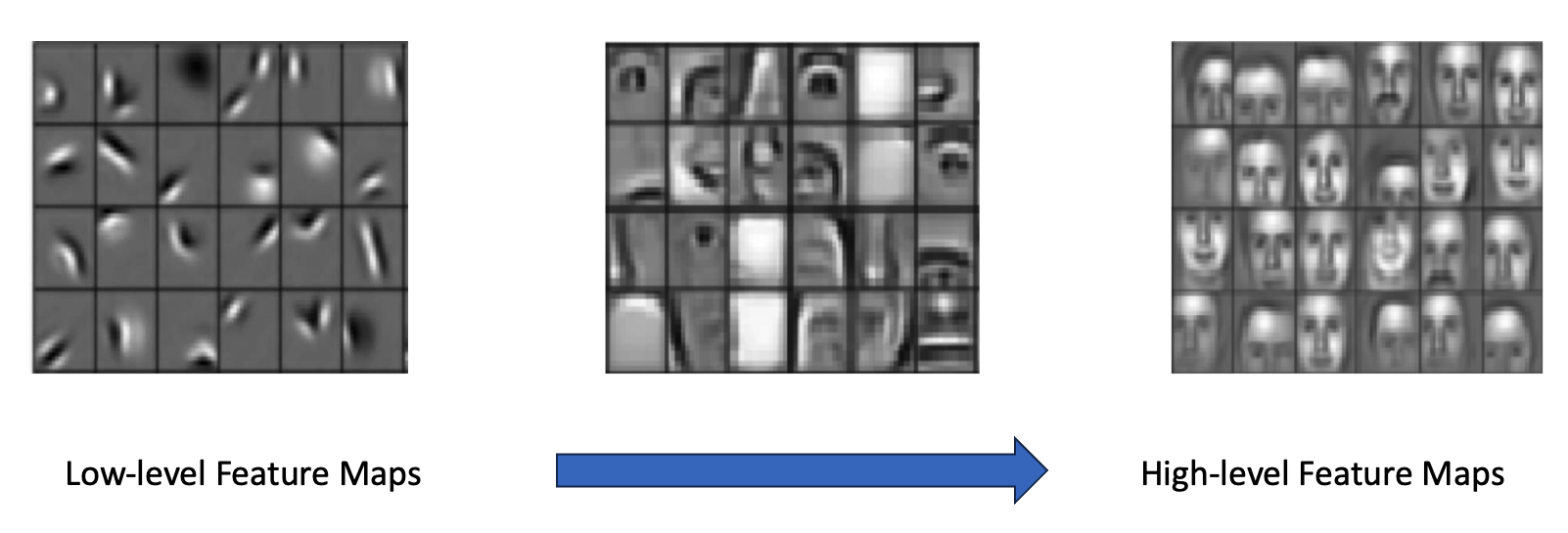}
    \caption{Convolutional Operation Extract More Concrete Features As Network Goes Deeper \cite{Ngcoursera}}
    \label{fig:cnnFeature}
\end{figure}

In Figure \ref{fig:ConvolOp} explains the fundamental procedure of applying a 3 \( \times \) 3 filter across a 6 \( \times \) 6 image, providing a detailed exposition of the process. This filter systematically scans a segment of the 3 \( \times \) 3 region within the image, executing multiplication and summation operations to calculate a new pixel value. As the filter traverses the entire image, it systematically searches for specific patterns, creating a feature map. Furthermore, as depicted in Figure \ref{fig:ConvolOp}, the 6 \( \times \)6 image exhibits a notable distinction in pixel values, with higher intensities at the top and lower values at the bottom. This distinctive pattern enhances the presence of a horizontal line crossing the image. The applied filter has the same values along each row and predominantly identifies horizontal features within the image. Upon convolution, the feature map shows the ability to remove horizontal edges within the image. At each layer of the neural network, different filters are applied systematically to the input image, resulting in the creation of feature maps. This convolutional procedure reduces the image's dimensions, consequently diminishing the number of parameters that necessitate learning throughout the training phase. This is because the same feature patterns are consolidated within the same feature map, which allows shared learning parameters for a given feature map. As the neural network deepens, the convolutional neural network progressively filters out features, starting from simple line segments and gradually evolving to more complex parts of objects, as shown in Figure \ref{fig:cnnFeature}. Ultimately, the network can even detect entire objects.

\subsubsection{Downsampling}
\begin{figure}[htb!]
    \centering
    \includegraphics[width=1\linewidth]{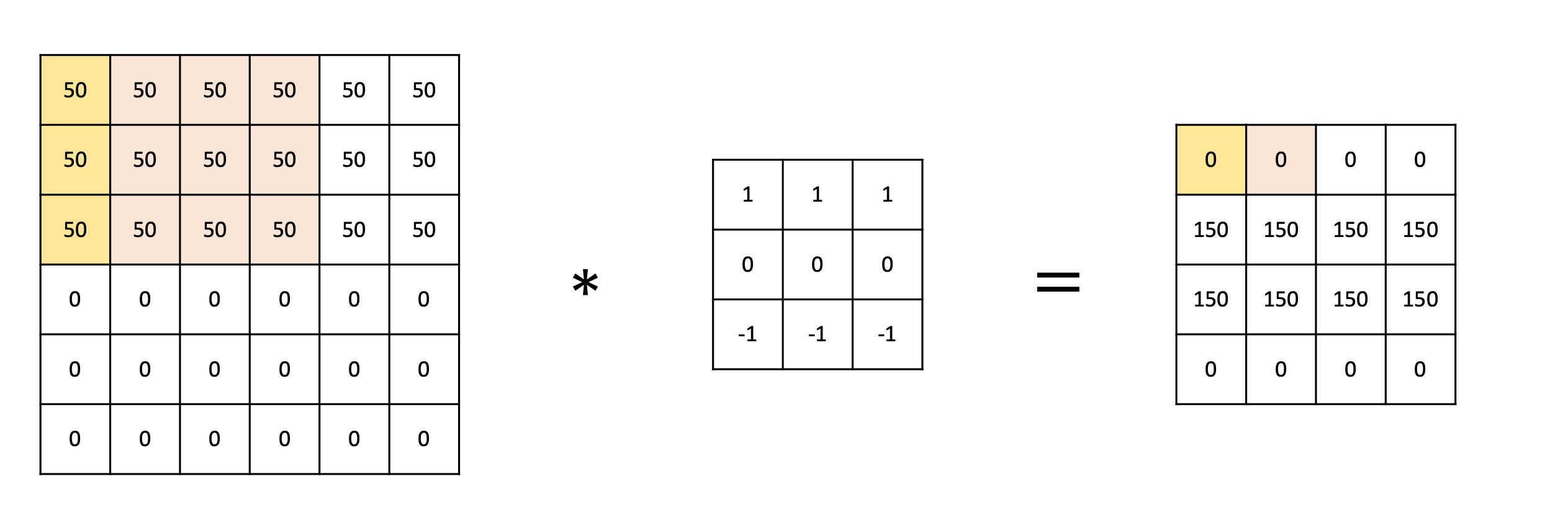}
    \caption{Illustration of the convolutional filter shifting one pixel horizontally to perform convolution at the subsequent location}
    \label{fig:stride}
\end{figure}
To enhance computational efficiency, Convolutional Neural Networks (CNNs) implement several optimization techniques, especially stride and pooling operations. Stride involves sliding convolutional filters across the input image at defined intervals. For example, Figure \ref{fig:stride} illustrates a 3  \( \times \) 3 filter shifting horizontally by one pixel at each step, performing a convolution operation in each position. Utilizing stride effectively reduces the dimensions of resulting feature maps. Specifically, given an input image of size m \( \times \) n, a filter dimension f \( \times \) f and stride s, the output feature map dimensions become:

\begin{align}  
    \left\lfloor \frac{m - f}{s} + 1 \right\rfloor
    \times 
    \left\lfloor \frac{n - f}{s} + 1 \right\rfloor 
\end{align}

\begin{figure}[htb]
    \centering
    \includegraphics[width=1\linewidth]{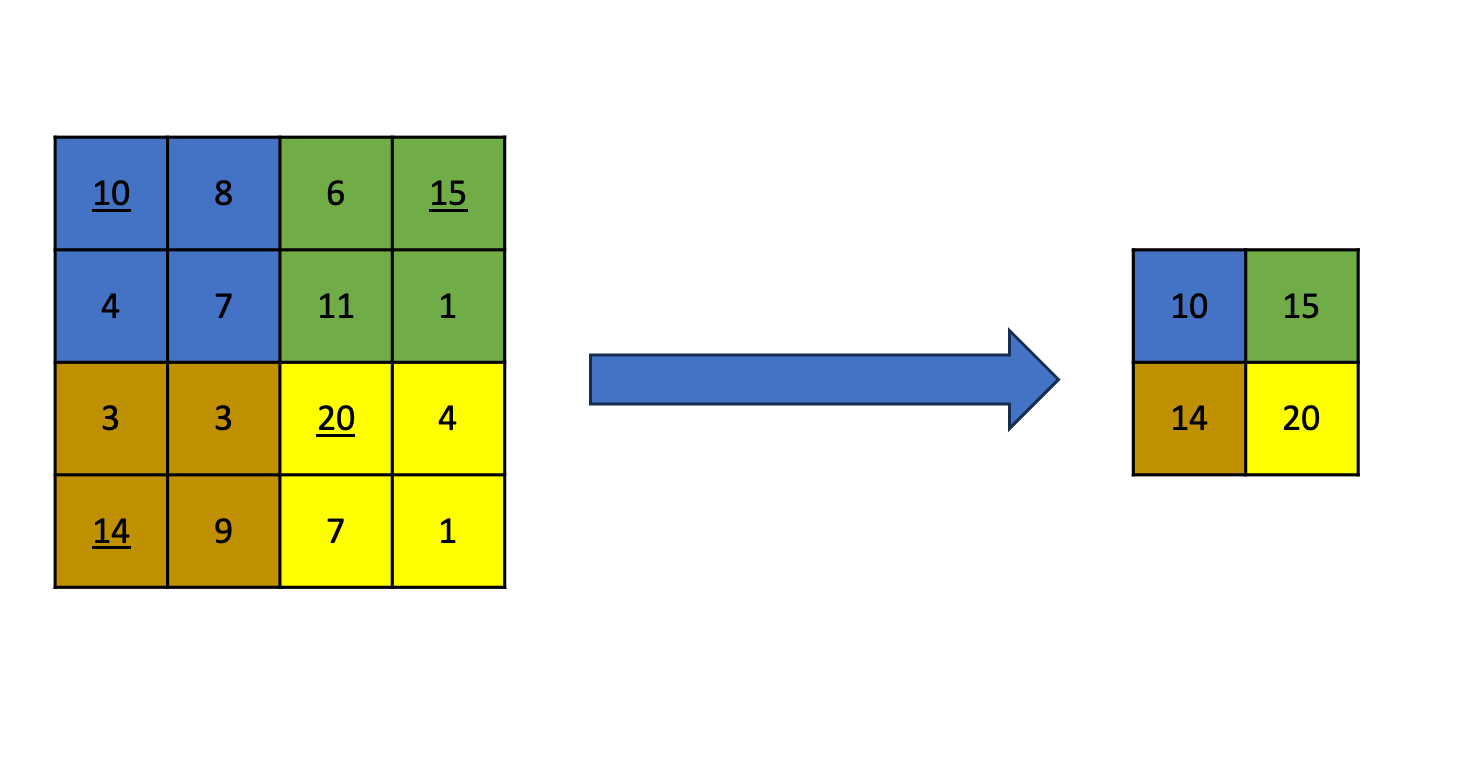}
    \caption{Max Pooling operation: A pooling window traverses the feature map, selecting the maximum value within each region to produce a reduced feature map.}
    \label{fig:maxpool}
\end{figure}

Pooling layers further reduce computational complexity by compressing the spatial dimensions of feature maps and emphasizing prominent features. A widely adopted pooling strategy is Max Pooling, as demonstrated in Figure \ref{fig:maxpool}. MaxPooling functions by selecting the highest activation value within each pooling window, thereby diminishing feature map size and computational demands.
CNN architectures considerably accelerate training and inference by significantly decreasing the number of learnable parameters and utilizing shared weights throughout the input image. For example, a conventional neural network with an input of a 256  \( \times \) 256 image and a hidden layer consisting of 10 neurons would need to optimize a total of:

\begin{align}  
    256 \times 256 \times 10 = 655360
\end{align}
parameters. Conversely, employing a CNN with 32 filters, each sized 3  \( \times \) 3, drastically reduces the parameter count to:
\begin{align}  
    3 \times 3 \times 32 = 288
\end{align}

As a result, CNNs efficiently perform the objectives by requiring significantly fewer parameters, thereby reducing computational load and improving performance.

\subsubsection{Typical Convolutional Neural Network Architectures}
\begin{figure}[htb]
    \centering
    \includegraphics[width=1\linewidth]{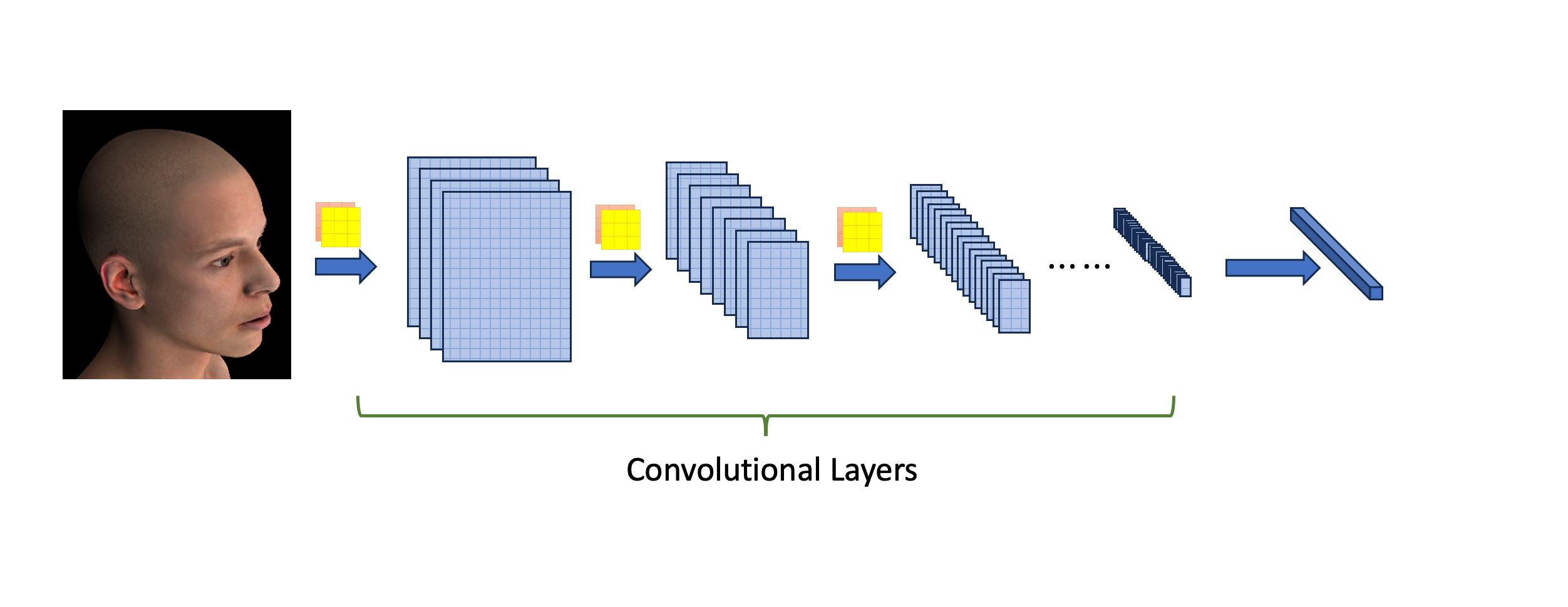}
    \caption{CNN Architecture: Input images are processed in sequential convolutional and pooling layers, progressively reducing spatial dimensions to compact, representative feature maps. These feature maps are then flattened and passed to subsequent neural network layers for further learning.}
    \label{fig:cnnarc}
\end{figure}

A typical Convolutional Neural Network (CNN) processes images through multiple convolutional layers, each employing a set of filters designed to detect specific visual patterns or features. These convolution operations are often paired with pooling layers, reducing spatial dimensions and highlighting dominant features. Several convolution and pooling layers are typically sequentially stacked, progressively distilling the input image into compact, meaningful feature representations. As shown in Figure \ref{fig:cnnarc}, an input image undergoes successive convolutions, reducing its spatial size while preserving essential feature information. The resulting small feature maps are then flattened into vectors, which are later used as input for fully connected layers or additional neural networks for further classification or regression tasks.

\begin{figure}[htb]
    \centering
    \includegraphics[width=0.5\linewidth]{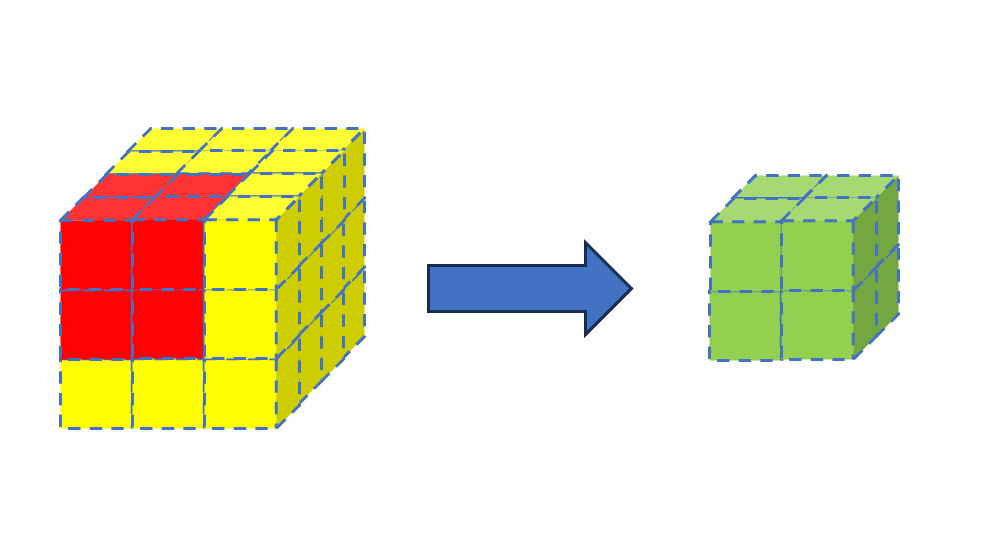}
    \caption{3D Convolutional Neural Network Operation: A three-dimensional filter, represented here as the red cube, moves across a volumetric input grid. This operation generates a new, smaller 3D volume that retains essential structural information from the original data.}
    \label{fig:3dnn}
\end{figure}

Convolutional Neural Networks (CNNs) are not limited to two-dimensional image data; they can also be used to process three-dimensional volumetric data. Instead of operating conventional 2D filters, three-dimensional convolutional filters are used to operate directly on 3D volumes. CNNs can recognize and interpret spatial patterns and structural shapes in volumetric data, extending their capabilities beyond standard 2D image analysis tasks. Figure \ref{fig:3dnn} illustrates the process of applying a 3D convolutional operation on volumetric data.

\subsection{Classic CNN Models}
In general, most CNN architectures are composed of the fundamental building blocks described in the previous sections. The primary objective of CNNs is to downsample the input images, progressively extracting meaningful features at each stage. Several pioneering CNN models have introduced groundbreaking innovations that significantly advanced deep learning. LeNet-5 \cite{lecun1998gradient} was one of the earliest successful CNN architectures, introducing foundational concepts used in document recognition tasks. AlexNet \cite{krizhevsky2012imagenet} was the first architecture to dramatically improve image classification accuracy, clearly demonstrating the superiority of deep learning methods over traditional statistical algorithms. Additionally, VGG-16 \cite{simonyan2014very} simplified network complexity by consistently utilizing small 3 \(\times\) 3 convolutional filters paired with max-pooling layers, facilitating deeper models without substantially increasing hyperparameters. Collectively, these influential architectures formed the foundation for contemporary deep-learning methodologies. 

Despite their effectiveness, these earlier CNN models encountered instability issues when attempting to increase depth, mainly due to vanishing gradient problems. ResNet \cite{he2016deep} addressed this challenge by introducing residual connections, which forward the output of the previous layers directly to the deeper layers, effectively stabilizing the training of significantly deeper networks. Combining elements from these models and stacking them strategically enables CNN architectures to surpass tasks such as object detection, semantic segmentation, and classification.

\begin{figure}[htb]
    \centering
    \includegraphics[width=0.5\linewidth]{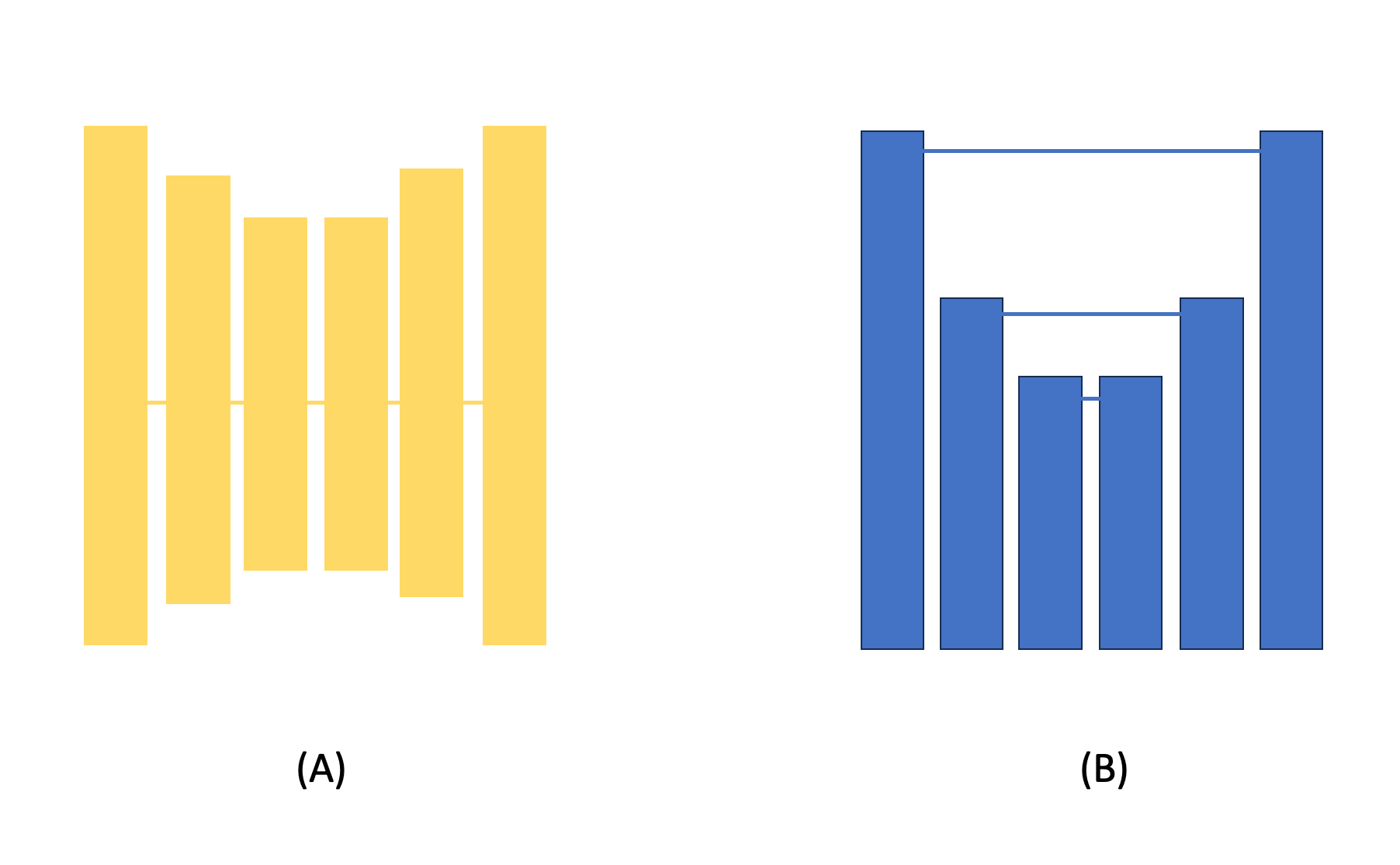}
    \caption{CNN Encoder-Decoder Architecture: (A) AutoEncoder (B) U-Net}
    \label{fig:hourglass}
\end{figure}

As noted previously, typical CNN architectures compress input images into smaller feature maps containing densely encoded information. Such feature compression is beneficial because the resulting small feature vectors efficiently represent the original images. Additionally, CNNs have the capability to reconstruct these compressed representations back to their original dimensions through upsampling. The resulting architecture resembles an hourglass shape when the compression and synthesis processes are combined sequentially. This specific CNN architecture, known as an Autoencoder, is illustrated in Figure \ref{fig:hourglass} (A). The first half of the Autoencoder compresses the input image down into a compact latent representation, while the next half decodes this representation back into the original input form. Autoencoders are particularly valuable in applications such as dimensionality reduction, denoising, and object detection tasks.

Although autoencoders can effectively denoise input data, sometimes preserving spatial details within the feature maps is crucial. To address this requirement, the U-Net architecture introduces skip connections, directly passing feature maps from the encoder stages to the corresponding decoder stages. These skip connections allow the decoder to retain fine spatial information throughout the synthesis process. Figure \ref{fig:hourglass} (B) clearly illustrates this U-Net structure with skip layers.

\subsection{Facial Alignment and Landmarking in Deep Learning}
In facial recognition and synthesis research, accurately detecting facial landmarks is a crucial step. This process involves identifying specific key points, or landmarks, on a facial image. Typically, these landmarks represent critical facial features, such as the corners of the eyes, the tip and sides of the nose, the corners and edges of the lips, and the contour of the face, including the jawline and chin.

For nearly two decades, Active Shape Models (ASM) and Active Appearance Models (AAM) have been among the most effective landmark-detection algorithms. However, recent advances in deep learning have introduced neural-network-based methods trained on extensive datasets, surpassing traditional statistical shape and appearance models in many applications.

Numerous studies have attempted to develop neural networks capable of predicting 3D facial geometry from a single 2D image. However, challenges such as variations in head pose, occlusions, image resolution, and illumination conditions significantly hinder accurate 3D predictions, particularly when employing convolutional neural networks. To address these limitations, Feng et al. proposed an end-to-end neural network approach that regresses 2D images to UV positional maps \cite{feng2018joint}. These positional maps effectively represent the full 3D facial geometry and provide dense correspondence among faces. Similarly, the 3D Face Alignment Network (3DFAN) \cite{bulat2017far} combines a 2D-to-3D face alignment module with a stacked heatmap subnetwork, predicting the depth (Z coordinate) along with 2D landmarks. Additionally, 3D Spatial Transformer Networks (3DSTN) \cite{3dstn} estimate a camera projection matrix to reconstruct 3D facial geometry and handle occluded facial regions through 2D landmark regression.

\subsection{Facial Synthesis in Deep Learning}

With the rapid development of deep learning, deep learning has also been widely used in computer vision and computer graphics to improve the accuracy of the model. In the context of facial synthesis, deep learning techniques have played a particularly significant role. One prior work is Deep Appearance Models for Face Rendering. Lombardi et al. leveraged the Activate Appearance Model (AAM) concept and proposed a method that utilizes a neural network to predict the appearance and mesh of a face \cite{Lombardi:2018}. The method in the paper aimed to disentangle the view conditions from a multi-view capturing system, by synthesizing the images of the subject into one view-independent average texture. The average texture, along with the face mesh, was then fed into an autoencoder to capture the non-linear behavior of textures, resulting in improved accuracy of the reconstructed face. However, achieving outstanding quality of human faces is a trade-off for acquiring millions of images of a single subject in a controlled environment during training. Mesh-based geometry representation often fails to capture complex dynamic scenes (e.g., fire, smoke) due to translucency, scattering, or unpredictable motions. Instead of representing the facial geometry by triangulated mesh, Lombardi et al. proposed an encoder-decoder-based neural network to represent 3D geometry by a 3D volume supervised by a set of 2D images captured from a multi-view capturing system \cite{Lombardi:2019}. However, the resulting rendering quality was constrained by the resolution of the voxel grids. The authors utilized warping functions from which the neural network decoded and the warp fields were used to index the final RGB$\alpha$ volume, resulting in a more detailed representation of the rendering result.

Traditional 3D scene representations, such as point clouds, meshes, or grids, suffer from limitations in their ability to represent complex geometries, and often require a large dataset to achieve high-quality renderings. To address these challenges,  Mildenhall et al. proposed a novel method in the paper ``NeRF: Representing Scenes as Neural Radiance Fields for View Synthesis'' for synthesizing 3D scenes in a multi-layer perceptron neural network \cite{mildenhall2020nerf}. NeRF takes 5D parameters specifying a point's location and viewing direction in 3D volume and produces RGB values and  volume density. ToFu \cite{li2020learning} and Tempeh \cite{TMPEH:CVPR:2023} adopted the inverse rendering idea, and project grids onto the images and sample features accordingly. They then estimate the 3D mesh points by employing volumetric surface-aware feature fusion. However, both approaches utilize raw multi-view stereo scans and rely on minimizing a geometric loss commonly applied in surface registration. As a result, the predicted meshes are noisy triangle meshes, requiring additional mesh wrapping steps before they can be used for rigging and production.

%% file: meshAcquisition.tex
\chapter{3D Mesh Acquisition} \label{cha:meshaacqu}
Avatars—digital representations of users in virtual environments—have gained significant popularity in interactive industries due to their ability to facilitate communication through both verbal and non-verbal behaviors, such as facial expressions and body postures\cite{gonzalez2007}. Although acquiring realistic human facial models remains complex, these techniques have found widespread applications in archaeology, forensic science, medical and healthcare research, and, most notably, 3D mesh acquisition for gaming and film production.

Despite these advances, the persistent challenge of the ``uncanny valley'' effect underscores that photorealistic human face modeling remains one of the most demanding problems in computer graphics. Recent progress in facial scanning technology, however, has provided extensive high-fidelity ground-truth data, substantially improving realism. Leveraging such data captured by advanced scanning devices, recent developments in both offline and real-time rendering have achieved notable improvements in performance and quality, enabling novel applications for digital avatars. Furthermore, the rapid evolution of deep learning has introduced greater automation in facial modeling, assisting the computer graphics community in developing more efficient algorithms and expanding their use in digital art and storytelling.

Given these advancements, a critical question arises: Can an automated facial modeling system be developed to generate animatable meshes while ensuring cross-application compatibility efficiently? Such automation is essential to handle the large volumes of data required to train sophisticated models, maintaining both scalability and adaptability across diverse use cases.

\section{History \& Traditional Pipeline for 3D Mesh}
The digitization of three-dimensional forms has played a pivotal role in visual storytelling since the earliest days of computer animation. 3D mesh acquisition, sitting at the intersection of art, engineering, and computational theory, has become fundamental to animated films, video games, and immersive virtual experiences. The field traces its roots back to Parke's pioneering work \cite{Park10.1145/800193.569955}, which introduced the first polygon-based human avatar capable of rudimentary animation. Although the mesh quality and animation fidelity were basic by today's standards, this early achievement laid the groundwork for subsequent advancements.

In 1984, \textit{The Adventures of André and Wally B.} marked one of the first explorations into animated 3D characters, showcasing the potential of computer graphics despite limited technological capabilities. Pixar Animation Studios further expanded this potential with the release of \textit{Tin Toy} (1988), featuring Billy, a fully 3D-rendered human baby character. This short film served as a crucial prototype, refining techniques that culminated in the groundbreaking feature film \textit{Toy Story} (1995)—the first fully computer-animated feature film. \textit{Toy Story} not only revolutionized the animation industry but also initiated a widespread adoption of computer-generated imagery (CGI) in film production.

Entering the 2000s, the visual effects (VFX) industry experienced a significant technological leap with the introduction of facial motion capture. Films such as \textit{The Polar Express} (2004) pioneered the use of performance-capture techniques to translate actors' nuanced facial expressions into digital characters, pushing realism to new heights. Following closely, the film \textit{Spider-Man 2} (2004) adopted the advanced Light Stage scanning technology developed at the Institute of Creative Technologies (ICT) lab, providing highly accurate facial scans of actors to enhance visual fidelity. The integration of facial scanning with performance capture became a standard approach in landmark films like \textit{The Curious Case of Benjamin Button} (2008), \textit{Avatar} (2009), and \textit{TRON: Legacy} (2010). These films demonstrated the power of combining cutting-edge scanning technologies with detailed performance capture, creating photorealistic CG characters that convincingly bridged the uncanny valley, thus setting new benchmarks in visual storytelling and digital character creation.
 
\begin{figure}[!htb]
    \centering
    \includegraphics[width=0.8\linewidth]{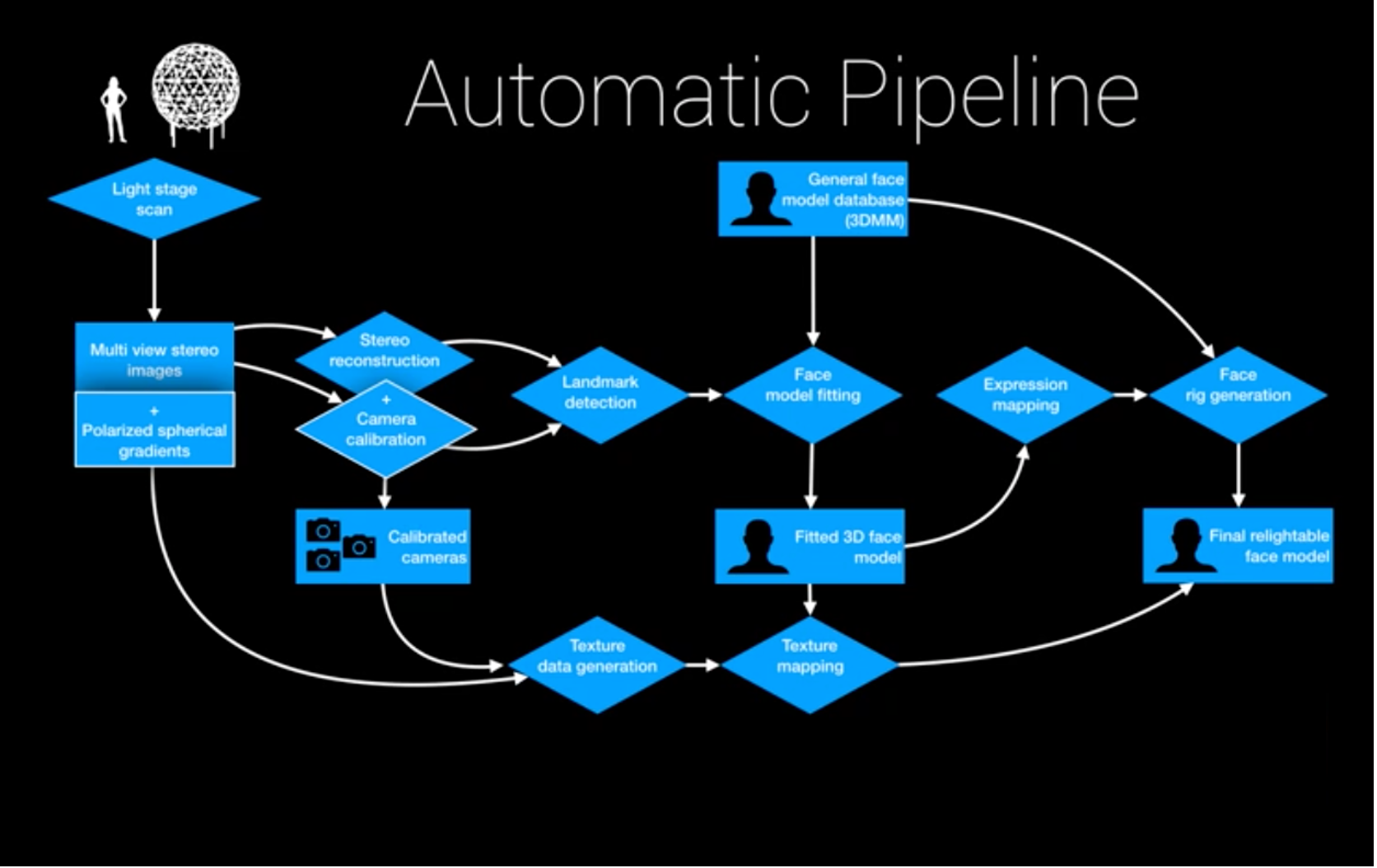}
    \caption{Digital Twin Creation Pipeline\cite{pipiline_image}}
    \label{fig:pipeline}
\end{figure}

While these advancements illustrate remarkable progress in computer graphics, the creation of photorealistic digital faces remains a challenging, multi-step task. A standard pipeline for producing lifelike facial models typically consists of several distinct phases, as depicted in Figure \ref{fig:pipeline}. These phases commonly involve facial data acquisition using advanced light stage scanning techniques, reconstructing three-dimensional geometry through stereo methods, accurately aligning and registering facial meshes, applying high-quality texture mapping to convey realistic surface details, and ultimately rendering the finalized digital character. Each step is crucial, significantly impacting the fidelity and realism of the resulting digital face.

The facial modeling process typically begins with multi-view stereo reconstruction, where a series of images captured from different perspectives are utilized to construct dense 3D point clouds representing the face geometry. To produce a mesh suitable for animation, the dense scans are processed using a retopology step, where a clean, quad-based mesh is fitted onto the original unstructured data. Importantly, this topology is designed with anatomical considerations in mind — the edges of each quad face closely follow natural facial muscles and key anatomical landmarks, thereby enabling realistic and convincing animation deformations.

\begin{figure}[!htb]
    \centering
    \includegraphics[width=0.4\linewidth]{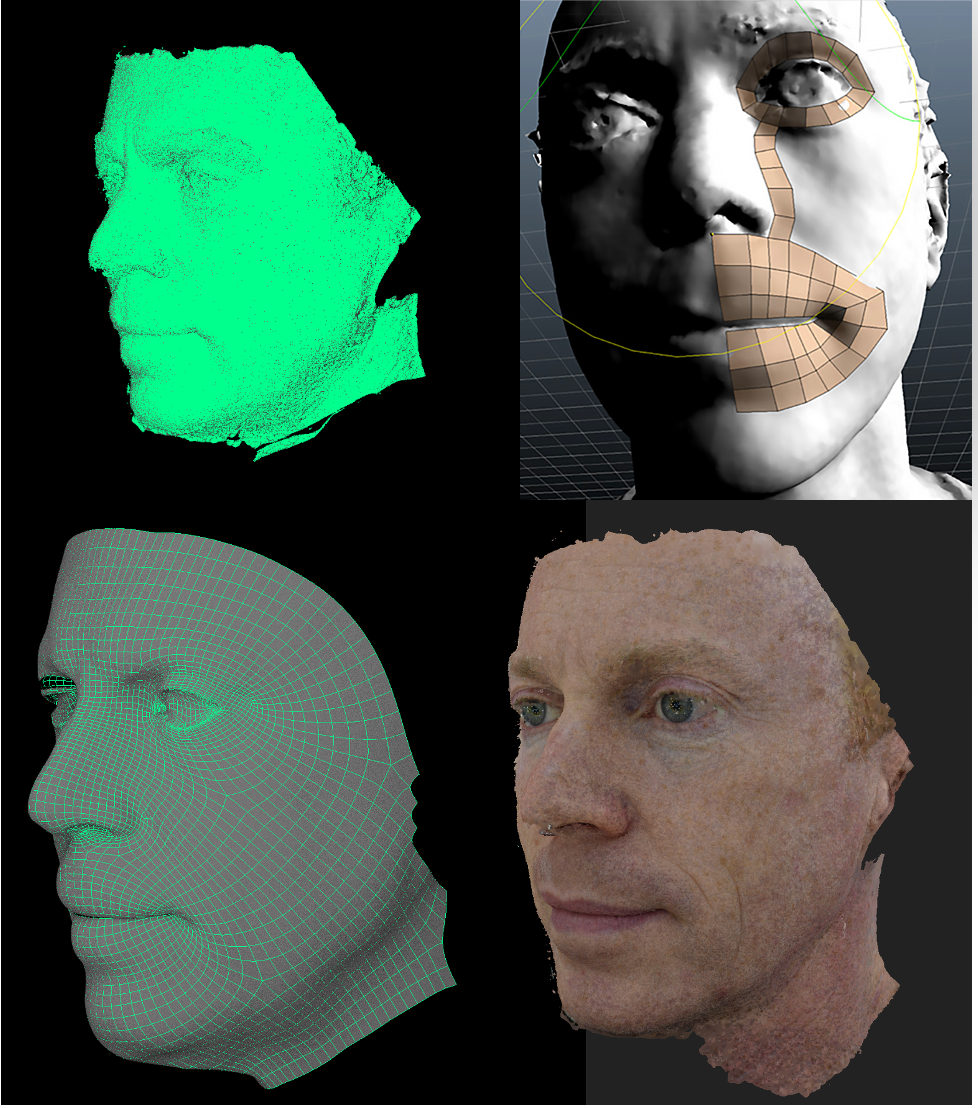}
    \caption{Example of Face Retopology}
    \label{fig:retopo}
\end{figure}

Figure \ref{fig:retopo} illustrates this vital retopology procedure clearly. The top-left panel depicts the initial dense yet unstructured scanned data. The top right panel demonstrates the careful manual placement of the quad topology that facilitates animation. Finally, the bottom panel showcases the completed, production-ready mesh with detailed texture mapping, highlighting the quality and fidelity achievable through this approach.

Despite significant technological improvements, the described facial synthesis pipeline remains highly labor-intensive. Subjects are required to remain completely still during the scanning process, ensuring accurate data capture. Hereafter, artists must invest considerable effort in manually refining the mesh topology and textures—a procedure that requires thorough attention, substantial technical proficiency, and informed artistic judgment. This process's inherently manual and intricate nature thus presents ongoing challenges in terms of time efficiency, resource allocation, and consistency of results.

\section{3D Mesh Acquisition Models}
The creation of animatable virtual-human digital twins has witnessed substantial advancements in the region of computer graphics, offering unprecedented potential to enhance human-computer interfaces across various applications. However, the development of a production-ready, re-topologized mesh of the human face remains a time-intensive undertaking, encompassing various intricate subtasks such as face detection, key points landmarking, face alignment, geometry acquisition, and retopology. The initial step in generating a human digital avatar involves accurately detecting and localizing faces within images. The Viola-Jones algorithm, renowned for its effectiveness, has been widely adopted in computer vision libraries (e.g. OpenCV) for face detection. The Viola-Jones algorithm employs a training process that leverages rectangular features, known as Haar filters, to differentiate between facial and non-facial regions within subregions of an image. This classification is achieved by evaluating a weighted sum of the Haar filters, enabling the algorithm to accurately recognize and differentiate faces from other elements present in the image. Subsequently, the process of localizing canonical landmarks typically involves face alignment to account for rotation, translation, and scale variations resulting from pose or view-direction differences. Traditionally, the Active Shape Model (ASM) \cite{COOTES199538}, a statistical method, has prominently featured in facial landmarking research, serving as a dominant approach for many years. ASM needs to pre-annotate face images based on a common landmarking schema and utilize these one-to-one corresponding landmarks for model training. Notably, ASM primarily focuses on identifying corresponding key points across all images. Building upon the foundation laid by ASM, the Active Appearance Model (AAM) \cite{cootes1998active} incorporates texture information into the model, facilitating the construction of a comprehensive 2D representation of human faces, encompassing both shape and texture attributes. This extended framework empowers a more holistic characterization of facial features, providing enhanced capabilities for analysis and synthesis. These algorithms exhibit exceptional proficiency in accurately localizing, landmarking, and reconstructing facial features within 2D images. The advancements in 2D face analysis have paved the way for significant progress in the field of 3D facial synthesis. In traditional production pipelines, the quest for achieving a convincing mesh geometry typically involves capturing images of actors using either a passive stereo system or an active lighting environment. The 2D facial analysis techniques mentioned above can also serve as a preprocessing step for containing human faces. 

Accurately reproducing a realistic 3D geometry of the human face presents significant challenges, primarily stemming from the intricate nature of facial structures. In computer graphics, the prevailing approach for achieving high-fidelity human representations is through a technique known as multiview stereo reconstruction. This methodology involves capturing multiple images of a subject from different viewpoints, followed by subsequent steps such as image rectification, feature matching, depth map acquisition, and back projecting to 3D point clouds. Beyond capturing facial geometry, the 3D point clouds are further converted into a standard polygonal mesh representation. This format typically has reduced point density, making it more suitable for rigging and animation workflows In addition to the previous challenges, the creation of a high-quality 3D facial mesh entails a complex setup and processing.  A comprehensive facial mesh suitable for production requires incorporating essential components such as 3D geometry, diffuse texture, and normal texture. Capturing these essential textures typically involves subjects positioned within a carefully controlled lighting environment. By employing a programmable active lighting environment that can change color, intensity, and direction, it becomes possible to replicate various real-world lighting conditions. This controlled setup enables the capture of an accurate reflectance field, allowing for the recreation of highly realistic human facial models.

In recent years, significant progress has been made in the field of facial analysis and synthesis, thanks to the adoption of deep learning-based neural network methods, particularly those involving deep convolutional neural networks. The deep-learning-based methods often have demonstrated outperformed statistical shape and appearance models in many areas. In most cases, those deep-learning-based neural-network methods are supervised learning, which requires an enormous amount of labeled data to train the model. Those publicly available face databases are labeled with some landmark scheme (iBUG-68 is one of the commonly used), which have been utilized to train neural networks using input facial images.

Numerous challenges and drawbacks are encountered throughout the facial synthesis pipeline. One notable issue is the laborious and time-consuming nature of creating a production-ready, re-topologized mesh of the human face, which involves many subprocesses requiring substantial manual effort. The transition from traditional statistical models to deep-learning-based neural network methods has shown promise in alleviating some of these subproceses to a certain degree. 
However, it is important to highlight that these methods heavily rely on large datasets comprising real individuals captured under controlled environments or the construction of datasets using statistical models from human face images.

\section{Issue in Facial Mesh Acquisition}

While digital human creation has advanced considerably, producing convincing animatable facial models continues to present substantial difficulties in computer graphics. The uncanny valley effect remains particularly problematic, where minor imperfections in virtual faces can trigger negative viewer reactions. To expedite the labor efforts in the traditional pipeline, recent research endeavors have steered towards integrating neural networks for facial analysis and acquisition. This paradigm shift harnesses the power of deep learning to generate high-fidelity retopologized face meshes while enhancing workflow efficiency.

Several papers \cite{johnston2018review, cceliktutan2013comparative} notice the shift of interest to deep-learning methods due to potential performance increases and techniques that also perform the 3D alignment. Given the inherent two-dimensionality of images, convolutional neural networks (CNNs) have emerged as the dominant force in recent computer vision research, including the domain of facial landmarking\cite{johnston2018review}, and has been used for facial landmarking since the early 90s \cite{vaillant1993}. Demonstrating their superiority over statistical models, deep learning-based facial alignment models have exhibited robustness in handling diverse scenarios, encompassing challenges such as head rotation, translation, and occlusions\cite{johnston2018review, xiang2021, cceliktutan2013comparative}. Efforts in research have been dedicated to overcoming the adverse effects of rotation, translation, and occlusions by employing strategies that either estimate head poses \cite{hu2018pose, burgos2013robust, wang2020} or infer the camera projection matrix\cite{bulat2017far}. The robustness of deep learning methods compared to statistical approaches is further strengthened by the availability of numerous publicly accessible databases for training neural networks, such as 300W\cite{300w}, COFW\cite{cofw}, WFLW\cite{wflw}, and AFLW \cite{aflw}. These datasets encompass a wide range of facial attributes, including age, ethnicity, skin color, expression, and pose variations. The utilization of such comprehensive databases contributes to the overall robustness of deep learning methods.

Moreover, the generation of a production-ready retopologized mesh for human faces has traditionally entailed substantial manual labor within the production pipeline. However, recent strides in deep learning techniques offer a promising avenue for end-to-end learning, eliminating the need for intermediate subprocesses. This innovative approach enables the direct conversion of human face photographs into cohesive 3D retopologized meshes with visually compelling appearances. By harnessing the power of deep neural networks, this streamlined workflow not only streamlines the production pipeline but also enhances the efficiency and realism of facial synthesis. Such breakthroughs have the potential to revolutionize the field of computer graphics, paving the way for accelerated advancements in facial analysis and synthesis. One widely adopted deep learning architecture for facial synthesis is the Variational Autoencoder (VAE). In the VAE framework, the encoder component compresses input images into a latent vector representation. In the decoding stage, researchers have diverged into two parts. A primary focus has been decoding the latent vector into a volumetric representation \cite{Lombardi:2019}, which is subsequently subjected to volumetric rendering techniques to generate re-rendered images. While this approach yields visually realistic results, it has inherent limitations regarding its suitability for production usage. An alternative approach directly decodes the latent space into a retargeted common mesh \cite{Lombardi:2018, zhang2022, dias2022, moser:semi}. This retopologized mesh empowers artists with greater flexibility in customizing facial geometry for production purposes. Similar to the concept of end-to-end learning, deep learning models have emerged as powerful tools for acquiring high-level feature representations of faces and facial geometry. In contrast to traditional methods that involve a series of steps for geometry and texture acquisition, deep learning offers a more straightforward process. The widely adopted technique for geometry acquisition is multi-view stereo, which entails capturing multiple images of the subject from different viewpoints\cite{beeler2010high}. These images are then processed using feature-matching algorithms to calculate depth maps, enabling the derivation of accurate facial geometry. This methodology is widely adopted by commercial photogrammetry software, such as Agisoft Metashape, to achieve accurate and detailed results. Other high-level featurs of faces, such as diffuse, normal and specular maps, can be obtained by polarized gradient illumination\cite{ma2007rapid, ghosh2011multiview}. However, end-to-end deep learning makes learning geometry and texture maps all at once. Zhang et al. applied a multi-VAE architecture, which allows for the comprehensive learning of diffuse, specular, and normal maps in a unified manner.

Deep learning models in facial analysis and synthesis have enabled real-time performance capabilities. The architecture of deep neural networks facilitates the inference process because the inference is a linear combination of weights and inputs. The computational power of GPUs allows for parallel processing, making predictions within milliseconds \cite{serra2022, Lombardi:2018}.  In contrast, traditional statistical approaches typically take the order of seconds per image for facial analysis and synthesis tasks\cite{cceliktutan2013comparative}. This optimized and efficient computation enables real-time feedback from the model, making it possible to create a dynamic digital avatar with various facial expressions using live stream video of an individual.\cite{zhang2022}.

In summary, facial analysis and synthesis have found extensive applications across various domains. The generation of high-fidelity facial avatars serves as a valuable tool for producing lifelike human animations in the realms of movies, games, and real-time avatars. Furthermore, these techniques have proven instrumental in efficiently creating humanoid creatures and enhancing rendering quality, thereby contributing to improved visual realism in computer graphics. The integration of facial analysis and construction methodologies in deep learning has thus opened new avenues for realistic visual representation and immersive experiences in diverse fields.


%
%

%% file: acquisitionSystem.tex
\chapter{Facial Data Acquisition Systems}
As previously discussed in Chapter \ref{cha:meshaacqu}, constructing a realistic facial avatar involves numerous technical and hardware challenges. Creating a high-quality digital face remains a complex and labor-intensive process, comprising specialized tasks such as face detection, landmark localization, geometry acquisition, alignment, establishing surface correspondences, capturing reflectance properties, and mesh retopology—each stage frequently requiring significant human artistic input. Achieving high-fidelity facial data further demands precise control over lighting, geometry, and material properties, typically necessitating custom-built capture setups and advanced computational processing methods. To tackle these challenges, we have developed VarIS, a versatile Light Stage system designed to efficiently acquire detailed facial geometry and accurate reflectance information under controlled illumination conditions. In parallel, we employ 3D Morphable Models (3DMMs) to convert captured raw data into highly accurate, production-ready, and animatable facial meshes applicable to diverse use-cases. This chapter introduces these tools and methodologies, showcasing our lab's approach to addressing and overcoming traditional barriers in facial data acquisition and modeling.

\section{Variable Illumination Sphere}

Historically, various 3D scanning technologies have played an important role in how facial geometry is captured and turned into usable models. Cyberware scanners, prevalent from the 1980s through the early 2000s, were extensively utilized to create early 3D face datasets, notably the BU-3DFE \cite{bu3dfe} and FaceWarehouse \cite{facewharehouse}. However, these scanners operated by moving a laser line across the subject or by rotating around the person, resulting in lengthy acquisition times and limited control over lighting and exposure. Consequently, the output meshes generally exhibited low-resolution textures and inadequate surface detail. To address these limitations, the 3DMD system introduced synchronized multi-camera arrays, capturing facial geometry through rapid projection of infrared or color patterns onto the subject. This approach significantly accelerated the capture process and enabled higher-resolution texture acquisition, producing detailed 3D point clouds more efficiently.

Since the early 2000s, the Light Stage developed by Debevec et al. \cite{debevec2000acquiring} has emerged as the state-of-the-art device in visual-effects production and research. The distinctive advantage of the Light Stage lies in its capability to capture not only accurate 3D geometry but also reflectance properties such as diffuse, specular, and surface normals. This is achieved through spherical gradient illumination combined with polarization techniques, offering comprehensive ground-truth data crucial for highly realistic rendering. Moreover, the availability of such detailed reflectance and geometry data has revolutionized how actors' performances are digitally captured and translated into animation within the film industry.

Nevertheless, substantial challenges persist throughout the pipeline of creating facial models. These include accurately scanning and registering facial geometry, retopologizing meshes suitable for animation, capturing physically based reflectance properties, and effectively rigging and animating the resulting avatars. To overcome these issues, we designed VarIS, the Variable Illumination Sphere, a versatile multi-purpose system specifically engineered for acquiring and processing high-fidelity geometric and appearance data for advanced computer graphics research and production workflows.
\subsection{Hardware Setup}
\begin{figure}[!htb]
    \centering
    \includegraphics[width=0.5\linewidth]{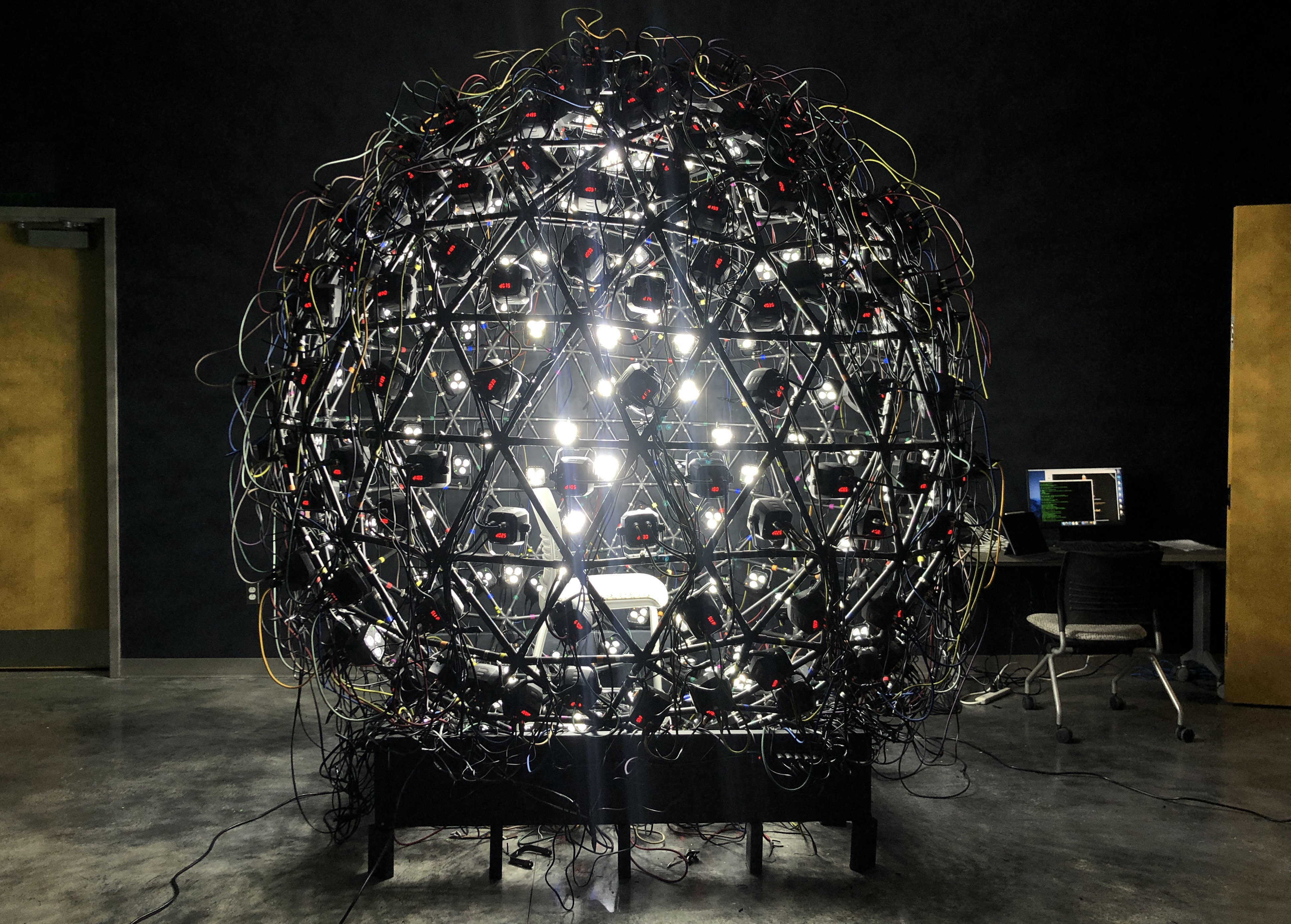}
    \caption{VarIS with all white lights activated.}
    \label{fig:varis}
\end{figure}
The physical support structure for VarIS consists of an 8-foot diameter 4V geodesic dome, precision-designed and fabricated by Sonostar Universal Structures \cite{sonostar}. The higher-frequency geodesic configuration was selected to provide both structural robustness and near-perfect spherical geometry for equipment mounting, while offering substantial cost savings over traditional custom metal frameworks. The system incorporates 364 modular lighting units, each containing three RGBW (red, green, blue, white) LED arrays with individual Fresnel lens collimation. These units are precisely mounted along optimized latitudinal and longitudinal arcs that conform to the geodesic geometry, facilitating efficient switching between dual polarization states.

For system control, we implemented a distributed architecture comprising six primary serial chains (36 lighting units each) and four supplementary chains for polarized pole configurations, totaling the complete 364-unit array. The imaging subsystem utilizes a hardware-agnostic approach, processing native RAW formats through the open-source gPhoto2 library \cite{gPhoto2} rather than proprietary SDKs, thereby ensuring greater flexibility and future compatibility. While Nikon NEF serves as the primary capture format, the system maintains full interoperability with Canon CR2 files.

Lighting control employs the ESTA DMX512A standard \cite{dmx}, chosen for its proven reliability in large-scale installations, extensive device support, and cost-efficient implementation through standardized components. This configuration provides precise, synchronized control of all lighting elements while minimizing custom electronic requirements. The complete VarIS geodesic sphere is illustrated in Figure \ref{fig:varis}.

\subsection{Software \& Application}
\begin{figure}[!ht]
    \centering
    \includegraphics[width=0.5\linewidth]{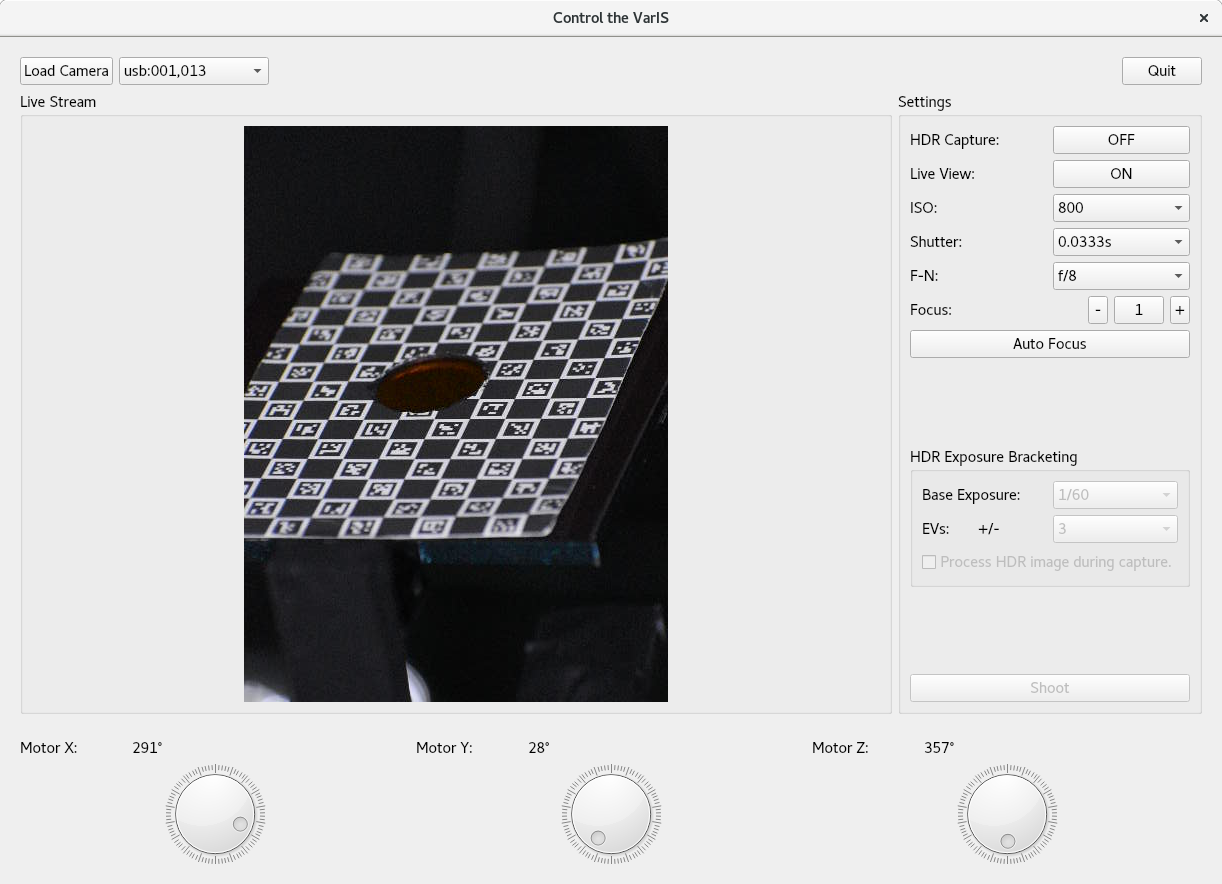}
    \caption{Graphical User Interface for Camera Control}
    \label{fig:qt}
\end{figure}

The VarIS software system is primarily developed in Python, chosen for its rapid development capabilities, cross-platform compatibility, and extensive support through various open-source libraries. Core libraries employed include built-in Python modules for file system operations, pySerial \cite{Liechti2020-pySerial} and gPhoto2 \cite{gPhoto2} for hardware communication, Qt5 \cite{Qt}, PyQt5 \cite{PyQt}, and OpenGL \cite{OpenGL} for user interfaces and visualization tasks, and NumPy \cite{NumPy}, SciPy \cite{SciPy}, and OpenCV \cite{OpenCV} for data processing and analysis. The lighting control subsystem utilizes pySerial to manage serial communication with the ten DMX lighting chains, facilitating various illumination patterns including axis gradients and one-light-at-a-time (OLAT), and handling spatial coordinates for each lighting element.

A graphical user interface (GUI), illustrated in Figure \ref{fig:qt}, was developed using PyQt5, gPhoto2, and OpenCV. This interface provides real-time camera views to assist users in fine-tuning camera focus, adjusting exposure settings, positioning subjects within the capture volume, and aligning material samples to specific viewpoints.
\begin{figure}[!ht]
\centering
\includegraphics[width=0.5\linewidth]{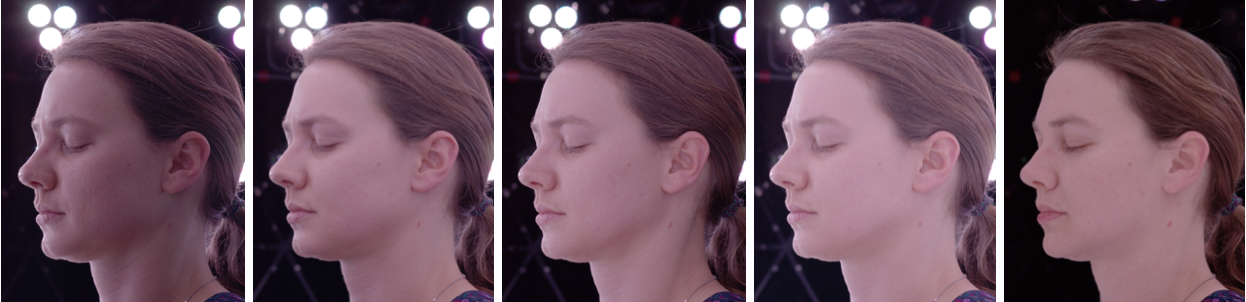}
\caption{Example of Gradient Illumination within VarIS}
\label{fig:gradient}
\end{figure}
For efficient and detailed texture acquisition, gradient illumination and polarization strategies \cite{ma2007rapid} were implemented. This process involves capturing four sequential photographs under distinct illumination conditions: gradient patterns along the X-, Y-, and Z-axes, followed by uniform cross-polarized illumination. Figure \ref{fig:gradient} shows sample photographs demonstrating these different lighting conditions. From these images, normal, specular, and diffuse texture maps are derived. Processed textures and resultant renders are presented in Figure \ref{fig:maps}.

\section{Visage Craft for Synthetic Data}
The 3D Morphable Model (3DMM) framework offers a robust parametric approach to statistically modeling variations in human facial geometry and appearance. Using principal component analysis (PCA) in data sets composed of densely aligned 3D facial scans, 3DMMs successfully encapsulate the natural variability found in human facial morphology. This capability enables two essential functionalities: (1) synthesizing novel facial shapes and textures through low-dimensional parameter adjustments, and (2) reconstructing precise 3D facial geometry from limited or partial data, thus simplifying complex facial representation tasks \cite{3d-morphable-models}.

Several prominent examples of publicly available 3DMMs have significantly impacted facial modeling research. The Basel Face Model (BFM), built from over 200 high-quality Cyberware scans, is one of the earliest and most widely adopted models, encompassing both shape and texture variations \cite{bfm09, bfm2017}. The Liverpool-York Head Model (LYHM), derived from a more diverse dataset comprising over 1,200 head scans, enhances demographic representativeness and coverage \cite{LYHM}. Furthermore, FLAME \cite{FLAME:SiggraphAsia2017}, extensively utilized in academic research and deep learning pipelines, integrates identity, expression, and pose blend shapes, making it particularly suited for expressive face animation and avatar generation.

Although these publicly accessible 3DMM datasets have facilitated considerable advances, they sometimes fall short in terms of the resolution or flexibility needed for specific, specialized applications. To address these limitations and to better fulfill our research and production objectives, we have developed a high-resolution Appearance 3D Morphable Model explicitly tailored to our project's requirements.

\subsection{Model Building Preparation}

To construct our Appearance 3D Morphable Model (A3DMM), we started with 48 face meshes obtained from the 3D Scan Store, each sharing the same topology and accompanied by corresponding diffuse, normal, and gloss maps. Initially, Procrustes Analysis (without scaling) was performed on these meshes to determine the optimal rigid transformations for precise alignment. Principal Component Analysis (PCA) was then separately applied to both the aligned mesh geometries and the associated texture maps. From each PCA result, eigenvectors and eigenvalues were extracted, with the eigenvectors sorted in descending order based on their corresponding eigenvalues, determining the primary directions of variation within the constructed model.

Most existing 3D Morphable Models (3DMMs) do not typically integrate Physically Based Rendering (PBR) techniques into their workflows. To enhance realism in face mesh rendering, we have adopted a microfacet-based bidirectional reflection distribution function (BRDF) model.

The full BRDF rendering equation is expressed as:
\begin{align}  
   L_o(x,\omega_o, \lambda,t) &= L_e(x,\omega_o, \lambda,t)+\int_{\Omega}f_r(x,\omega_i,\omega_o, \lambda,t)L_i(x,\omega_i,\lambda,t)(\omega_i\cdot n)d\omega_i
\end{align}

where $x$ is the surface point, $\omega_o$ and $\omega_i$ represent outgoing and incoming light directions respectively, $\lambda$ denotes wavelength, and $t$ denotes time. This equation calculates radiance at a surface by integrating incoming radiance over the hemisphere, factoring in the surface's reflective characteristics.

In typical computer graphics scenarios, wavelength and time dependencies are usually neglected, simplifying the equation to:

\begin{align}  
   L_o(x,\omega_o) &= L_e(x,\omega_o)+\int_{\Omega}f_r(x,\omega_i,\omega_o)L_i(x,\omega_i)(\omega_i\cdot n)d\omega_i
\end{align}

The BRDF $f_r(x,w_i,w_o)$ combines diffuse and specular components:
\begin{align}  
   BRDF &= k_df_{diffuse} + k_sf_{specular}\\
   k_d + k_s &= 1
\end{align}
By accounting for Fresnel reflectance, we simplify $k_s$ allowing $k_d$ to be computed as $k_d = 1-k_s$ 

For diffuse reflection, we use the Lambertian model:
\begin{align}  
   f_{diffuse} = f_{Lambert} = \frac{color}{\pi}
\end{align}

The Cook-Torrance microfacet model \cite{cook-torrance} represents specular reflection and is given by:

\begin{align}  
   f_{specular} = f_{Cook-Torrance} = \frac{D(h)G(\omega_i,\omega_o,h)F(\omega_o,h)}{4(\omega_i\cdot n)(\omega_o\cdot n)}
\end{align}

Here, $F(\omega_o,h)$ represents the Fresnel function, which models view-dependent reflectance:
\begin{align}  
   &F=F_{schlick}=F_0+(1-F_0)(1-(\omega_o\cdot h))^5
\end{align}
where $h$ is the halfway vector computes as $h = \frac{\omega_i+\omega_o}{||\omega_i+\omega_o||}$, and $F_0$ is the reflectance at normal incidence (typically around 0.04 for human skin). Since the $F$ counts for the Fresnel term, which simplifies the BRDF to: 
\begin{align}  
   & BRDF = k_df_{diffuse} + f_{specular}\\
   & k_d = 1 - k_s, k_s=F
\end{align}

The normal distribution function $D(h)$ describes microfacet orientation distributions on the surface. We utilize the Trowbridge-Reitz normal distribution \cite{walter2007microfacet}:
\begin{align}  
   &D(h) = D_{Trowbridge-Reitz}(h) = \frac{\alpha^2}{\pi((n\cdot h)(\alpha^2-1)+1)^2}
\end{align}
where $\alpha$ represents the squared roughness ($\alpha=roughness^2$).
Finaly, the geometry shadowing function $G(\omega_i,\omega_o,h)$ accounts for microfacet shadowing and masking effects and is defined by the Schlick-Beckmann approximation:
\begin{align}  
   G(\omega_i,\omega_o,h) &= G_{Schlick-Beckmann}(\omega_i,n) G_{Schlick-Beckmann}(\omega_o,n)\\
   &\Rightarrow G_{Schlick-Beckmann}(\omega, n)=\frac{n\cdot \omega}{{(n\cdot \omega)(1-k)+k}},k=\frac{\alpha}{2}
\end{align}

\begin{align}  
   \sigma_i = \sqrt{\lambda_i
   }
\end{align}

\subsection{Visage Craft}
\begin{figure}[ht]
    \centering
    \includegraphics[width=0.5\linewidth]{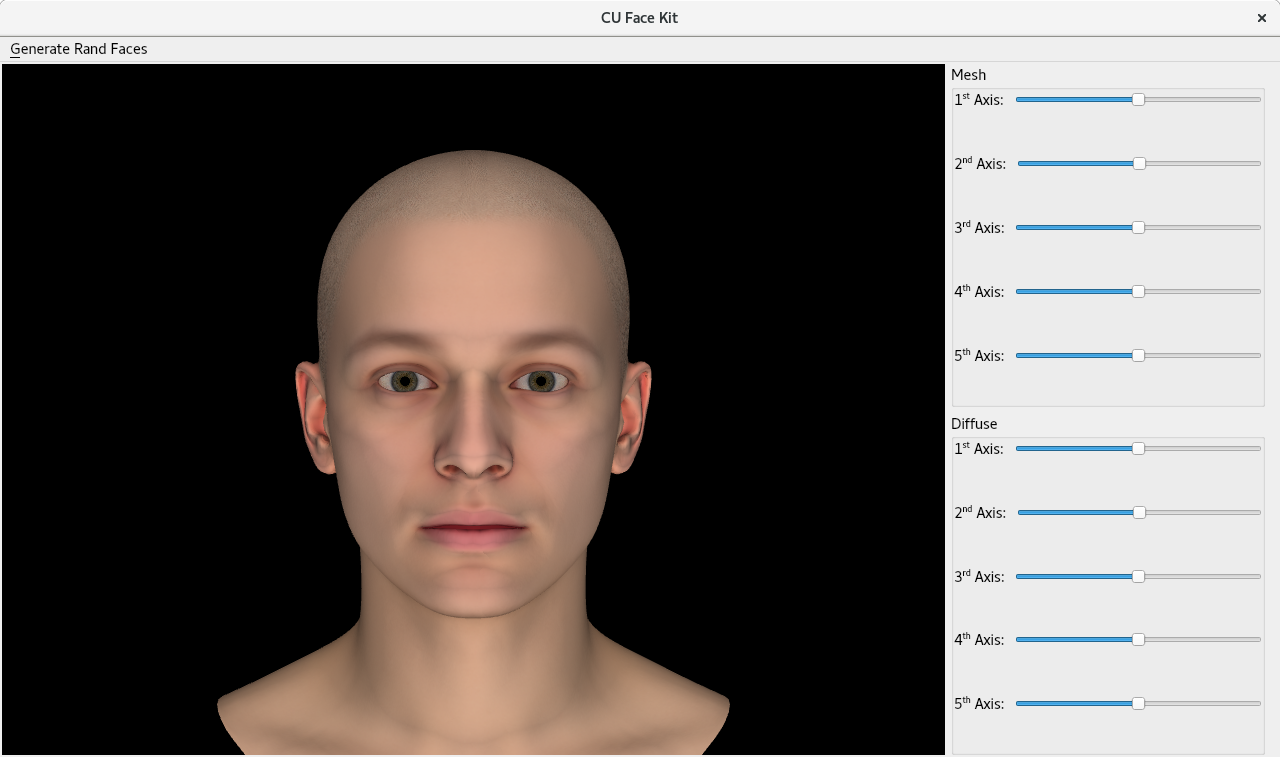}
    \caption{Visage Craft GUI}
    \label{fig:visage}
\end{figure}

Visage Craft was primarily developed using Python, relying on a variety of core libraries. It includes native Python libraries for file operations and utilizes Qt5 \cite{Qt}, PyQt5 \cite{PyQt}, and OpenGL \cite{OpenGL} for handling user interfaces and visualization operations. NumPy \cite{NumPy}, SciPy \cite{SciPy}, and OpenCV \cite{OpenCV} are used for mesh processing and PCA. We also incorporated the Cook-Torrance micro-facet Bidirectional Reflectance Distribution Function (BRDF) in the OpenGL fragment shader directly. The key elements of this BRDF implementation include the Trowbridge-Reitz (GGX) normal distribution function, the Schlick-Beckmann geometry-shadowing function, and Schlick's approximation for the Fresnel term. Each of these components uses associated maps to control parameters such as albedo, gloss, and normal direction \cite{walter2007microfacet}. A screenshot of the Visage Craft interface is provided in Figure \ref{fig:visage}.

\begin{figure}[ht]
    \centering
    \includegraphics[width=0.7\linewidth]{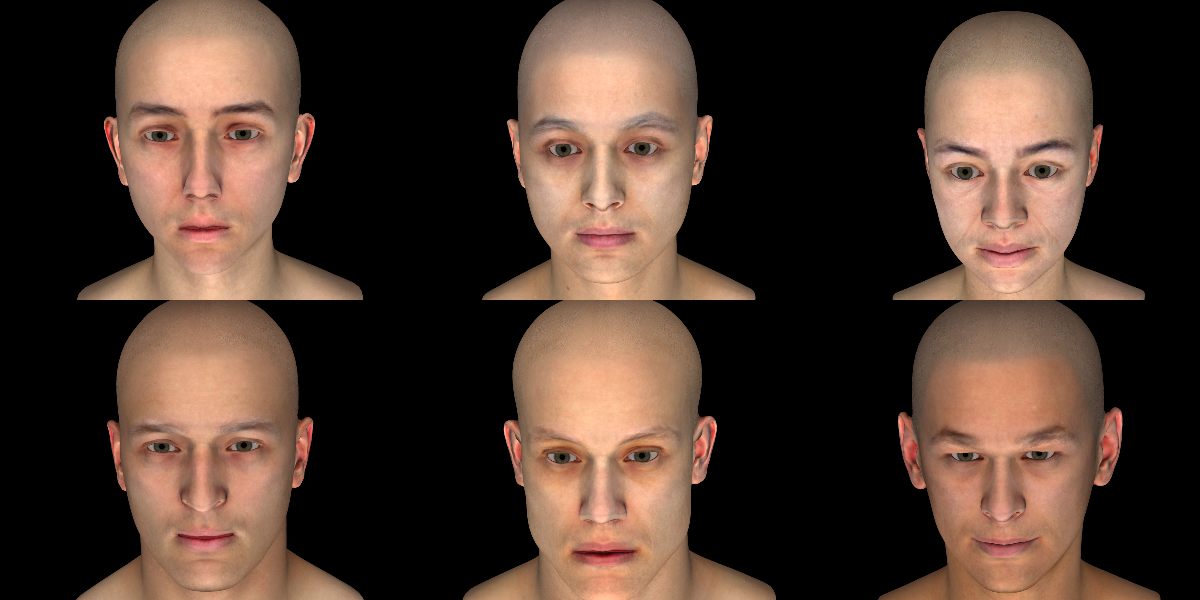}
    \caption{Sample Faces Generated using Visage Craft}
    \label{fig:vcexample}
\end{figure}

In the graphical user interface, sliders correspond to the top five principal components of the PCA models for both mesh geometry and diffuse textures. Adjusting these sliders allows users to visually observe the impact each component has on facial shape and texture appearance. Since the standard deviation associated with each principal component is the square root of its eigenvalue ($\sigma_i = \sqrt{\lambda_i}$), varying the sliders within $\pm3\sqrt{\lambda_i}$ covers approximately 98\% of the variance found in the training dataset. Examples of faces generated through Visage Craft are depicted in Figure \ref{fig:vcexample}.

\begin{figure}[ht]
    \centering
    \includegraphics[width=0.7\linewidth]{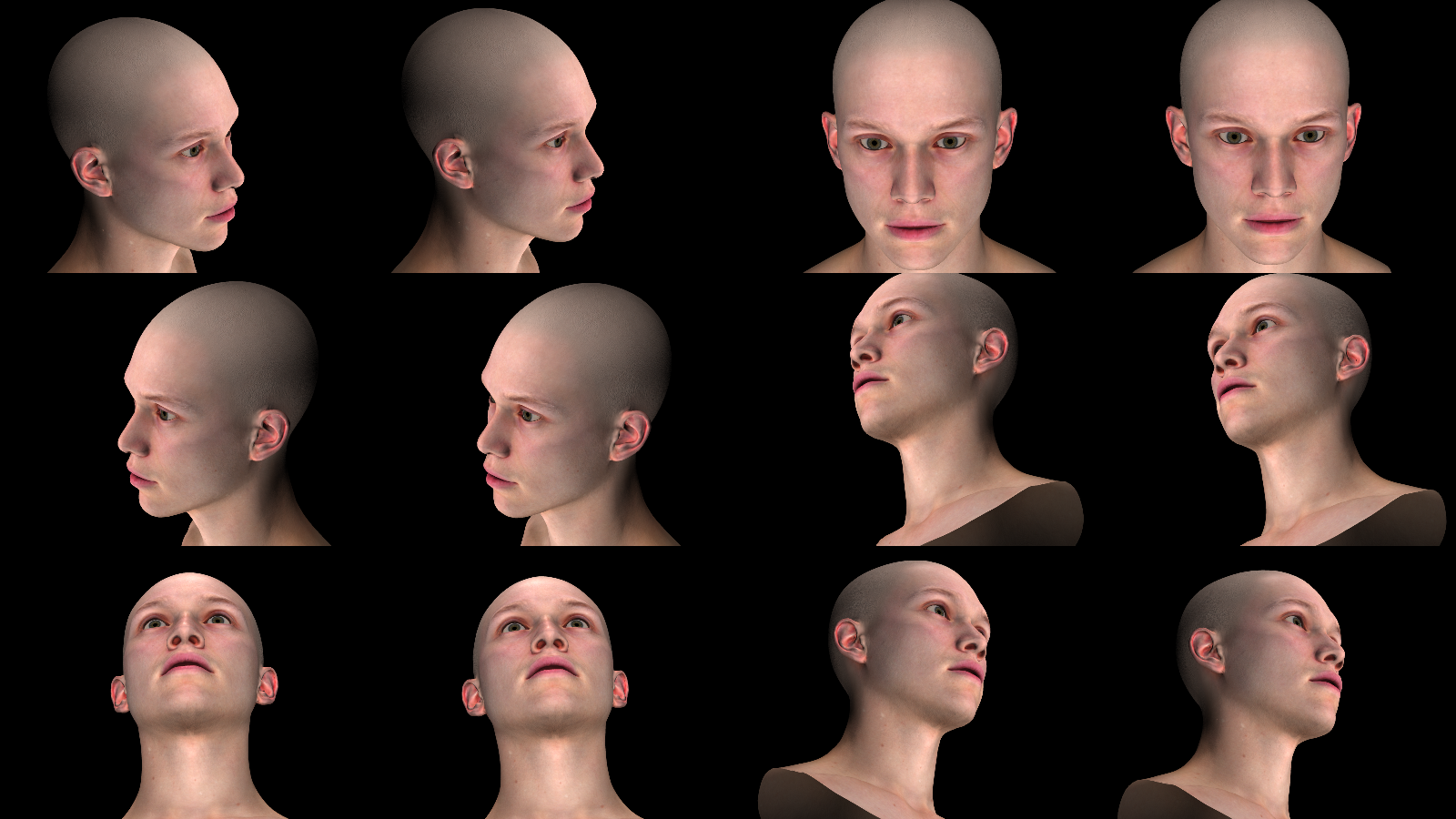}
    \caption{Multiview face samples generated by Visage Craft, showing six stereo pairs with known camera intrinsic and extrinsic parameters}
    \label{fig:multiviewSamples}
\end{figure}

Moreover, as previously mentioned, most publicly available 3D Morphable Models (3DMM) typically include variations only in facial shape and diffuse textures. This limitation makes adapting them for specialized research challenging. To address this, our mesh model is designed to be adaptable, allowing customization to suit specific research needs. Recently, significant attention has been given to multiview stereo reconstruction techniques in both traditional computer graphics and deep-learning contexts. Therefore, our system supports the generation of synthetic multiview camera setups along with facial meshes. A major advantage of having our proprietary 3DMM is the flexibility to position virtual cameras anywhere, with precisely known intrinsic and extrinsic camera parameters available for downstream applications. Figure \ref{fig:multiviewSamples} demonstrates synthetic multiview face images generated by Visage Craft, depicting six stereo image pairs along with their camera parameters.

\begin{figure}[ht]
    \centering
    \includegraphics[width=0.7\linewidth]{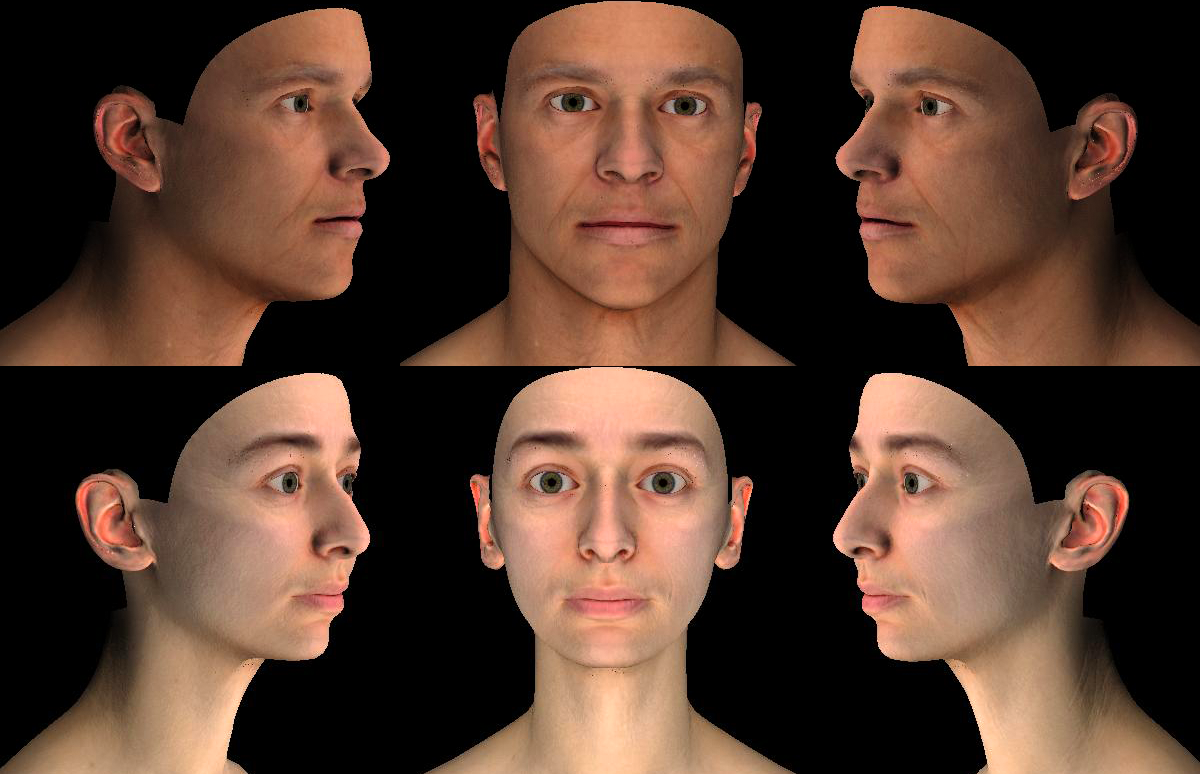}
    \caption{Multiview Samples}
    \label{fig:facesample}
\end{figure}
Additionally, because all generated meshes share a consistent topology, geometric correspondences between meshes are inherently established. These correspondences greatly facilitate subsequent tasks, such as automatic facial landmark extraction or isolating and masking specific facial regions. Figure \ref{fig:facesample} provides examples of synthetic training faces with hair regions automatically masked.

In summary, the Visage Craft Appearance 3D Morphable Model is constructed using 48 high-quality facial scans. To enhance rendering realism, we adopted the physically-based Cook-Torrance shading model. Visage Craft is capable of synthesizing random face samples within the standard deviation range, and the software is flexible enough to be customized for various research applications. This includes configuring camera and lighting setups with known parameters and utilizing inherent one-to-one mesh correspondences, thus significantly benefiting tasks like facial registration and landmark localization.

%% file: camera.tex
\chapter{Importance of Camera Intrinsic and Extrinsic Matrices}
Recent attention to facial alignment and landmark detection methods, particularly with
application of deep convolutional neural networks, have yielded notable
improvements. Neither these neural-network nor more traditional methods, though,
have been tested directly regarding performance differences due to camera-lens focal
length nor camera viewing angle of subjects systematically across the viewing
hemisphere. This work uses photo-realistic, synthesized facial images with varying
parameters and corresponding ground-truth landmarks to enable comparison of
alignment and landmark detection techniques relative to general performance,
performance across focal length, and performance across viewing angle. Recently
published high-performing methods along with traditional techniques are compared
in regards to these aspects.
\section{Introduction}
Face detection, tracking, and recognition continue to be employed in a variety of ever
more common-place biometric applications, particularly with recent integrations in
mobile-device security and communication. Most of these applications, such as identity
verification, pose tracking, expression analysis, and age or gender estimation, make use
of landmark points around facial components. Correctly locating these key points is crucial
as they often are used to abstract main features such as the jaw, eye-brows, eyes,
nose shape, nostrils, and mouth \cite{Wu_2018}. Due to the complexity of head gestures , automatic
localizing of canonical landmarks usually first involves face alignment to account for rotation,
translation, and scale due to pose or view-direction differences \cite{Burgos-Artizzu_2013_ICCV, SHI2006117, kae2014incorporating}. Furthermore,
2D images photographically captured by cameras are affected by perspective and lens
distortion, an important aspect considered in this work.

This review aims to compare performance of five notable facial landmark and alignment methods under the effects of different camera focal lengths and positions, particularly under conditions that have been  ignored or difficult to test. Previously, {\c{C}}eliktutan et al. completed a thorough survey of facial  landmark detection algorithms and comparative performance in 2013, which at the time primarily focused on 2D  techniques such as Active Shape Model (ASM) and Active Appearance Model (AAM) variations\cite{Celiktutan2013}.  In 2018, Johnston and Chazal published work that built on the earlier survey, noting the shift of interest to deep-learning methods due to potential performance increases as well as techniques that also perform 3D alignment \cite{Johnston2018}.  Several strong-performing neural-network methods have been published since, however, and in general, no performance comparisons have included lens-perspective effects nor systematic evaluation across the range of viewing angles.  This study is not an exhaustive survey of recent methods but rather an investigation in the effects of focal-length and viewing angle on both traditional and more recent neural methods (published after the 2018 article).  Focal-length-based perspective and viewing-angle are both important considerations if designing a biometric or other system in order to account for the lens chosen, viewing angle, and proximity necessary for the system.

The effects a lens imparts on acquisition have often been ignored in face-related research. A fundamental technique in computer vision is estimating a camera-projection matrix and has been regarded in many studies; however, the datasets used to train and test landmark detection do not usually include camera meta-data (particularly large datasets gleaned from the Internet for deep-learning approaches), or datasets have been captured in very controlled situations with a single lens. The most widely used databases in training recent deep networks are 300W\cite{sagonas2016300}, COFW\cite{SHI2006117}, WFLW\cite{wayne2018lab}, and AFLW\cite{aflw}. Those cover large variation over age, ethnicity, skin color, expression, and pose and have been used by top-performing deep neural networks [\cite{Wang_2019}, \cite{Valle_2018_ECCV}, \cite{su_2019}, \cite{Kowalski_2017},\cite{Lv_2017}]. None of them explicitly note focal-length as a parameter. In short, there is no data-set published online that has considered focal-length/field-of-view versus proximity for training alignment or landmark detection methods. We assume perspective distortions caused by focal length will likely affect the final annotation results. If so, training sets including camera and lens parameters could increase accuracy of a system or at least aid in designing systems.

\section{Method of Comparison}
\subsection{Landmark Schemes}
There have been a variety of landmark schemes used in related projects, but a few have been most used in recent work and make a logical choice for comparative evaluation. Following the categories in \cite{Celiktutan2013}, there are two major groups of facial landmarks schemes: primary landmarks and secondary landmarks. Primary landmarks usually define the eye corners, the mouth corners, and the nose tip. Those landmarks are located at 'T' sections between boundaries or at high curvatures on a face which may be detected by image processing algorithms, e.g. multi-resolution shape models\cite{Burt1984}, Harris Corner Detection model \cite{harris1988combined}, or Image Gradient Orientation (IGO) model\cite{tzimiropoulos2011robust}. Secondary landmarks outline the contour of main features that are guided by primary landmarks, such as the jaw line, eyebrows, and nostrils. Wu et al. \cite{Wu_2018} provide a thorough survey on facial landmark databases and their corresponding landmark schemes. A common 68-point landmark is supported by many face databases, e.g. AFLW\cite{koestinger11a}, BU-4DFE\cite{zhang2013high}, Helen\cite{le2012interactive, sagonas2016300} etc. For easiest consistency, the 68-point scheme from Multi-PIE \cite{gross2010multi}, and further popularized by iBUG's 300W \cite{sagonas2016300}, was chosen for this study.

\begin{figure}[htb]
    \centering
    \includegraphics[width=0.5\linewidth]{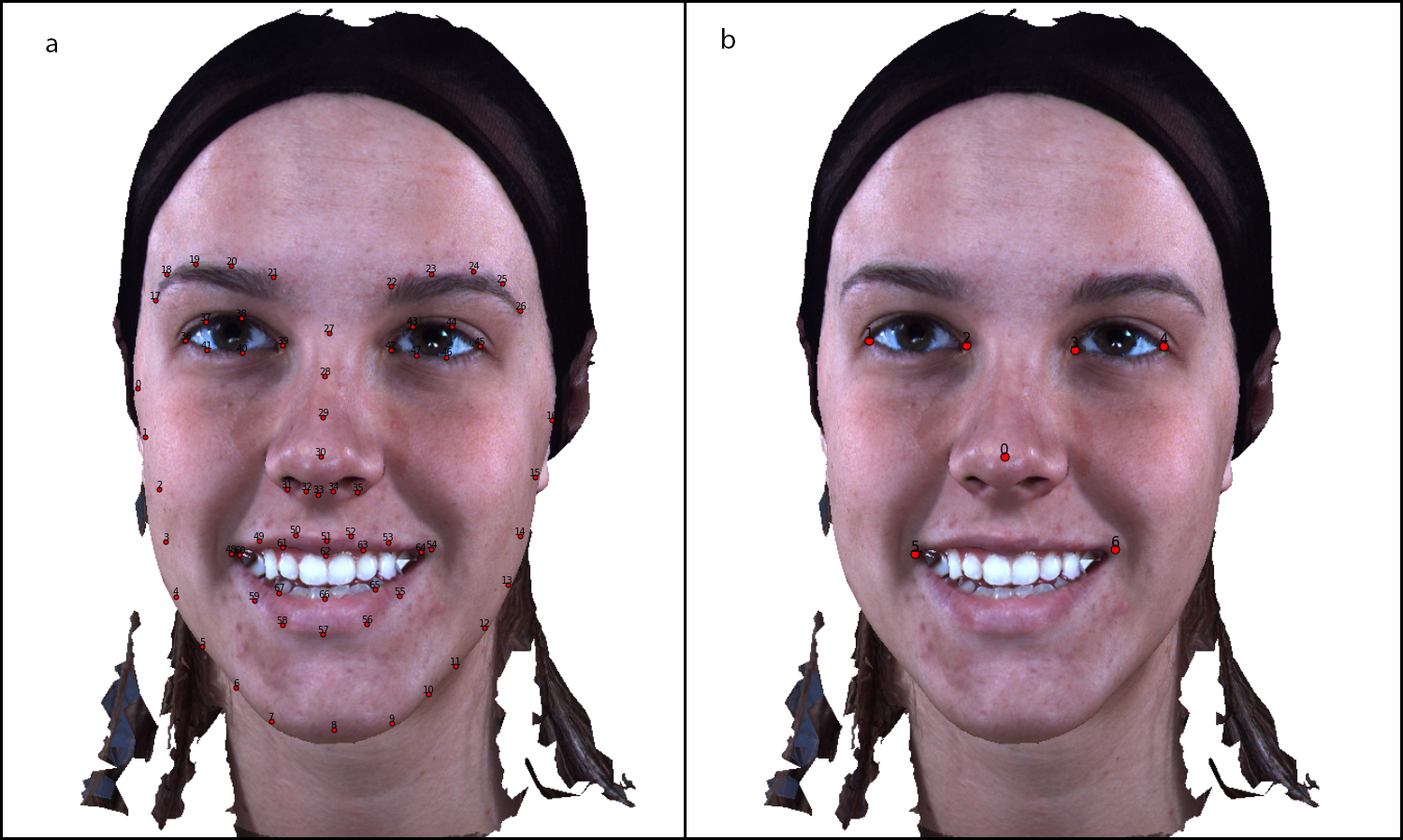}
    \caption{Two different landmarks schemes: (a) iBUG-68 (b)M7}
    \label{fig:lmkschemes}
\end{figure}

Sagonas et al. and Johnston1 et al. \cite{sagonas2016300, Johnston2018} state that primary landmarks are more easily detected than secondary landmarks while annotating the ground-truth reference. The ``m7 landmarks'' including the 4 eye corners, 1 nose tip, and 2 mouth corners are also included here in some comparisons with the idea that they provide higher importance information. Figure \ref{fig:lmkschemes} shows the two landmarks schemes used in this paper.

\begin{figure}[htb]
    \centering
    \includegraphics[width=0.5\linewidth]{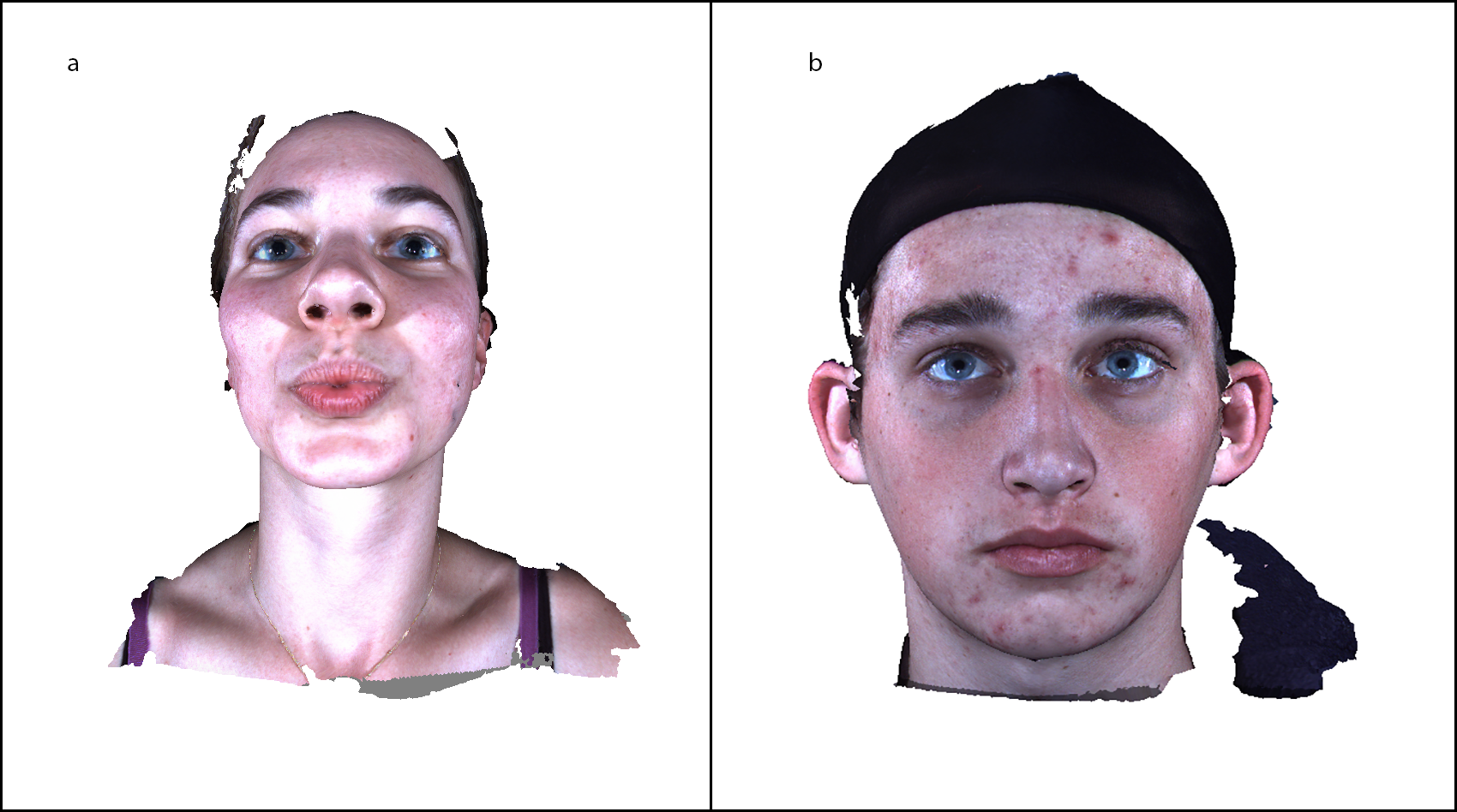}
    \caption{FACS \& NFACS faces: (a) It is a lip pucker expression and its corresponding action unit is 18\cite{}. (b) The participant provides a neutral expression}
    \label{fig:FACSandNFACS}
\end{figure}

In order to generate face images at controlled focal lengths and precise angle selections, we synthesized photo-realistic images using detailed 3D meshes captured from a structured-light 3dMD system. Our facial capture participants were asked to make different expressions following the Facial Action Coding System (FACS). FACS was created by the anatomist Carl-Herman Hjortsj\"{o} \cite{hjortsjo1969man} and further developed by Ekman etc. \cite{ekman1993facial} It provides a coding system which describes how to categorize facial expressions into Action Units (AUs) with muscle movements. We manually annotated the ground-truth landmarks in 3D for 84 faces from our participants, 64 from a set of FACS-capture expressions of two individuals and 20 of unique individuals with a range of ethnicity, age, and gender where the pose was neutral or a slight smile. Figure \ref{fig:FACSandNFACS} shows an example of FACS and neutral faces in our dataset. Landmark variation often occurs between in datasets, particularly for areas such as the jawline or eyebrows. For consistency,  we keep jawline points evenly distributed along the chin. In some projects, eyebrow points are placed at the center, bottom, or top of brow arcs. Good choices for landmarks points include those near high curvature or boundaries on objects. Here, eyebrows are marked anatomically at the supraorbital ridge or eyebrow ridge. 

\subsection{Evaluation Metrics}
We use ground-truth based localization error to evaluate performance in each case via root mean squared error (RMSE). Accurate landmarks are generated for each synthetic image by projecting manual 3D landmarks to match the rendered angle and field of view.  We use the method proposed by Jonston et al. \cite{Johnston2018} for calculating the RMSE:

\begin{align}  \label{RMSE}
  RMSE &= \frac{1}{K} \sum_{k=1}^{K} \sqrt{(x^{k} - \widetilde{x}^{k})^2+(y^{k} - \widetilde{y}^{k})^2}
\end{align}

  where \( x^{k}\), \( y^{k}\) denote each of the K predicted landmark $k$ in an image, and \( \widetilde{x}^k\), \( \widetilde{y}^k\) indicate the corresponding ground-truth landmark. 
Normalizing for face size in pixels is useful due to the variance across images Previously, RMSE is normalized by the ground-truth outer corners of the left eye and right eye landmarks(equation \ref{NRMSE:2})\cite{sagonas2016300}.
The error per landmark in image i is given as:

\begin{align}
\label{NRMSE:1}
\epsilon_i^k &= \frac{\sqrt{(x_i^k - \widetilde{x}_i^k)^2+(y_i^k - \widetilde{y}_i^k)^2}}{{d_{norm}}^i}\\
\label{NRMSE:2}
{d_{norm}}^i &= \sqrt{(\widetilde{x}_{le} - \widetilde{x}_{re})^2+(\widetilde{y}_{le} - \widetilde{y}_{re})^2}
\end{align}

where $(\widetilde{x}_{le},\widetilde{y}_{le})$ and  \((\widetilde{x}_{re},\widetilde{y}_{re})\) are the ground-truth outer corners of the left eye and right eye in the image i. In our case, however, our synthetic images vary with camera positions. The distances of outer-eye corners may have small impacts at side angles due to perspective projection.  Hence, we calculate Normalized Root Mean Squared Error (NRMSE) by normalizing per width of the head bounding box. We calculate the percentage of accepted points among all points to show the performance for each algorithm:

\begin{align}
P(k) &= 100 \frac{1}{I} \sum_{i=1}^{I}[i:\epsilon_i^k < Th]
\end{align}

where \( [i:\epsilon_i^k < Th] \) is a mask function that if the normalized distance \(\epsilon\) is less than $Th$, it is acceptable, and i is set to 1. Otherwise, the result is not acceptable, and i's value is set to 0. So, the overall performance over  K landmarks in each image for  I image set is:
\begin{align}
P &= 100 \frac{1}{K\times I} \sum_{k=1}^{K}\sum_{i=1}^{I}[i:\epsilon_i^k < Th]
\end{align}

\subsection{Camera Position \& Focal Length}
\begin{figure}[htb]
    \centering
    \includegraphics[width=0.5\linewidth]{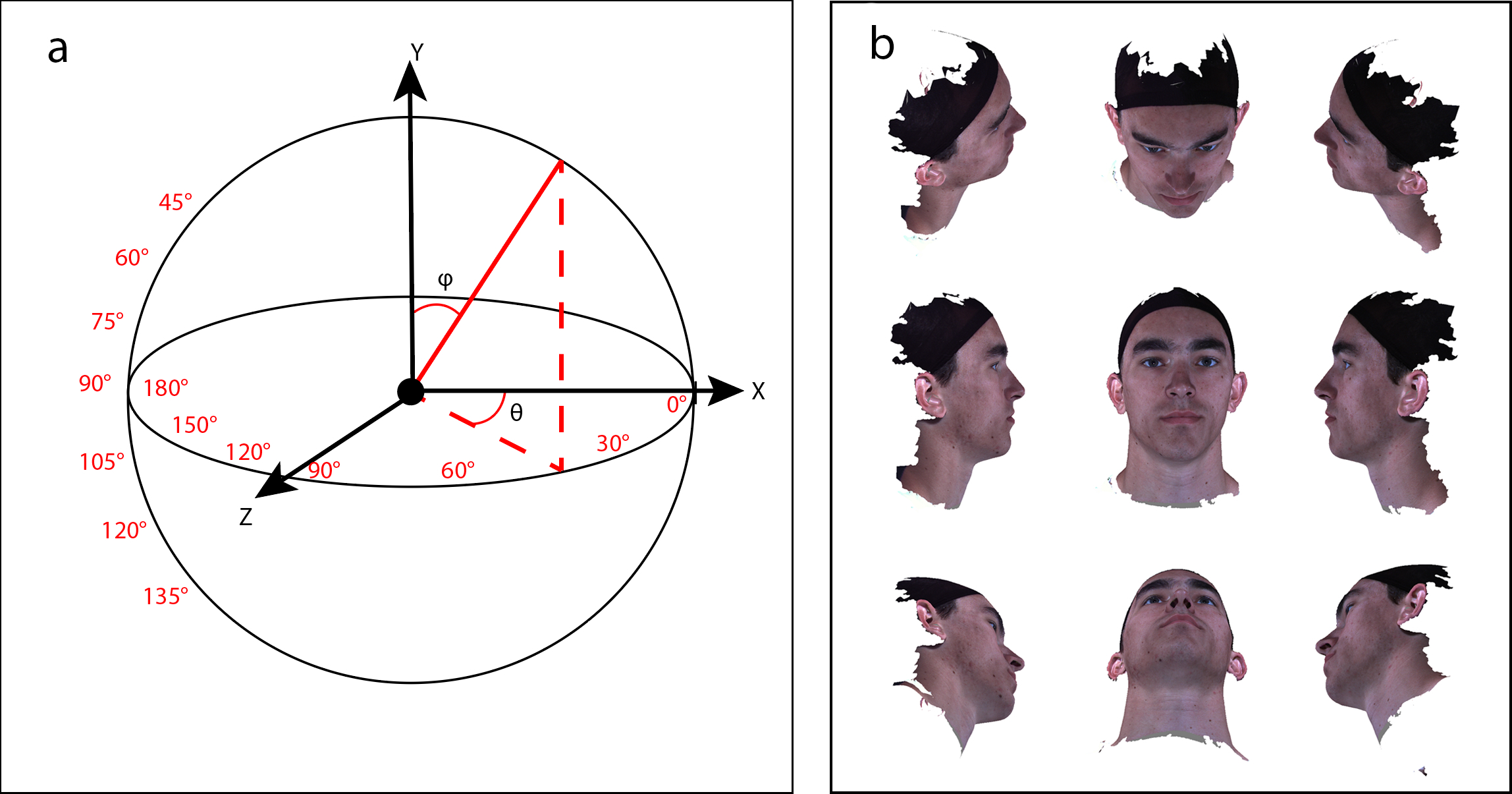}
    \caption{Camera Positions: (a) Right-handed system spherical coordinates. $\phi$ is the polar angle and  $\theta$ is the azimuthal angle. (b) samples for real face images taken from different viewing angles.}
    \label{fig:camPos}
\end{figure}

Our coordinate system follows the typical computer-graphics right-handed coordinate system convention, where the X-axis points to horizontal right, Y-axis points to vertical up, and Z-axis perpendicular to both X and Y points outward from the screen. In order to track the camera around each face, we use spherical coordinates to represent camera positions. Our interests are analyzing multiple viewing angles at a wide range of specific viewing angles. We define camera positions in spherical coordinates at \((r,\phi,\theta)\), where \(\phi\) is the polar angle (also known as zenith angle) from the positive Y-axis with \(45^\circ\leq\phi\leq135^\circ\), at $15^\circ$ each. We define $\theta$ to be the azimuthal angle in the xy-plane from the positive x-axis with \(180^\circ\leq\theta\leq0^\circ\) at intervals $30^\circ$. Lastly, $r$ varies for simulated focal length. Overall, we have 49 camera positions so that various front views of the face and some extreme camera positions could be tested. Figure \ref{fig:camPos} shows the position of spherical coordinates and samples of face images with different viewing angles.

\begin{figure}[htb]
    \centering
    \includegraphics[width=0.5\linewidth]{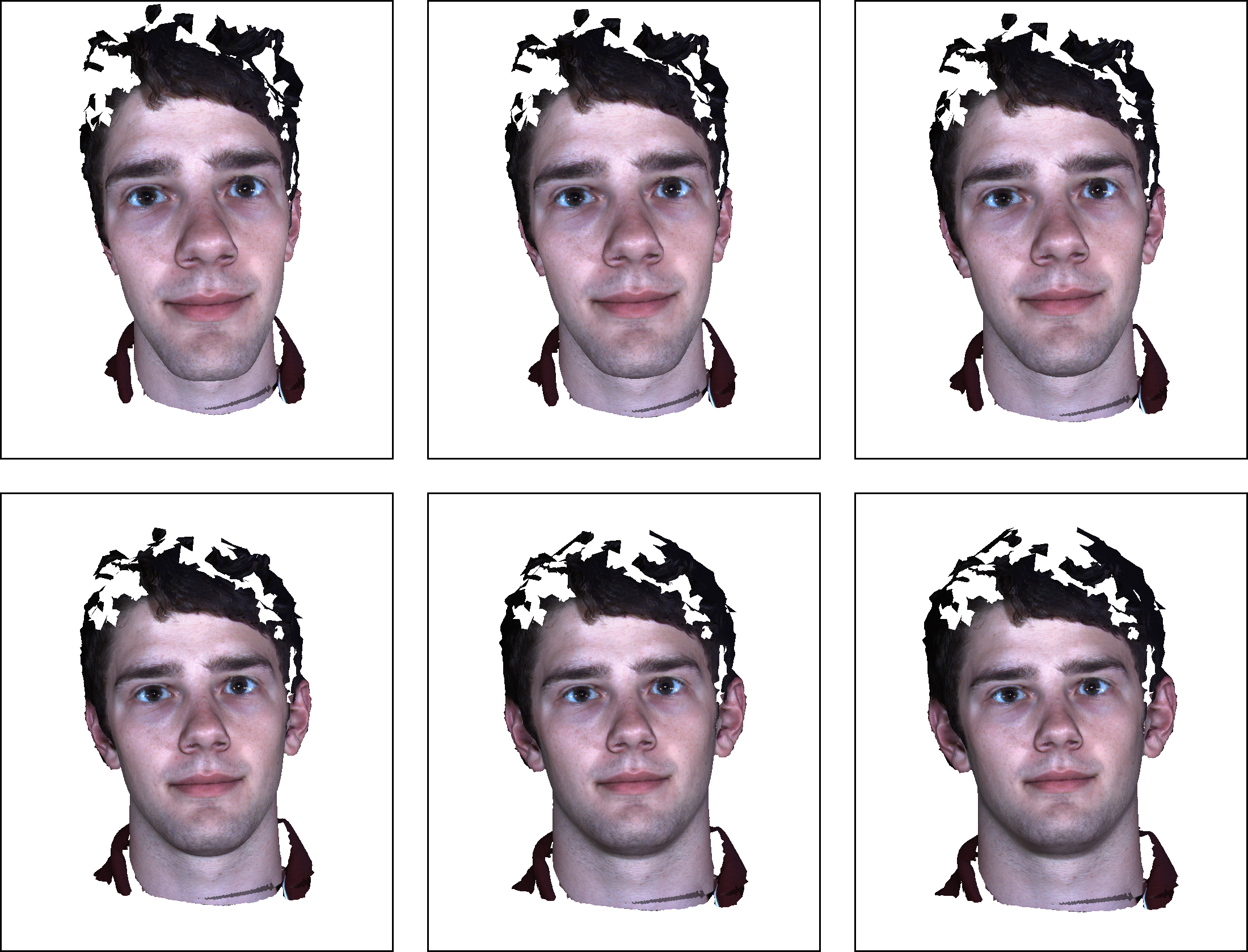}
    \caption{Camera Focal Lengths: First row from left to right: 24mm, 28mm, and 35mm; Second row from left to right: 50mm, 85mm, and 135mm}
    \label{fig:focalVaring}
\end{figure}

Focal length, relative to the dimensions of the film or digital sensor, determines the field-of-view on a physical camera, and there are also radial distortion issues relative to physical lenses and typical of certain optical designs such as pincushion and barrel distortion (these are not specifically included here but could warrant a follow-up study). In photography, a common standard of comparison of focal length to express field-of-view is relative to the standard of the 35mm-film frame size used for much of the twentieth century and carried forward into digital sensors.  This ``35mm'' frame size of 36mm across by 24mm down came to be a standard for still photography when Oskar Barnack doubled the individual frame from motion-picture film (standardized by Thomas Edison) to use in still cameras. The relation between angle of view and focal length is given by:

\begin{align}
 \label{equ:focalToAngle}
     \alpha &= 2\arctan\frac{d}{2f}
\end{align}

where $\alpha$ is the angle of view, $d$ denotes the size of film, and $f$ is the focal length (This could be calculated for horizontal,  vertical, or angular field of view). 
As can be seen by the relationship, shorter focal lengths widen the field of view and vice-versa.  To maintain a face of a relative size in images captured with different focal lengths, the distance to the camera needs to be changed. Perspective effects are modified as this occurs, as can be noted in Figure \ref{fig:focalVaring}.  Short focal lengths (wide-angle lenses) introduce a fair amount of facial distortion whereas longer lengths begin to approximate an orthographic projection that maintains relative distances among landmarks better. Although not tested here, these effects can be more pronounced near the edges of a capture frame.

As mobile phone photography increases, some of the most common focal lengths relative to the standard of comparison noted would equate to the 28mm to 35mm range of focal lengths, or a relatively wide field of view.  Interchangeable lens cameras or cameras with zoom lenses can vary the focal length. As can be seen from formula \ref{equ:focalToAngle}, a larger focal length lens has a narrower angle of view at the same camera-to-object distance which offers magnified, detailed photos. Focal lengths greater than 50mm are often used in longer range photography, long range biometric acquisition, and especially in head-and-shoulder portrait photography. For this study, common focal lengths of prime lenses used in still photography were chosen as the range, from 24mm (wide-angle on a 35mm system) to 135mm (slight telephoto on a 35mm system), with the range covering typical focal lengths used in photography and not including extreme wide-angle lenses nor extreme telephoto lenses. We chose six different types of common lens focal lengths (24mm, 28mm, 35mm, 50mm, 85mm, and 135mm) as our test domains for comparison.

\subsection{Face Landmark \& Alignment Methods}
 Wu et al.\cite{Wu_2018} mention classifying technology as holistic methods, constrained local model methods, and regression-based methods. Holistic methods treat a whole face image as the entire appearance and shape to train models. Constrained local models locate landmarks based on the global face but emphasizing local features around landmarks. Regression-based methods mostly are adopted for deep-learning, using regression analysis to map landmarks to images directly. Johnston et al. \cite{Johnston2018} believe that facial landmark detection methods can be divided into generative methods, discriminate methods, and statistical methods. Generative methods minimize the error between models and facial reconstructions. Discriminative methods use a dataset to train the regression models. Statistical models are a combination of generative methods and discriminate methods. {\c{C}}eliktutan et al. classify facial landmark detection into model-based (using the entire face region) and texture-based (matching landmarks to local features) \cite{Celiktutan2013}. Here we consider landmarking algorithms based on either statistical methods or deep-learning methods.  Statistical methods calculate the positions of landmarks using mathematical algorithms. Most of the traditional methods(e.g. AAM and ASM) can fall into this group. Deep learning methods feed facial images to train deep neural networks to locate landmarks.
 
 ASM and AAM models have performed among some of the best landmark-detection algorithms for nearly two decades. ASM, first introduced by Cootes et al., attempts to detect and measure the expected shape of a target in an image. ASM requires a set of landmarked images for training the model. The first step is using Procrustes Analysis to align all object images. A mean shape is calculated by PCA, which is applied to find eigen vectors and eigen values\cite{cootes1998active}. 
As an improvement of ASM, an active appearance model matches both shape and texture simultaneously and gives an optimal parameterized model. PCA is also applied for texture and once again for finding combined appearance parameters and vectors. Menpo provides five different AAM versions with two main groups: Holistic AAM (HAAM) and Patch AAM (PAAM)\cite{alabort2014menpo}. HAAM warps appearance information using a nonlinear function, such as Thin Plate Spline (TPS) and takes the whole texture into account when fitting, while the PAAM uses rectangular patches around each landmark as texture appearance. We test both HAAM and PAAM as separate techniques for comparison here. For building the AAM, we chose the widely used Helen Dataset which provides a high-resolution set of annotated facial images containing different ethnicities, ages, genders, head poses, facial expressions, and skin colors,  similarly used by Jonston et al. \cite{Johnston2018}. In order to reduce errors caused by facial detection, we extract faces from images using bounding boxes calculated from ground-truth landmarks and dilated by 5\%. 

In the past few years, deep-learning based neural-network methods have leveraged very large datasets for training and recently outperformed statistical shape and appearance models in many areas. We gathered three recent high-performing methods where implementations were available to compare in our various cases.  The first method is called the Position Map Regression Network\cite{feng2018joint}. 
The main idea of PRNet is creating a 2D UV Position Map which contains the shape of an entire face to predict 3D positions. PRNet employs a convolutional neural network (CNN) trained 2D images along with ground truth 3D dense position clouds created via 3D morphable model (3DMM). 3D positions are projected to the UV texture-map format and used in training the CNN. The UV texture map preserves 3D information, even posed with occlusions. 

The second method is the 3D Face Alignment Network (3D-FAN). Bulat and Tzimiropoulos use a 2D-to-3D Face Alignment Network combined with a stacked heat-map sub-network to predict Z coordinates along with 2D landmarks\cite{bulat2017far}. 

The third method from Bahagavatula et al. uses a 3D Spatial Transformer Network (3DSTN) to estimate a camera projection matrix in order to reconstruct 3D facial geometry. The method forms occluded faces with 2D landmark regression and predicts 3D landmark locations\cite{bhagavatula2017faster}.

These methods were trained on 300W-LP except for 3D-FAN which was trained on the 230,000 + 300W-LP. It would be prohibitive to attempt to include all recent deep-learning methods in this comparison, but these were chosen based on strong performance in recent publications, and we believe other recent methods would very likely perform similarly based on similar overall performance on the same datasets.

\section{Procedure}
\begin{figure}[htb]
    \centering
    \includegraphics[width=1\linewidth]{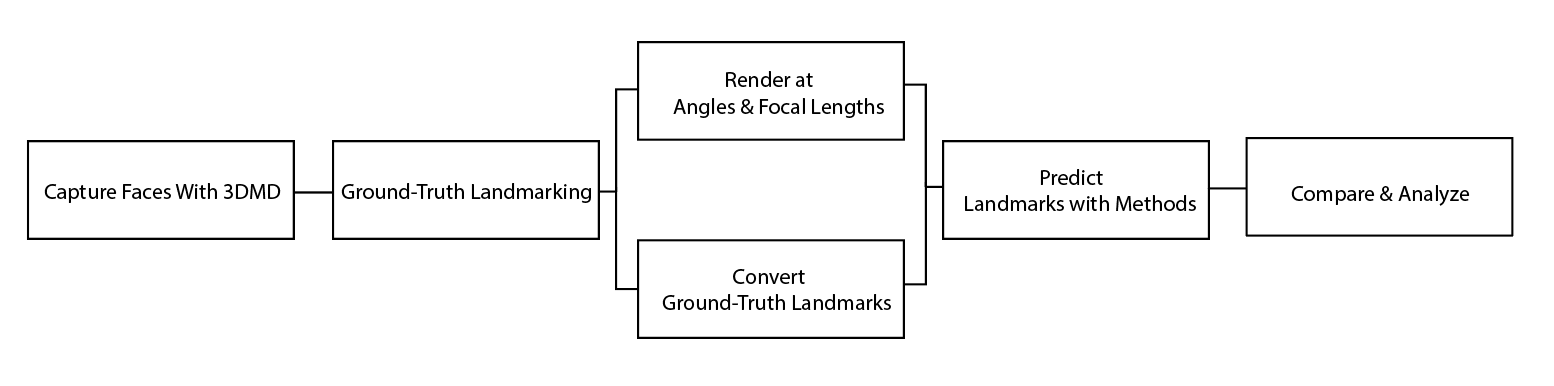}
    \caption{Flow chart of the main process for measurement}
    \label{fig:workflow}
\end{figure}

Figure \ref{fig:workflow} illustrates the main work flow of our approach to evaluate facial landmark and alignment algorithms.

\begin{figure}[htb]
    \centering
    \includegraphics[width=1\linewidth]{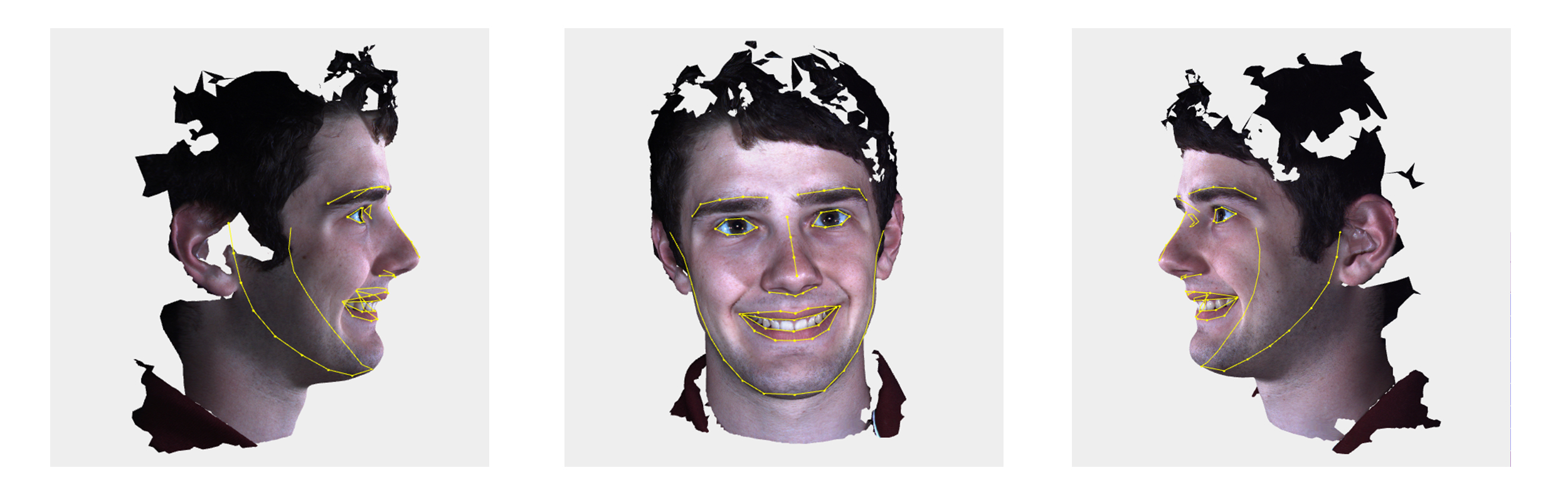}
    \caption{3D Ground-Truth Landmarking Using landmarker.io}
    \label{fig:3Dlandmarker}
\end{figure}

To calculate the RMSE (\ref{RMSE}) and NRMSE (\ref{NRMSE:1}), (\ref{NRMSE:2}) on landmarks, all measurements require ground-truth as references. All facial meshes with texture were manually marked using landmarker.io to create these ground-truth landmarks. Figure \ref{fig:3Dlandmarker} shows an example of 3D facial annotation in landmarker.io as performed on our dataset\cite{alabort2014menpo}.

\begin{figure}[htb]
    \centering
    \includegraphics[width=0.5\linewidth]{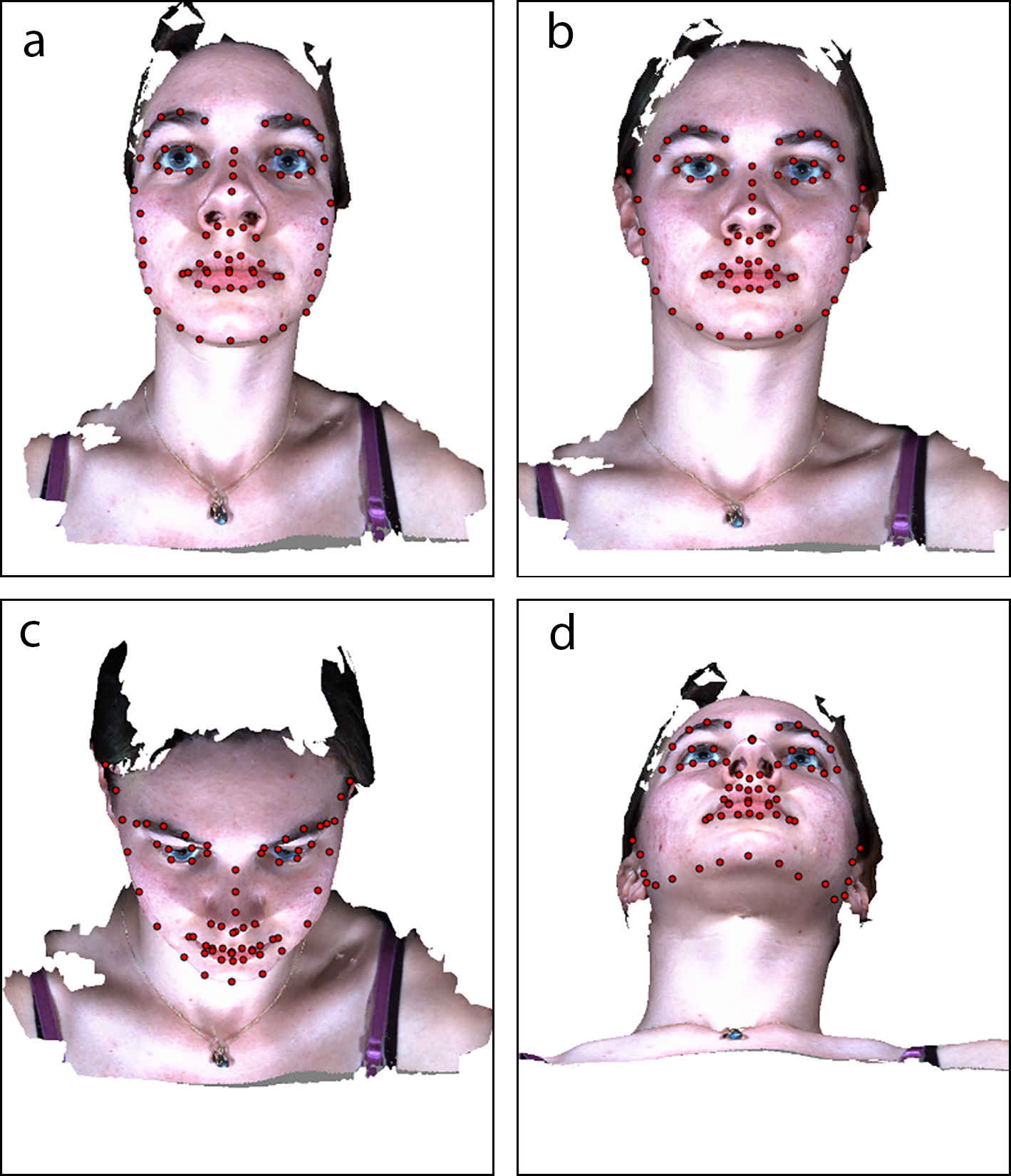}
    \caption{Converted 2D Ground-Truth on images: (a) and (b) are ground-truth landmarks with 24mm and 135mm focal length at the center view respectively. (c) and (d) are rendered landmarks from top view and down view with 135mm focal length}
    \label{fig:2DGroundTrueth}
\end{figure}

Using our own Python-, Qt-, and OpenGL-based lab application, Countenance Tool, we render 3D facial positions given varying angles and focal lengths. Since we compare how view angles and focal lengths affect landmark methods, we move the virtual camera to 49 different locations shown in Figure \ref{fig:camPos}. At each location, we rasterize faces with 6 different synthesized focal lengths (24mm, 28mm, 35mm, 50mm, 85mm, and 135mm) by changing the focal length parameter shown in equation \ref{equ:focalToAngle} before rendering. Overall, there are images at 49 angles and 6 focal lengths for each face. At the same time, we use the same camera matrices (varying with view of angles and focal length parameters)  to project the 3D ground truth landmarks to yield the ground-truth 2D landmarks at image coordinates. Figure  \ref{fig:2DGroundTrueth} shows a set of images with ground-truth landmarks of different focal lengths and viewing angles.

To summarize the workflow demonstrated in Figure \ref{fig:workflow},  we first performed facial geometry capture with a 3dMD system. The 3dMD system provided 3D meshes along with texture information. We then imported those into landmarker.io to annotate each face manually to generate 3D ground-truth landmarks. After getting the ground-truth, we rasterized each face at 49 angles and 6 focal lengths and calculated the ground-truth 2D landmark locations. Finally, we analyzed performance of each method by calculating NRMSE error between a method's predicted landmarks and the 2D ground-truth locations.

\section{Results and Discussion}
\begin{figure}[htb]
    \centering
    \includegraphics[width=1\linewidth]{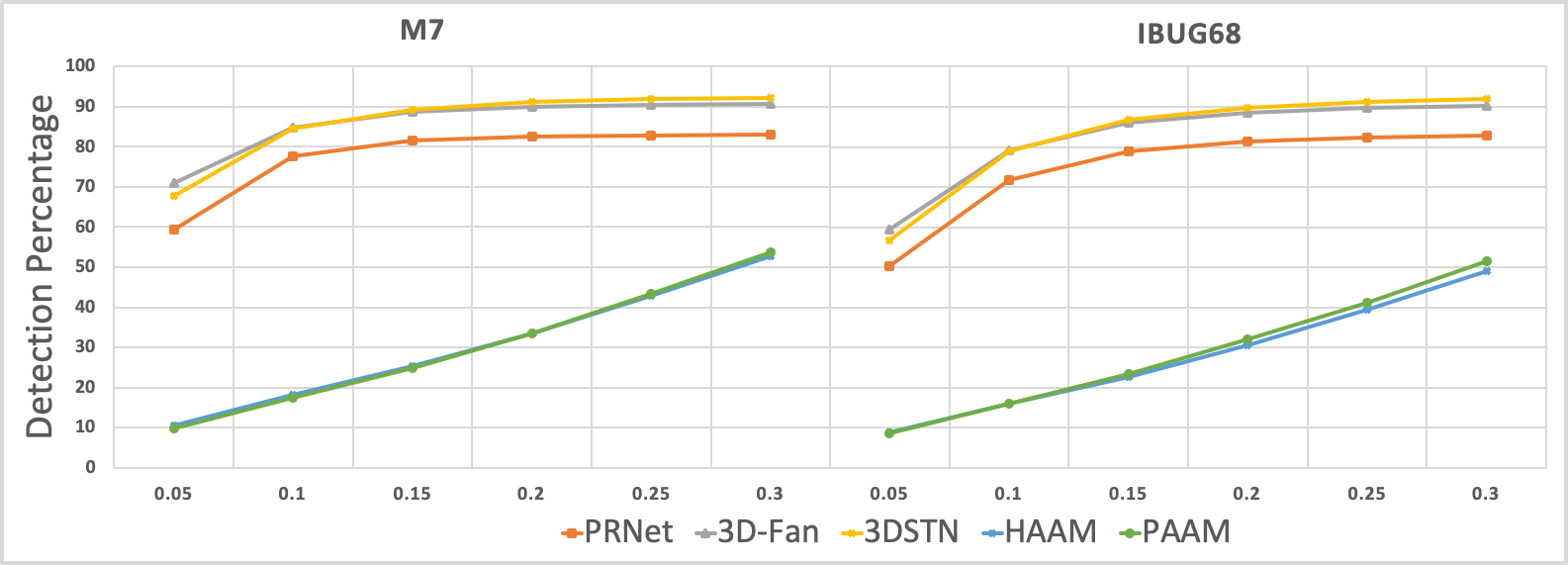}
    \caption{Performance Respect to Thresholds: Left is percentage of correctness of five methods with threshold increasing using 68 pts.; Right uses M7. }
    \label{fig:threshold}
\end{figure}

In this section, we compare the RMSE performance of the five methods with the full 68-point scheme and the reduced m7 scheme against 6 threshold levels. Figure \ref{fig:threshold} plots the percentage correctly accepted for each facial landmark and alignment method with both schemes. Generally speaking, as expected, the overall acceptance performance for each algorithm increases as the threshold widens. The m7 landmarking scheme tends to show better performance as a smaller set located at distinct ''corners.''
In general, the CNN methods perform better, but all are still subject to performance effects due to focal lengths and viewing angles.  It would be remiss to declare one method particularly better than another here, particularly since 3D-FAN was trained on an augmented dataset versus the others;  we used the publicly available pre-trained networks.  Compared to the neural-network techniques, the performance of traditional statistical methods is typically lower. As Cootes explains \cite{cootes1999comparing}, the performance of ASM and AAM is highly dependent on the starting position of landmark displacement. Higher accuracy of face detection tends to improve landmark detection.  

\begin{figure}[htb]
    \centering
    \includegraphics[width=1\linewidth]{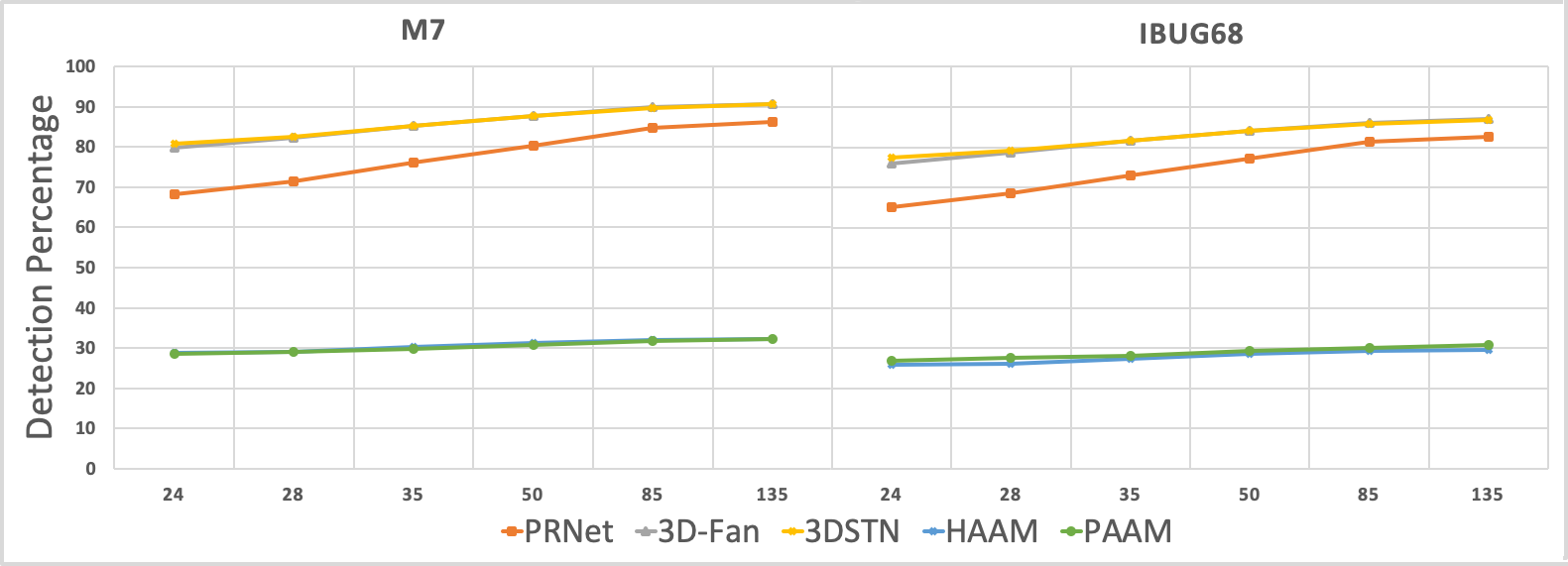}
    \caption{Focal Length Varying: Left half of the chart uses M7, right half uses 68 points.}
    \label{fig:focal}
\end{figure}

One of the main contributions of this paper is demonstrating the effect of focal-length on landmarking accuracy. Figure \ref{fig:focal}  demonstrates lower performance with a wider field-of-view, associated with strong perspective effects, and better performance as focal length increases. There is expected leveling in performance with focal length increase.

\begin{figure}[htb]
    \centering
    \includegraphics[width=1\linewidth]{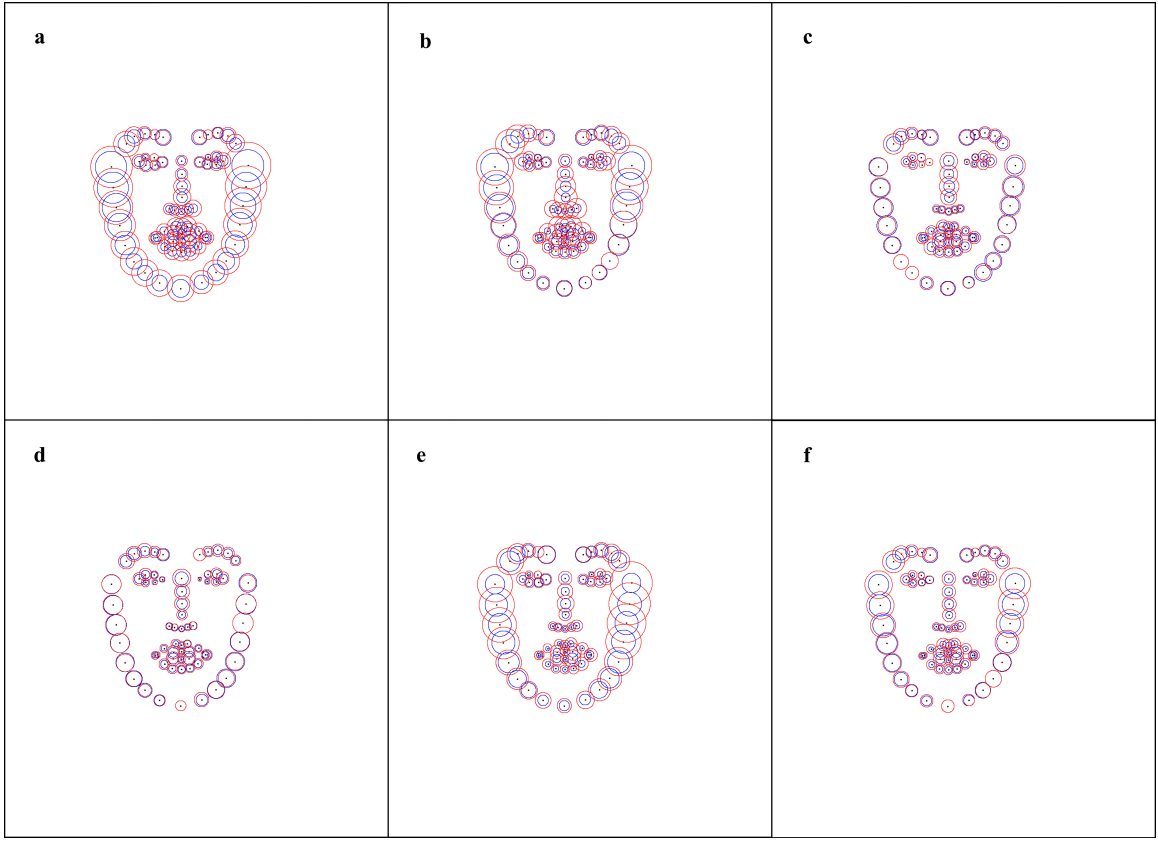}
    \caption{RNSE Distance: Blue circles are RMSE at 135mm, red circles are 24mm.(a) HAAM, (b) PAAM, (c) PRNet, (d) 3DSTN, (e) 3D-Fan, and (f) average result for all algorithms.}
    \label{fig:ransDist}
\end{figure}

In order to visualize effects on specific landmarks at different focal lengths, we drew the 68-point  landmarks located by each method and the average of the frontal view for the extremes (135 mm lens in blue circles based on RMSE and 24mm in red. This shows which landmarks are most affected by the focal-length perspective warping. Figure \ref{fig:ransDist} also reflects the data depicted in the Figure \ref{fig:focal}. The radius of the RMSE presents how far each predicted landmark is from the ground-truth. The result shows that all of the landmarks that are close to the center of faces have more accurate predictions, while landmarks along facial edges have lower accuracy predictions due to projective distortions; particular, corners of eyes and lips seem affected.

\begin{figure}[htb]
    \centering
    \includegraphics[width=1\linewidth]{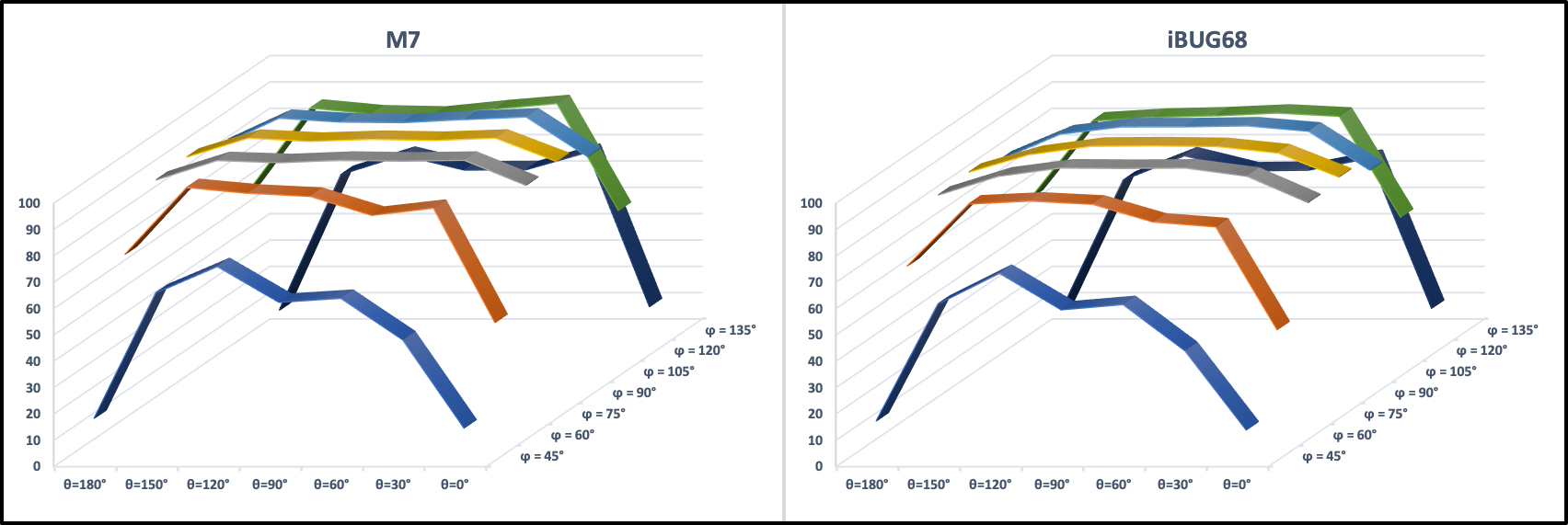}
    \caption{PRNet:Camera Position Varying Across $\phi$ (from $45^\circ$ to $135^\circ$) and $\theta$ (from $0^\circ$ to $180^\circ$).}
    \label{fig:camVaryingPRNet}
\end{figure}

\begin{figure}[htb]
    \centering
    \includegraphics[width=1\linewidth]{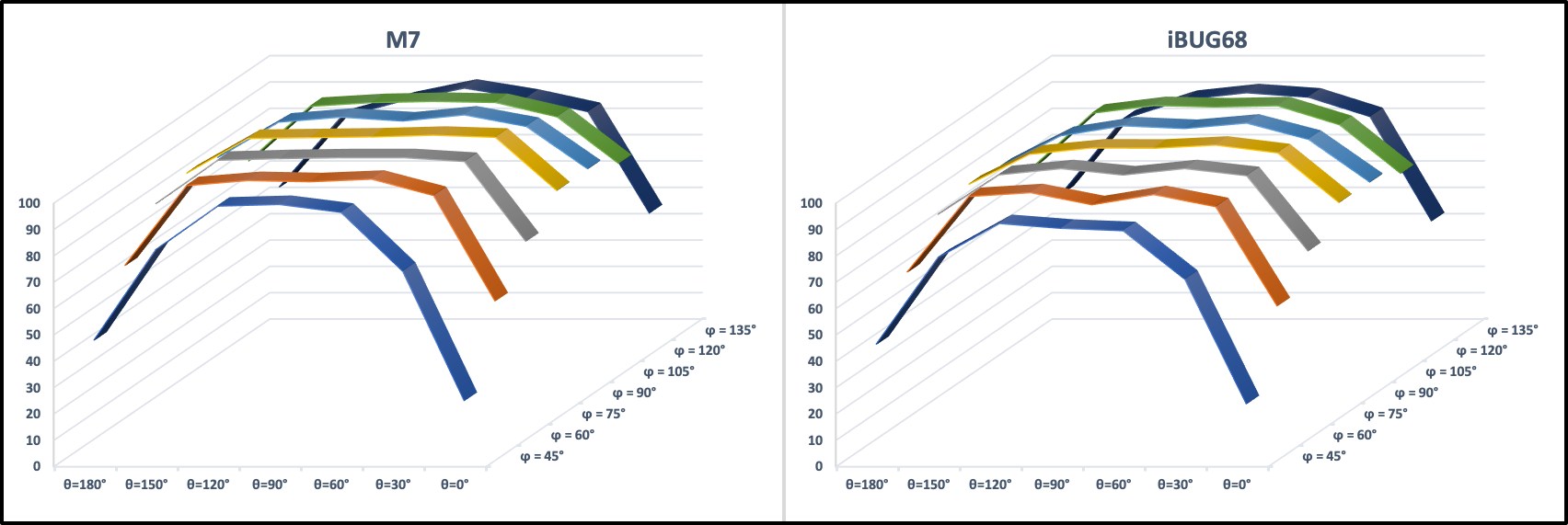}
    \caption{3DFan: Camera Position Varying Across $\phi$ (from $45^\circ$ to $135^\circ$) and $\theta$ (from $0^\circ$ to $180^\circ$). }
    \label{fig:camVarying3DFan}
\end{figure}

\begin{figure}[htb]
    \centering
    \includegraphics[width=1\linewidth]{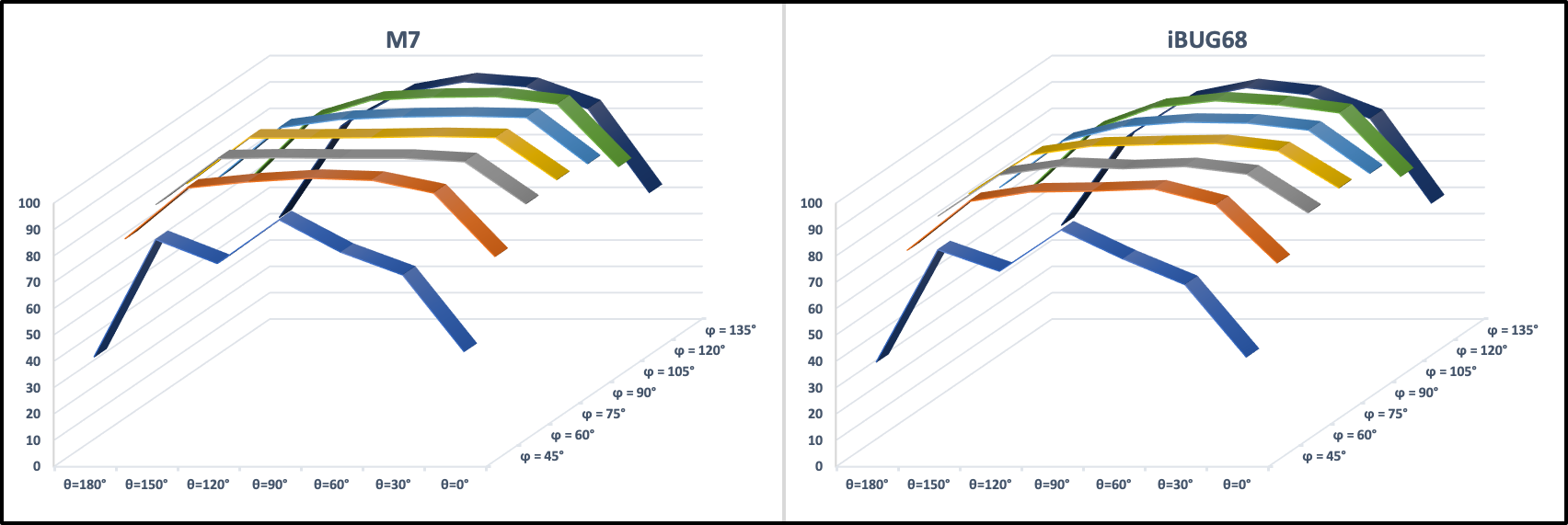}
    \caption{3DSTN: Camera Position Varying Across $\phi$ (from $45^\circ$ to $135^\circ$) and $\theta$ (from $0^\circ$ to $180^\circ$). }
    \label{fig:camVarying3DSTN}
\end{figure}

\begin{figure}[htb]
    \centering
    \includegraphics[width=1\linewidth]{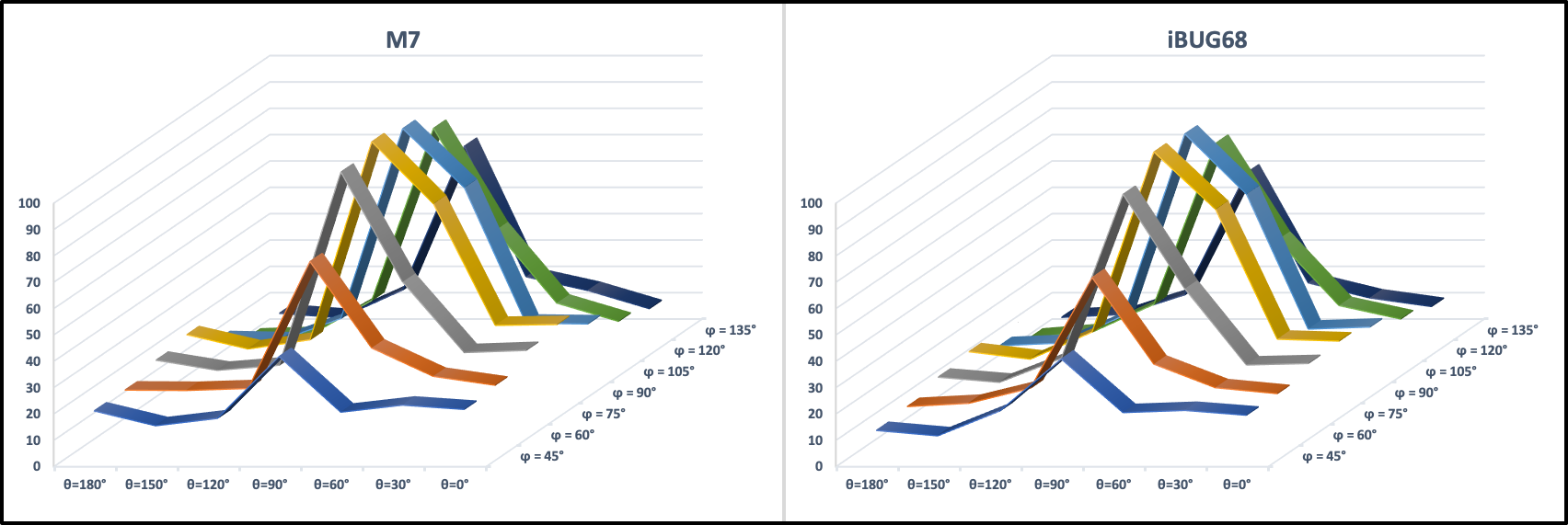}
    \caption{HAAM: Camera Position Varying Across $\phi$ (from $45^\circ$ to $135^\circ$) and $\theta$ (from $0^\circ$ to $180^\circ$).}
    \label{fig:camVaryingHAAM}
\end{figure}

\begin{figure}[htb]
    \centering
    \includegraphics[width=1\linewidth]{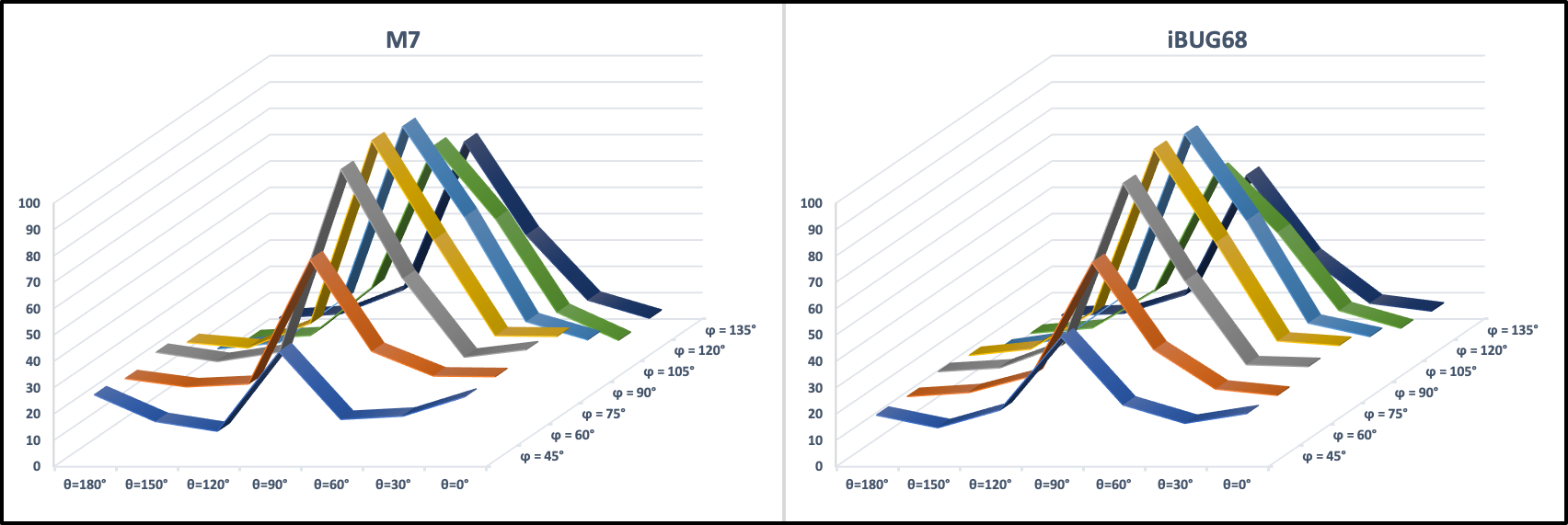}
    \caption{PAAM: Camera Position Varying Across $\phi$ (from $45^\circ$ to $135^\circ$) and $\theta$ (from $0^\circ$ to $180^\circ$).}
    \label{fig:camVaryingPAAM}
\end{figure}

The last consideration for this paper is systematic adjustment of the camera's viewing angle across the viewing hemisphere. We place the camera at 49 different positions with extreme poses included. When the camera views from the center ( $\theta \cong 90^\circ$ and $\phi \cong 90^\circ$), the performance results are better than when the camera view from the sides. The landmark predictions at $\phi$ around $45^\circ$ and $135^\circ$ have the lowest performances due to extreme viewing angles.  As expected, performance drops as the view moves to the more extreme angles, and the rate of effect for each method are shown in Figures \ref{fig:camVaryingPRNet} to \ref{fig:camVaryingPAAM}.


\begin{figure}[!h]
    \centering
    \includegraphics[width=1\linewidth]{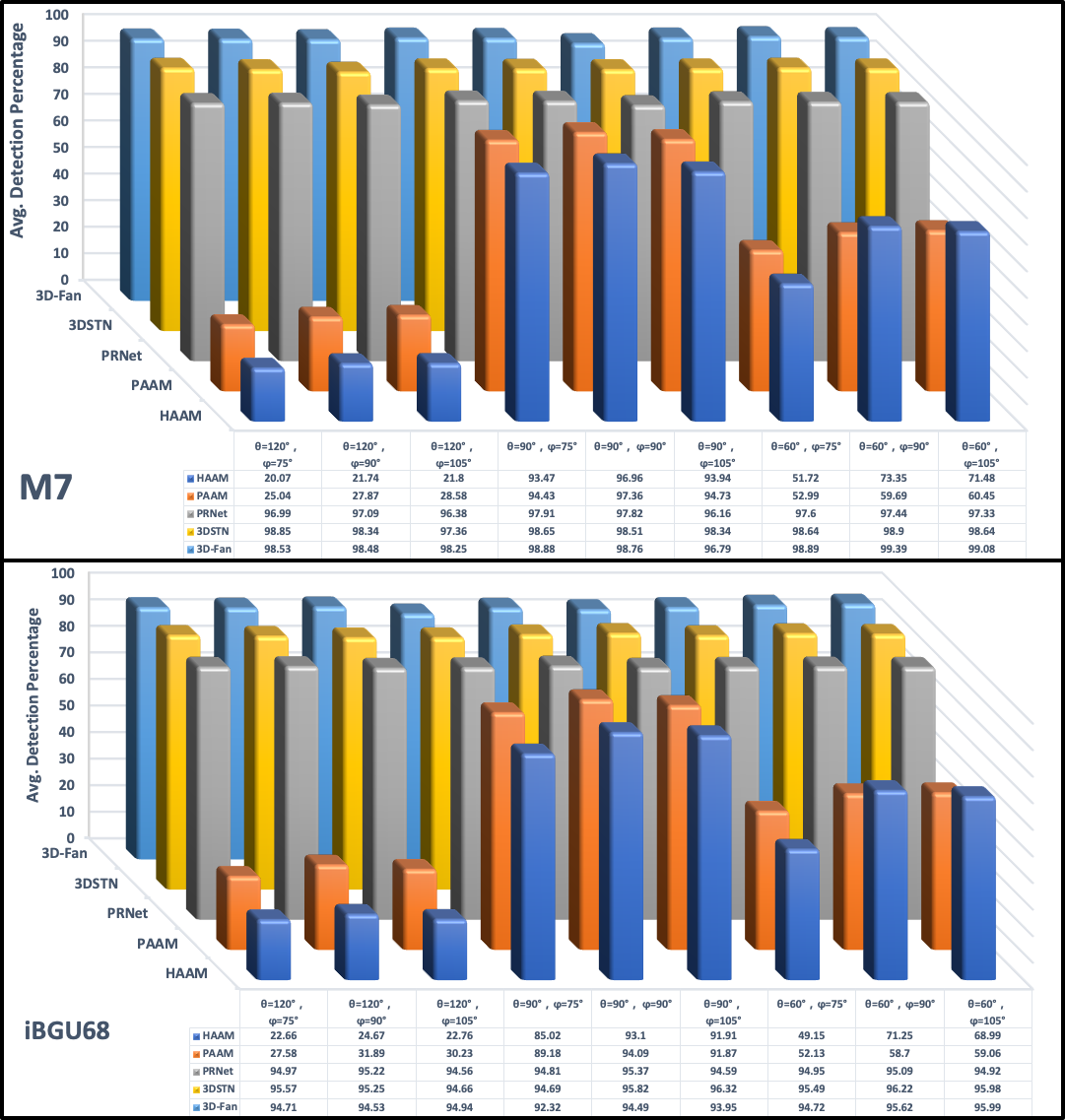}
    \caption{Camera Position with Group of frontal View: The frontal view positions where $\phi$ = $75^\circ$, $90^\circ$, and $105^\circ$; $\theta$ = $60^\circ$, $90^\circ$, and $120^\circ$.}
    \label{fig:FrontInGroups}
\end{figure}

\begin{figure}[!h]
    \centering
    \includegraphics[width=1\linewidth]{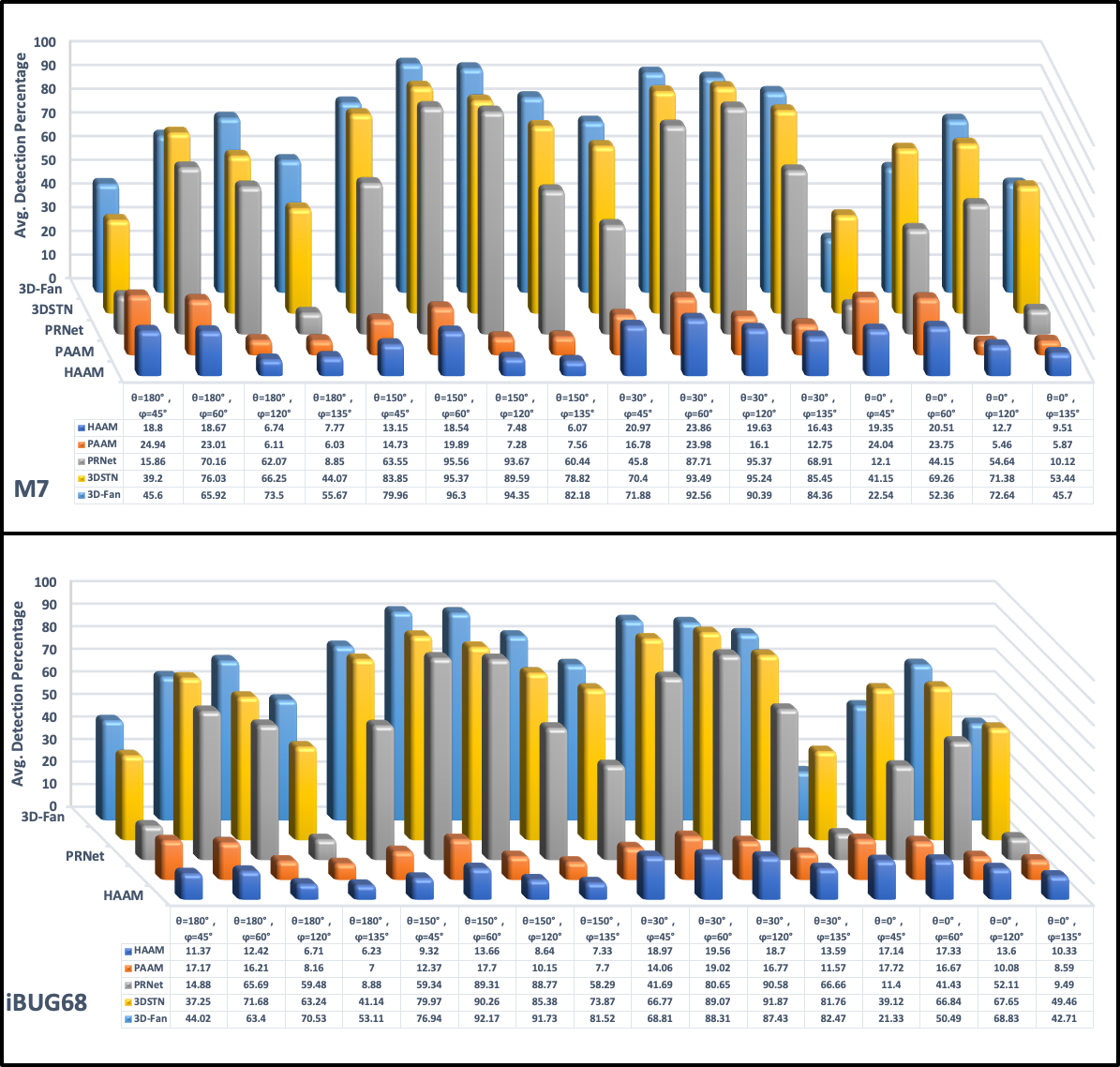}
    \caption{Camera Position with Group of Outer View: The outer view position where $\phi$ = $45^\circ$, $60^\circ$, $120^\circ$, and $135^\circ$; $\theta$ = $0^\circ$, $30^\circ$, $150^\circ$, and $180^\circ$.}
    \label{fig:FrontOutGroups}
\end{figure}

Most of the facial landmark and alignment algorithms perform well at frontal views, and the detection precision relies on the training set variability. Attempting to delineate prediction differences between extreme-view cases  and center view cases, we chose the most centered view image ($\phi=90^\circ$ and $\theta=90^\circ$), as well as 8 images surrounding by it, to be the frontal group (Figure \ref{fig:FrontInGroups}). The rest of the images are the outer group (Figure \ref{fig:FrontOutGroups}). Front view detection can approach almost 100\% accuracy especially at center view for deep neural networking methods. The precision rate drops more than 50\% approaching extreme angles ($\theta=0^\circ$ and $\theta=180^\circ$). 

Part of the set of the images used were also based on 3D captures of action units from FACS which taxonomizes individual physical expression of emotions. The results shown in Figure \ref{fig:FACSandNFACS} illustrate that in general the landmark-prediction methods work better on neutral faces due to FACS faces having more facial expressions which increase prediction difficulties. Performance decreases across wide field of view and view angle are consistent.

\begin{figure}[!h]
    \centering
    \includegraphics[width=1\linewidth]{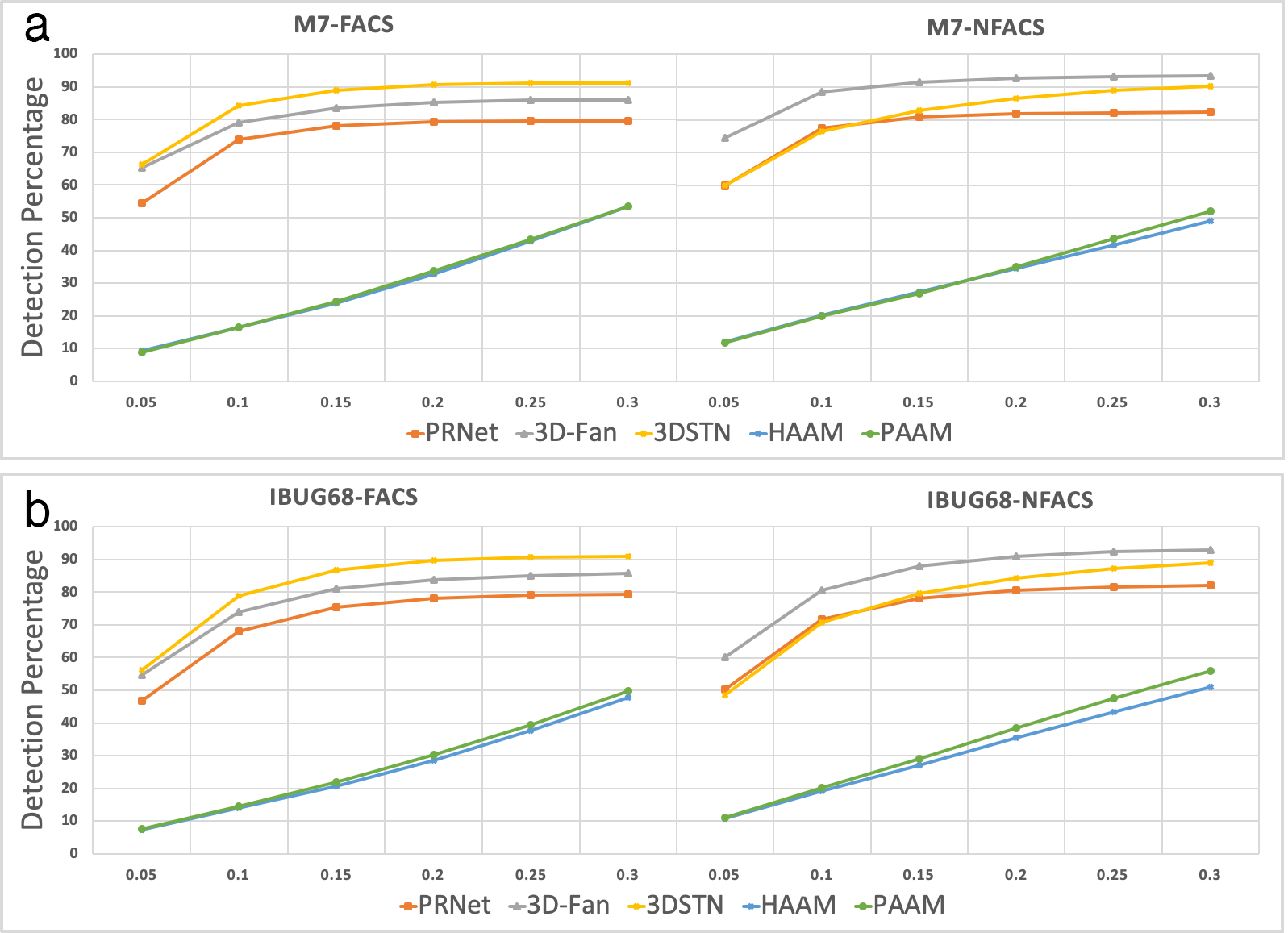}
    \caption{FACS and non-FACS Comparison: (a) FACS NRMSE and non-FACS NRMSE with M7 and (b) is iBUG-68 landmark schemes.}
    \label{fig:FACSnNonFACS}
\end{figure}

\section{Conclusion}

In conclusion 3DSTN, PRNet, and 3D-FAN methods generally work better than traditional statistical methods.  Deep-learning methods have become become the prevalent research direction for the time being, but they are still subject to viewing angles and also, particularly, lens effects that have rarely been considered during any performance evaluations.

Increasing focal length tends to improve the landmark and alignment performances due to less projection distortion. This could inform design decisions for camera system and lens chosen for a biometric system, or it could be used to inform future algorithm design. 

Given experimental results, all methods, as expected, work best from frontal-viewing angles.
It is also interesting to note that the slope of fall-off for the performance decrease introduced by shorter focal lengths (wider field-of-view) is less for the AAM based methods and the 3DSTN approach.  This is likely due to the AAM methods being based on image features, and the PAAM more specifically emphasizing local image features.  3DSTN likely does well as part of the method specifically estimates a camera projection matrix, which in some sense should help counteract some of the focal-length introduced perspective issues.  PRNET and 3D-FAN methods using more general 3D data are likely more affected, and the larger training set for 3D-FAN likely assists its performance here.

One limitation of statistical algorithms is the landmark detection performance is tied to the head pose variation in the training set. When applying PCA, the first N eigen vectors are chosen as the main components. Typically, these are chosen based on representing $\pm3$ standard deviations from the mean value. Based on this limitation, the landmarking performance for extreme view angles, as often shown, drops. However, the CNN methods that all incorporate some system of 3D reference tend to do better as viewing angles move from the center; however they still suffer performance drops and are still affected by shorter focal lengths. 

Since focal length variance does affect final face landmark and alignment performance, future work could include use of this to augment training data. This could be done through data collection or use of synthetic data.

Meta-data from capture lenses stored in digital photographs is often removed by the time images reach large datasets, but it would be interesting to note such effects from in-the-wild photographs. In the meantime, training with synthetic data that includes controlled variance of viewing-angle ranges as well as varying focal length, added to photographic datasets, should likely improve results.  

In the future, image acquisition should not only cover pose, illumination, expression, ethnicity, skin color, etcetera, but also include consideration of full camera and lens parameters when possible.


%% file: deepnet.tex
\chapter{Multiple View Neural Regression of Facial Shape Model}
The creation of production-ready retopologized meshes for human faces necessitates substantial manual effort and time investment. Recent advancements in deep neural networks show promise in directly learning mappings to 3D face meshes from facial photographs. While monocular image-based 3D face synthesis is a widely explored area of research,  its limitations are in acquiring sufficient information, such as depth and image quality information. Conversely, multi-view 3D face reconstruction, leveraging inputs from multiple viewpoints, offers enhanced accuracy. However, the absence of accurate global correspondences among all views presents obstacles in neural network training. In this paper, we propose a novel approach for automatically predicting riggable 3D Morphable Model (3DMM) face meshes from multi-view images. Our methodology comprises three key features: 1) To train the neural network, an Appearance 3D Morphable Model (A3DMM) is built from a small set of scans of individuals' faces and used with physically based shading and random generation to create a synthetic dataset, using our software Visage Craft. 2) Leveraging the synthetic dataset to automatically establish one-to-one correspondences across multi-view images, enabling supervised training of neural networks. 3) With supervised learning, the method could be easily extended to other neural network architectures.

\section{Introduction}
The prevalence of digital characters in feature films and video games underscores the growing importance of achieving realistic human faces in computer graphics. However, overcoming the uncanny valley phenomenon remains a significant research obstacle. High-fidelity facial datasets enable the creation of lifelike human animations and facilitate the efficient generation of humanoid creatures and enhancement of final rendering quality. Traditional film and game-production pipelines involve multiple intricate processes to develop animation-ready face assets. These include geometry acquisition through photogrammetry, mesh refinement and retopology, and texture transfer and correction. Each of these subprocesses presents unique challenges stemming from the intricate geometry and nuanced physical characteristics of human faces, compounded by inherent limitations in computational resources. As a result, each step represents a distinct research domain with numerous unresolved obstacles.

Acquiring three-dimensional vertices represents a crucial stage in 3D facial synthesis \cite{egger20203d}. A common approach involves reconstructing facial geometry using multi-view stereo (MVS) techniques to obtain detailed 3D scans. The fundamental principle of MVS entails identifying point-to-point correspondences across multiple image pairs. Active systems, employing techniques such as laser projection, polarized lighting, or infrared patterns during image capture, are often utilized for this purpose. However, such approaches typically require professional setups or specialized hardware \cite{blanz1999morphable,3dmd_2024, Ghosh_polorized, zhangCyberware}. Alternatively, Beeler et al. \cite{beeler2010high} proposed a facial synthesis pipeline using passive multi-view images, wherein correspondences are identified by analyzing features from stereo images along epipolar lines. Despite its potential advantages, the passive setup is hindered by its time-consuming nature.  An alternative approach involves the use of learning-based methods, such as the 3D Morphable Model (3DMM), initially proposed by Blanz and Vetter in 1999 \cite{blanz1999morphable}. The 3D Morphable Model (3DMM) is trained on a limited set of 3D facial meshes that share a common topology. Using principal component analysis (PCA), it encodes shape variation in a low-dimensional space, with texture modeled in a similar fashion. Early approaches in face synthesis applied 3DMM parameters to image-based facial representation by minimizing energy functions that measure the difference between the model projection and the input image. That analysis-by-synthesis led the research trending for the past 20 years\cite{egger20203d}. It is a known challenge in the computer graphics and computer vision research community that the ability to acquire a large facial dataset for 3DMMs is quite limited.  It wasn't until a decade later that the first 3D morphable model, the Basel Face Model, became publicly available for use \cite{Pascal5279762}. Nonetheless, both methods necessitate manual clean-up to achieve satisfactory results.

In 2018, the research conducted by Johnston and Chazal wrote a survey and they observed a significant shift of interest towards deep-learning techniques, motivated by their potential to enhance performance and facilitate the execution of 3D alignment and construction tasks\cite{johnston2018review}. This paradigm shift improves the power of deep learning to generate high-fidelity retopologized face meshes while enhancing workflow efficiency.  Regardless of the neural network architectures employed, the predominant approaches for regressing 3D shapes from 2D images involve either single-view or multiple-view images. Single-view-based deep learning utilizes training datasets of in-the-wild monocular 2D images and video sequences  \cite{koestinger11a, zhang2013high, le2012interactive, sagonas2016300}. However, due to the variability in head poses, occlusions, lighting conditions, and depth ambiguity in the training images, the performance of single-view-based regression often degrades when applied to non-frontal view facial images\cite{pengfei_mv}. To overcome these drawbacks, multi-view-based 3D face regression leverages several images of the same subject from different viewing angles. This approach provides neural networks with additional information to infer 3D geometry more accurately.  However, capturing multiple viewpoints typically involves camera calibrations, often requiring professional devices like the Light-Stage, initially developed at the USC ICT lab\cite{debevec2000acquiring}. As a result, there are limited publicly available training databases for this purpose. Moreover, most multi-view-based neural network methods encounter difficulties in learning 3D meshes due to the challenges of establishing accurate feature correspondences among all views. To solve this issue, inspired by \cite{carreira2016human}, some approaches \cite{wu2019mvf, richardson20163d} render predicted 3D meshes into 2D images and incorporate them as extra channels in the multi-view images. Not only is rendering during training time-consuming, but it also requires masking out occluded facial parts due to varying view angles\cite{wu2019mvf}. Alternatively, recent methods \cite{dias,richardson20163d} utilize 2D landmarks to refine final predictions. Nevertheless, these methods heavily rely on the accuracy of landmark detection \cite{richardson20163d, dias}. \cite{li_2021}  highlights the impact of camera information, such as focal length and viewing angles, on landmark detection accuracy, which is rarely considered in training datasets.

This paper primarily focuses on learning retopologized meshes from multiview images. Traditional 3D morphable models often lose high-frequency facial details due to linearity constraints \cite{li2021tofu, TMPEH:CVPR:2023}. However, parametric models remain widely used due to a key advantage: the ability to quickly generate the mesh's appearance by using a small set of dataset\cite{porter_2022}.  This paper proposes a multiview image-based 3D facial synthesis pipeline and investigates the feasibility of training the neural network with fully synthetic data. Our approach draws inspiration from pioneering work in Deep Appearance Models for Face Rendering \cite{Lombardi:2018}, which aimed to reconstruct a view-independent face mesh using a Variational AutoEncoder (VAE) \cite{kingma2022autoencoding}. The neural network architecture is implemented similarly to Instant Multi-View Head Capture through Learnable Registration \cite{TMPEH:CVPR:2023}. Synthetic data generation and follows a similar approach to \cite{dias}. Unlike their work, this pipeline can generate multiview synthetic data with camera information. Furthermore, to enhance the performance of the neural networks, we adopt a feature sampling process combining \cite{li2021tofu,TMPEH:CVPR:2023, richardson20163d, wu2019mvf}. While previous approaches for learning facial models from images have either relied on large datasets of real people captured in controlled environments or constructing datasets from single human face images using statistical models. We present work toward nearly instantly generating a face asset from pairs of image views. An Appearance 3D Morphable Model (A3DMM) is built from a small set of scans of individuals' faces and used with physically based shading and random generation to create a synthetic dataset, using our software Visage Craft. Leveraging fully controllable synthetic data, projecting mesh points as preprocess gives an accurate ground truth for supervised training. Integrating mesh point samples into the loss function to establish one-to-one correspondences aligns our method with Beeler's \cite{beeler2010high} passive stereo reconstruction approach, easily adapting other neural network architectures. With dense correspondences readily established in the dataset, predictions can be conveniently tested on both passive and active data.

In summary, the contributions of this work are:
\begin{itemize} \item Generate synthetic multi-view facial data with associated camera parameters using our internally developed software, Visage Craft. Visage Craft is built upon a small collection of real human facial scans and employs physically-based shading to achieve photorealistic rendering.

\item In addition to multi-view imagery, our method leverages a 3D Morphable Model (3DMM) to produce a consistent, retopologized mesh. Since the synthetic dataset is fully controlled, manual annotations, such as landmarks and dense point-to-point correspondences, can be efficiently created and easily integrated into the neural network training process.

\item Given the synthetic and fully controllable nature of the data-generation process, datasets can be flexibly customized according to specific training requirements. \end{itemize}

\section{Related Work}

Our approach is guided by the task of learning 3D retolopolgized mesh points from synthetic multi-view face images. The following sections will dive into the most relevant prior research concerning our approach. Furthermore, we will explain the distinctions between our methodology and previous work within these domains.

{\bfseries Appearance Models:} Cootes et al. introduced an Active Shape Model (ASM), which used landmarks to manually annotate faces in images and train a statistical face model \cite{COOTES199538, cootes2000introduction}. The landmarks brought one-to-one correspondence across all training faces. Applying PCA to those landmarks allowed ASM to describe meaningful features of faces during feeding time. Cootes et al. proposed an Active Appearance Model (AAM) by integrating the texture information into the ASM while training \cite{cootes1998active}.  AAM combined the shape eigenvectors and color eigenvectors into one joint model. The combined parameters would be learned during the fitting stage. Additionally, a more comprehensive publicly accessible 3D Morphable Model (3DMM), known as the Basel Face Model, was developed from a large-scale dataset. \cite{bfm09}.

{\bfseries Single View 3DMM Based Neural Network:} The robustness of deep learning methods, in contrast to statistical approaches, is further enhanced by the abundance of publicly accessible databases tailored for training neural networks. Notable examples include 300W \cite{300w}, COFW \cite{cofw}, WFLW \cite{wflw}, and AFLW \cite{aflw}, which cover a broad spectrum of facial attributes, encompassing variations in age, ethnicity, skin color, expression, and pose. Leveraging such extensive datasets contributes significantly to the overall robustness of deep learning methodologies.

Regression from 2D to 3D typically requires prior parametric models to supervise the learning process \cite{richardson20163d, tewari2019fml}. Given that the quality of the model is highly dependent on the accuracy of the ground-truth linear models, obtaining precise and controllable face geometry becomes critical. One approach to achieving this is through synthetic data generation. Notably, Dias et al. pioneered one of the earliest methods to train models using fully synthetic data \cite{dias}. However, due to the limited variability in fully synthetic datasets, additional landmarks are often required to refine the final predictions. To address this limitation, recent advancements by Moser et al. and Serra et al. \cite{serra2022, moser2021} have proposed methods that first learn to generate images of computer-generated animated characters and subsequently derive the PCA model from these intermediate 3D characters. Despite their effectiveness, these approaches introduce additional complexities to the models.

{\bfseries Multi-View 3DMM Based Neural Network:} To our knowledge, Lombardi et al. were the first to explore learning appearance models by consolidating multiview images into a single texture map \cite{Lombardi:2018}. Subsequently, their work extended to learning neural volumes from novel viewpoints. A major limitation of single-view-based deep learning models is the absence of depth information in the input images. In contrast, multiview approaches involve aggregating images from various viewpoints to train parametric models \cite{pengfei_mv, Ramon2019, wu2019mvf}. However, the lack of feature correspondences across different views poses a challenge for neural networks in learning the relationships between multiple frames. To address this issue, methods such as ToFu and Tempeh employ grid-based sampling of image features \cite{li2021tofu, TMPEH:CVPR:2023}.

{\bfseries Multi-View Volumetric Neural Network:} In order to establish one-to-one correspondences from multi-view images, ToFu and Tempeh project grids onto the images and sample features accordingly \cite{li2021tofu, TMPEH:CVPR:2023}. They then estimate the 3D mesh points by employing volumetric surface-aware feature fusion. However, both approaches utilize raw multi-view stereo scans and rely on minimizing a geometric loss commonly applied in surface registration. As a result, the predicted meshes are noisy triangle meshes, requiring additional mesh wrapping steps before they can be used for rigging and production.

{\bfseries Neural Network Architectures:} Several studies \cite{johnston2018review, cceliktutan2013comparative} have observed a notable shift towards deep learning methods, driven by their enhanced performance and capability for handling complex 3D alignment tasks. Given the inherent two-dimensional structure of images, convolutional neural networks (CNNs) have become central to recent advancements in computer vision. Various neural network architectures have effectively integrated CNNs for improved outcomes \cite{he2015resnet,huang2018densnet,krizhevsky2014alexnet}. Among these, ResNet \cite{he2015resnet} introduced residual blocks specifically designed to mitigate the vanishing gradient issues, enabling the training of deeper neural networks. Additionally, U-Net \cite{ronneberger2015u} implemented skip connections to transfer of high-resolution features across network layers, thereby preserving detailed information that might otherwise be lost during training.

\section{Method}

\begin{figure*}[thb]
    \centering
    \includegraphics[width=0.8\textwidth]{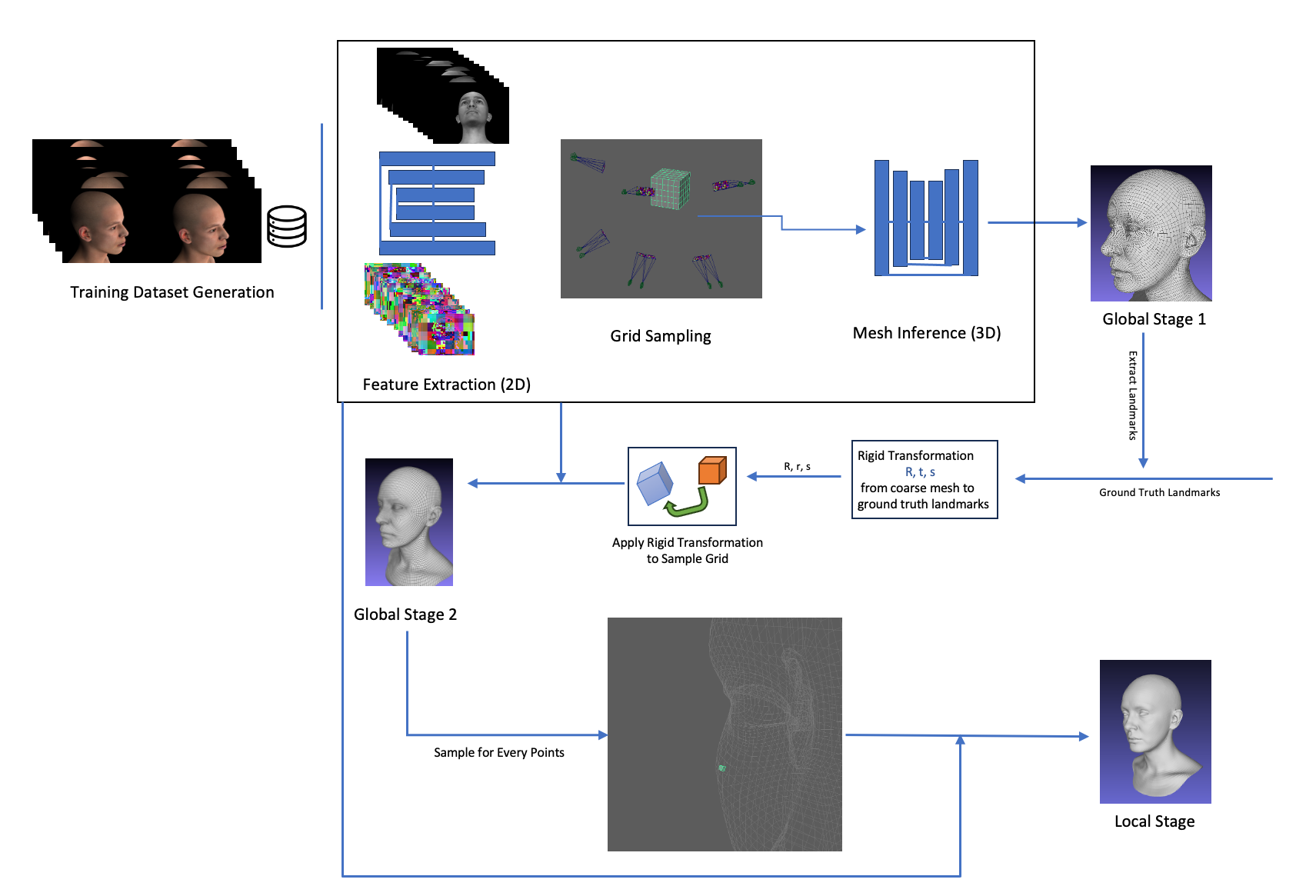}
    \caption{Overview}
    \label{fig:oveview}
\end{figure*}

The overall pipeline of the proposed model is illustrated in Figure \ref{fig:oveview}. The training dataset is initially generated using a proprietary Appearance 3DMM-based software, Visage Craft. Similar to the approach used by TEMPEH, the training process consists of three stages. The first stage, global stage one, predicts feature maps by sampling from grids projected on the feature maps. From this initial prediction, a set of approximate landmarks is extracted and passed to global stage two. A rigid transformation is applied to these rough predicted landmarks, allowing computation of rotation, translation, and scale relative to the ground truth landmarks. The spatial sampling grid is then transformed to align accurately with the true facial positions for mesh refinement. In the final refinement stage, a smaller grid is applied to each vertex, and local features are individually sampled from the feature maps, enabling precise adjustments to each vertex position. The core idea of our multiview deep learning approach consists of four main stages: (i) Generation of a synthetic, multiview, photorealistic training dataset using proprietary software called Visage Craft, which incorporates physically-based shading and an Appearance 3D Morphable Model (A3DMM). The A3DMM is built from scanned data of real faces along with known camera parameters. (ii) Since all the training data is retopologized, accurate landmark correspondences can be readily established between ground truth and predicted meshes. (iii) Creation of visibility masks from the synthetic dataset to indicate non-occluded facial regions, thus enabling dense, one-to-one correspondences across multiple viewpoints during training. (iv) Aggregation of information from multiview images through a neural network, facilitating precise prediction of the final 3D parametric facial model.

\subsection{Training Data} \label{sec:ldmks}

Achieving high-quality synthesis for a specific human face traditionally requires capturing extensive image datasets of the subject under strictly controlled conditions, often requiring significant budget and time. Generating semantically consistent facial meshes further complicates the process, given the intricate geometry of the human face. To address these challenges, we employ a variant of the 3D Morphable Model (3DMM), a widely-used approach based on Principal Component Analysis (PCA). The 3DMM represents novel faces through linear combinations of eigenvectors derived from representative facial data \cite{blanz1999morphable}. By manipulating a set of component weights, the model can effectively represent diverse facial geometries. However, existing methods commonly utilize meshes of relatively low resolution associated with a single diffuse texture map, often lacking detail. This limitation results in a significant loss of fine features and mesoscopic details. To this point, we integrate both shape and appearance (diffuse, normal, and gloss maps) into the PCA framework, an approach analogous to Active Appearance Models. Specifically, 48 retopologized face meshes, each paired with their corresponding diffuse, normal, and gloss texture maps, were acquired from the 3D Scan Store\footnote{https://www.3dscanstore.com/}. PCA was then individually applied to mesh geometry and each of the texture maps to build our comprehensive 3DMM. The eigenvectors and eigenvalues obtained from these analyses were subsequently leveraged to construct a detailed and diverse training dataset, laying the foundation for the subsequent stages of the proposed method.

Given a set of cameras $C_i$, with known intrinsic matrices $K_i$ and extrinsic matrices $[R_i | t_i]$, a series of multi-view stereo images $I_i$ from viewpoints $V_i$
can be rendered. In this work, we adopt the Cook-Torrance bidirectional reflectance distribution function (BRDF) model for shading and rendering of human faces \cite{cook-torrance}. Specifically, the Lambertian model is employed to compute the diffuse component. To accurately simulate specular reflections on the rendered faces, we utilize the GGX model for representing the normal distribution function, while the geometry shadowing function is modeled using the Schlick-Beckmann approximation. Additionally, Schlick's approximation is used to calculate the Fresnel reflectance factor \cite{walter2007microfacet}. The dataset generated through this rendering process is subsequently partitioned into two subsets: one for training and the other for validation purposes.

\begin{figure*}[htb]
    \centering
    \includegraphics[width=0.8\textwidth]{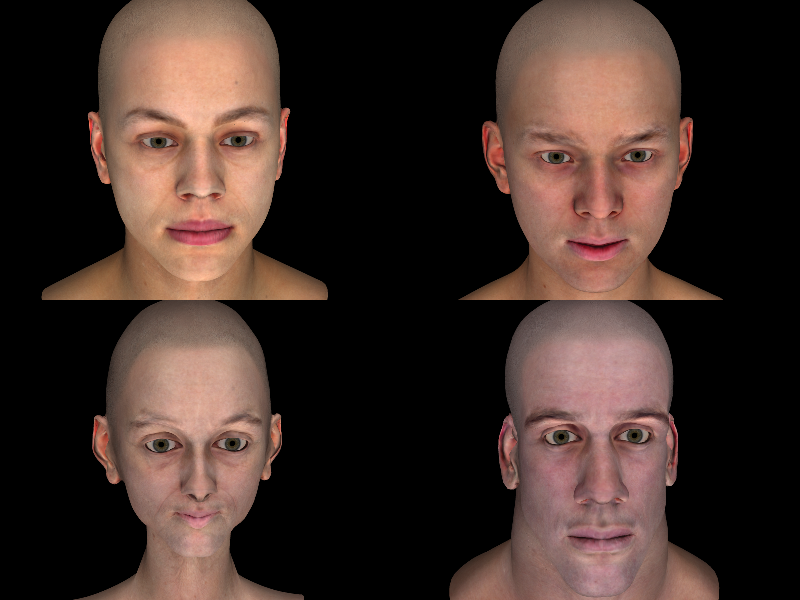}
    \caption{Face Samples: Randomly generated faces: upper row—normal sampling; bottom row—uniform sampling.}
    \label{fig:samples}
\end{figure*}

In the preparation stage of the face training dataset, the eigenvectors and eigenvalues derived from PCA are imported into the proprietary software Visage Craft (see Fig. \ref{fig:visage}). This software enables the generation of novel face meshes by randomly sampling parameters across each principal component. To ensure comprehensive and representative coverage of facial variations, the parameter sampling range is set to $\pm$3 standard deviations from the mean for each component, thus capturing approximately 98$\%$ of the potential shape variations within the generated dataset. Furthermore, a notable advantage of synthetic data generation is the ability to sample beyond typically observed shapes. To enhance the diversity of the dataset, normally sampled shapes are combined with uniformly sampled shapes. This process introduces atypical facial geometries that are rarely encountered in real-world data, thereby augmenting the training set and improving the model's generalization capabilities. Fig. \ref{fig:samples} illustrates examples of face meshes generated through both normal and uniform sampling methods.

\begin{figure}[htb]
    \centering
    \includegraphics[width=0.8\linewidth]{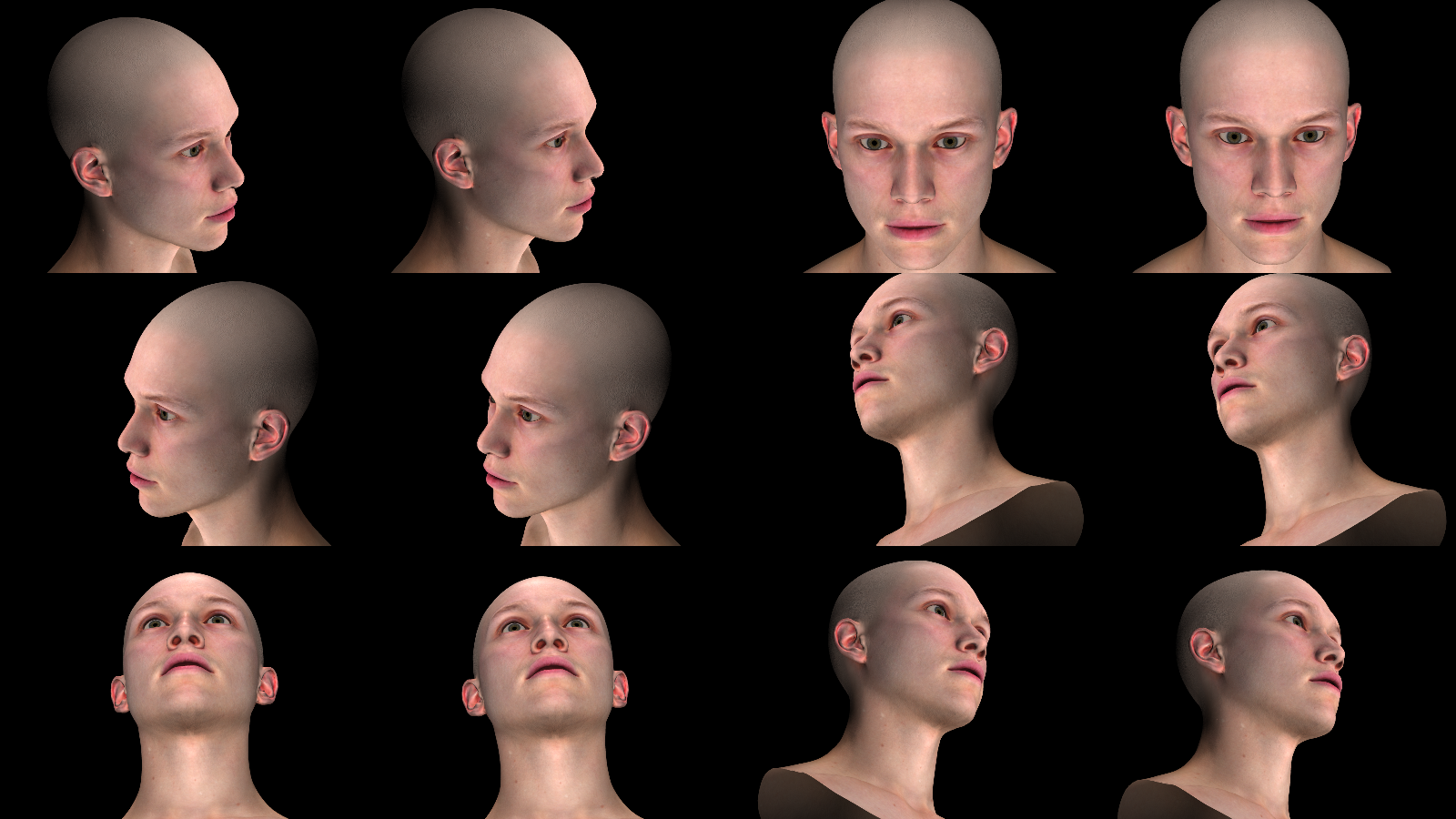}
    \caption{Multiview sample images generated by Visage Craft. There are six pairs of stereo images with known camera intrinsic and extrinsic matrices}
    \label{fig:facekit}
\end{figure}

Another benefit of using synthetic data is the ability to precisely control the rendering camera setup. In this study, a conventional multi-view stereo camera configuration is employed, comprising six pairs of stereo cameras strategically placed around each facial model. This setup enables rendering a total of 12 different viewpoints, effectively capturing diverse facial features from multiple perspectives. Figure \ref{fig:facekit} illustrates an example of synthetic faces generated for the training dataset.

\subsection{Neural Network Architecture}

Figure \ref{fig:oveview} illustrates an overview of the proposed end-to-end learning pipeline. The architecture draws inspiration from various prior works \cite{TMPEH:CVPR:2023, li2021tofu, Lombardi:2019, Lombardi:2018}, particularly the TEMPEH method. The training procedure is structured into three stages. In the initial stage, referred to as global stage 1, multi-view images are input into a convolutional neural network based on a ResNet34 \cite{he2015resnet} combined with a U-Net architecture \cite{ronneberger2015u}, employing shared parameters across views. This network extracts two-dimensional feature maps from each input image independently. Next, a 32-by-32 sampling grid is projected onto each view, and features are aggregated across views to form sampled feature representations. These aggregated features are then fed into a 3D U-Net \cite{iskakov2019learnabletriangulationhumanpose}, producing an initial rough prediction of the facial mesh. From this mesh, a set of landmarks is extracted, which, together with ground truth landmarks, is provided as input to global stage 2. In this second stage, a rigid transformation—including rotation R, translation t, and scale s—is computed by aligning the roughly predicted landmarks with their corresponding ground truth landmarks. Utilizing this computed transformation, the sampling grid is repositioned to more accurately align with the correct head location. This improved alignment enhances the accuracy of feature sampling. Finally, in the refinement stage, a smaller and more precise 8-by-8 grid is applied individually to each vertex of the mesh, enabling detailed vertex-wise adjustments. This approach results in a more refined and accurate final mesh representation. 

{\bfseries Sample Grid Generation:} TEMPEH adopts a cube-shaped sampling grid, where the width, height, and depth are equal in dimension. However, human head geometry more closely resembles a cylindrical structure, with the vertical axis (head height) typically exceeding both depth and width. When the neck region is included in the mesh, a cubic grid becomes unnecessarily large, resulting in significant unused space, particularly in the anterior and posterior regions. This inefficiency leads to wasted computational resources during training.

\begin{figure}[htb]
    \centering
    \includegraphics[width=0.4\linewidth]{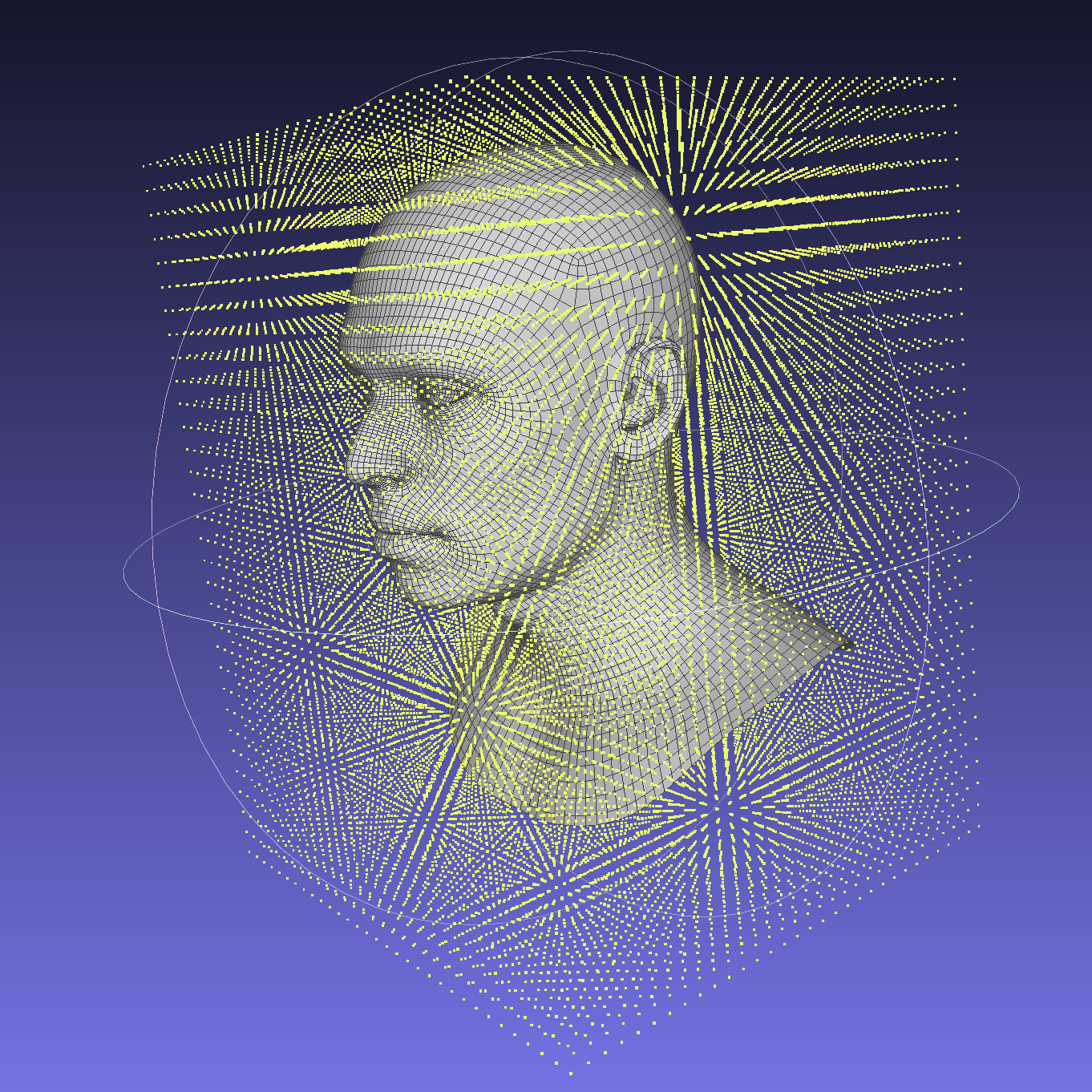}
    \caption{Sample grid derived from the dataset's bounding box}
    \label{fig:grid}
\end{figure}

To address this, the maximum bounding box encompassing all training samples is computed and used to define a more compact and representative sampling grid. As illustrated in Figure \ref{fig:grid}, the resulting grid deviates from a cubic form; instead, its dimensions are derived from the dataset itself, aligning more closely with the actual proportions of human head geometry. This nonuniform grid better accommodates the training data and improves sampling efficiency during learning.

{\bfseries Rigid Transformation on Landmarks:} ToFu \cite{li2021tofu} directly infers the head mesh from a fixed spatial grid. However, TEMPEH argues that since the head occupies only a small portion of the volumetric space, directly predicting the 3D head from the entire grid reduces the learning efficiency and lowers the reconstruction accuracy. To solve the problem, TEMPEH integrates a small neural spatial transformer to predict rotation, translation, and scale parameters for improved head localization. According to their analysis, incorporating a head localization network significantly enhances the mesh synthesis quality.

In this paper, the idea of applying a rigid transformation to the sampling grid to improve learning efficiency is maintained, but with modifications leveraging the 3DMM-generated meshes. Specifically: (i) the sampling grid is generated directly based on the training dataset. By computing a bounding box over the entire synthetic training database generated by Visage Craft, the 32-by-32 sample grid can be optimally determined, ensuring all faces are entirely enclosed within the capture volume. (ii) All meshes generated by Visage Craft are inherently retopologized, and one of the primary challenges associated with learning 3D meshes using multiview images is the difficulty in establishing surface point correspondences across these images \cite{Ramon, Lombardi:2019, Lombardi:2018, pengfei_mv, wu2019mvf}. Unlike previous methods requiring neural networks to predict the rigid transformation, this method leverages explicit landmarks extracted from meshes predicted in the global stage one. These landmarks are used to compute the rigid transformation directly against ground truth meshes, helping accurate alignment of the sample grid with face meshes.

\begin{figure}[htb]
    \centering
    \includegraphics[width=1\linewidth]{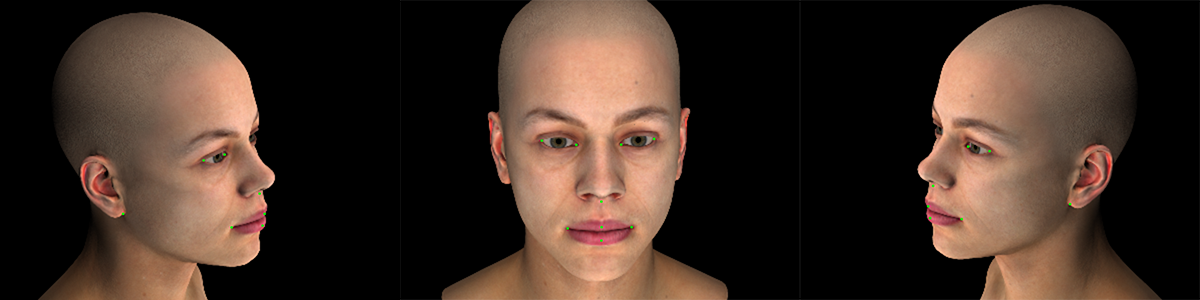}
    \caption{Landmarks projected onto three selected views, displaying only the visible points.}
    \label{fig:landmarks}
\end{figure}

An alternative method for obtaining landmarks involves using standard detection libraries (e.g., OpenCV or dlib). However, Xiang et al. \cite{xiang} found that variations in focal length (influencing perspective distortion) and viewing angles (related to head poses) negatively impact both traditional and deep learning-based facial landmark detection accuracy. Therefore, acquiring landmarks directly from ground-truth meshes avoids the distortion issues associated with standard landmark detection methods. Additionally, landmark selection benefits from identifying clearly distinguishable object boundary connectivities, particularly "T" junctions, as suggested by Cootes et al. \cite{cootes2000introduction}. Ambiguous regions like jawlines complicate facial outline definition. For clarity and consistency, 11 distinct landmarks are selected, including two outer eye corners, two inner eye corners, two mouth corners, two midpoints along the upper and lower lips, two earlobe junctions, and the philtrum. These points are utilized for precise rigid alignment computations. Fig. \ref{fig:landmarks} illustrates an example set of landmarks projected back onto multi-view images.

{\bfseries View Feature Vectors and Fusion:} The multi-view deep learning regressor faces two primary challenges: (i) accurately mapping multiview images onto a single 3D facial mesh, since neural networks often have difficulty determining which mesh regions are visible from specific viewpoints; and (ii) establishing dense correspondences across multiple views, which remains a complex task for neural networks. Prior approaches, such as those proposed by \cite{pengfei_mv, richardson20163d}, tackle the first challenge by employing predefined visibility masks applied uniformly to all faces. However, this uniform application neglects potential individual variations in visibility, which can negatively impact the learning process. For the second challenge, methods introduced by \cite{li2021tofu, TMPEH:CVPR:2023} project a grid onto each image, enabling the neural network to learn correspondences from sampled grid points.

Following ToFu and TEMPEH, during the refinement stage, each mesh vertex is sampled across 12 different 2D feature maps through perspective projection and bilinear sampling. While ToFu uniformly aggregates sampled features across views without considering visibility, TEMPEH improves upon this by computing visibility masks based on the dot product between surface normals and camera viewing directions. Points with positive dot product values are classified as visible from the corresponding viewpoint. These computed visibility masks thus serve as weights for surface-aware feature aggregation. However, TEMPEH uses raw scans as supervisions for training, which lack consistent topology in the final predictions, requiring visibility masks to be recalculated at each training iteration.

\begin{figure}[htb]
    \centering
    \includegraphics[width=1\linewidth]{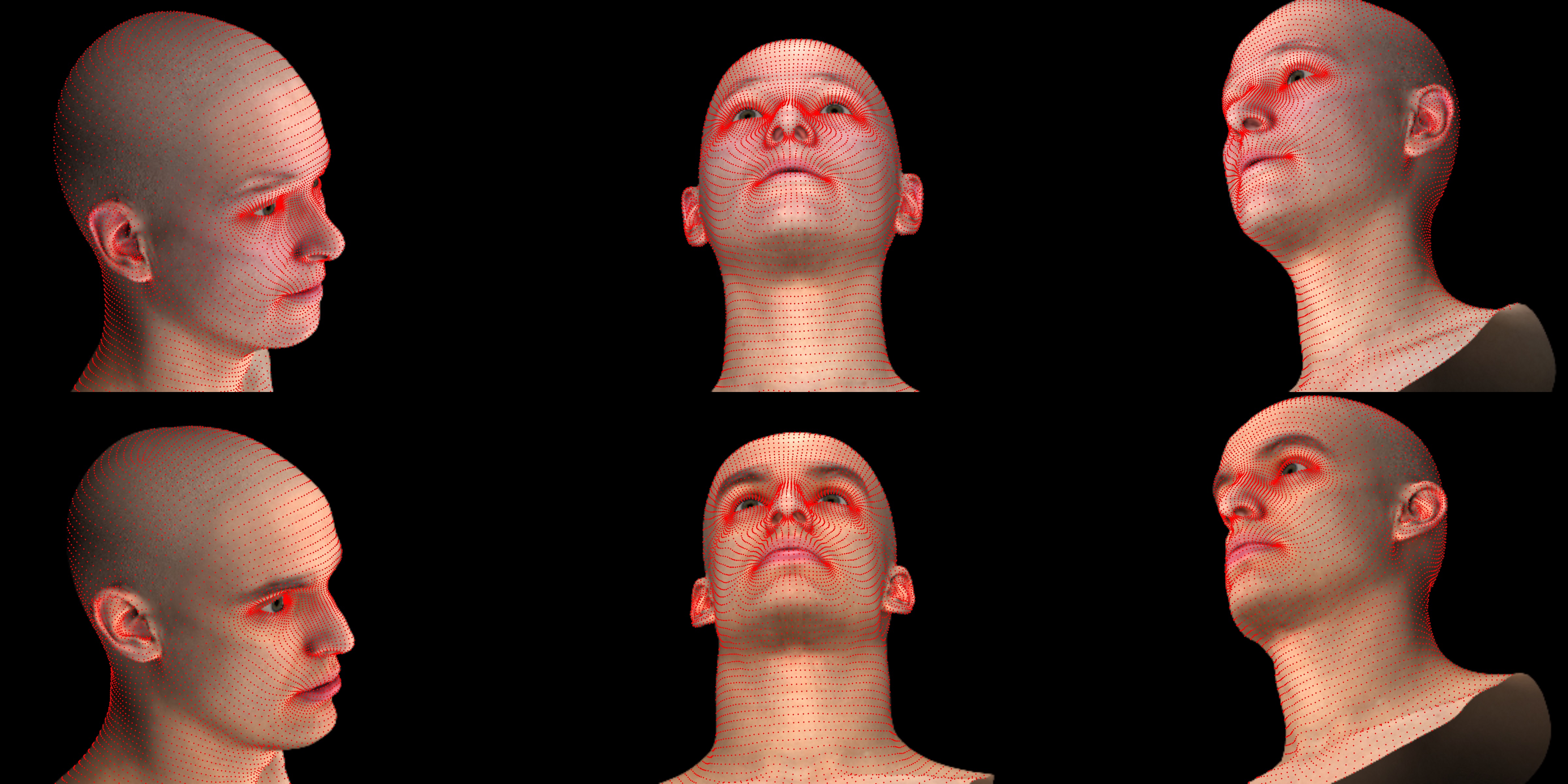}
    \caption{Visibility Masks Project Back to Images}
    \label{fig:visiMask}
\end{figure}

Our proposed approach combines and refines these methods by integrating visibility masks into the feature aggregation process. A central aspect of our method is leveraging retopologized meshes generated using the 3D Morphable Model (3DMM) algorithm implemented within Visage Craft. Given the uniform topology across all generated meshes and known camera parameters, visibility masks can be efficiently computed for each viewpoint. Specifically, applying Z-buffer testing with known camera matrices and meshes from Global Stage 2 enables precise determination of mesh vertex visibility for every view. These visibility masks serve as precomputed weights indicating surface visibility, similar to the approach used by TEMPEH, but without needing additional computations during the training process. To further accelerate visibility mask generation, a multi-threading library in Python is utilized, reducing the computation time to approximately 2–4 hours for an 8000-sample dataset across all 12 views. Figure \ref{fig:visiMask} demonstrates two examples of synthetic facial meshes with mesh vertices back-projected onto three selected views, clearly showing the applied visibility maps.

{\bfseries Loss Functions:}
In contrast to TEMPEH, which uses raw scans for supervision and thus incorporates surface distance and surface regularization losses to ensure close alignment with raw data and to compensate poor registrations due to overlaps. Because our method employs meshes with a consistent and clean topology for training, only a simple point-to-point distance loss is required for supervision. However, to enforce the sampled grid in our second stage to follow rigid transformations aligned with ground-truth landmarks, we introduce an additional landmark-based loss. This loss term ensures the final predicted meshes closely translate and align with the ground-truth meshes, improving overall reconstruction accuracy and alignment precision. The point loss $L_p$ is defined as the mean squared error between the predicted mesh vertices and the corresponding ground-truth vertices:
\[
L_p = \frac{1}{n} \sum_{j=1}^{n} \left( p_j - \hat{p}_j \right)^2
\]

where $p_i$ represents the predicted vertex positions, and $\hat{p}_i$ denotes the ground-truth vertex positions. Similarly, given the predicted landmarks $lmk_i$ and the corresponding ground-truth landmarks $\hat{lmk}_i$ the landmark loss $L_lmk$ is defined as:
\[
L_{lmk}= \frac{1}{n} \sum_{k=1}^{n} \left( lmk_k - \hat{lmk}_k \right)^2
\]

The final loss function used in training is formulated as a weighted combination of the vertex position loss and the landmark position loss:
\[
L = w_p L_p +  w_{lmk}L_{lmk}
\]

where $w_p$ and $w_{lmk}$ represent the weighting factors assigned to the vertex and landmark losses, respectively. These weights control the relative importance of each term during training. Notably, during Global Stage 1 and the refinement stage, the landmark loss weight $w_{lmk}$ is set to 0, whereas Global Stage 2, $w_{lmk}$ is set to 1.

In summary, while the overall architecture of the proposed network draws inspiration from TEMPEH, several important differences distinguish it. First, instead of using a uniform cubic sampling grid, a head-shaped bounding box grid is adopted to better match the anatomical structure of the human head and reduce wasted space during training. Second, landmarks are integrated not only to compute a rigid transformation that aligns the sampling grid with the facial region but also as a regularization component within the loss function, improving geometric accuracy. Lastly, visibility masks and view-specific feature vectors are derived from the predicted coarse meshes themselves, allowing for more targeted refinement in the later stages.

\subsection{Testing Data}
\begin{figure}[htb]
    \centering
    \includegraphics[width=1\linewidth]{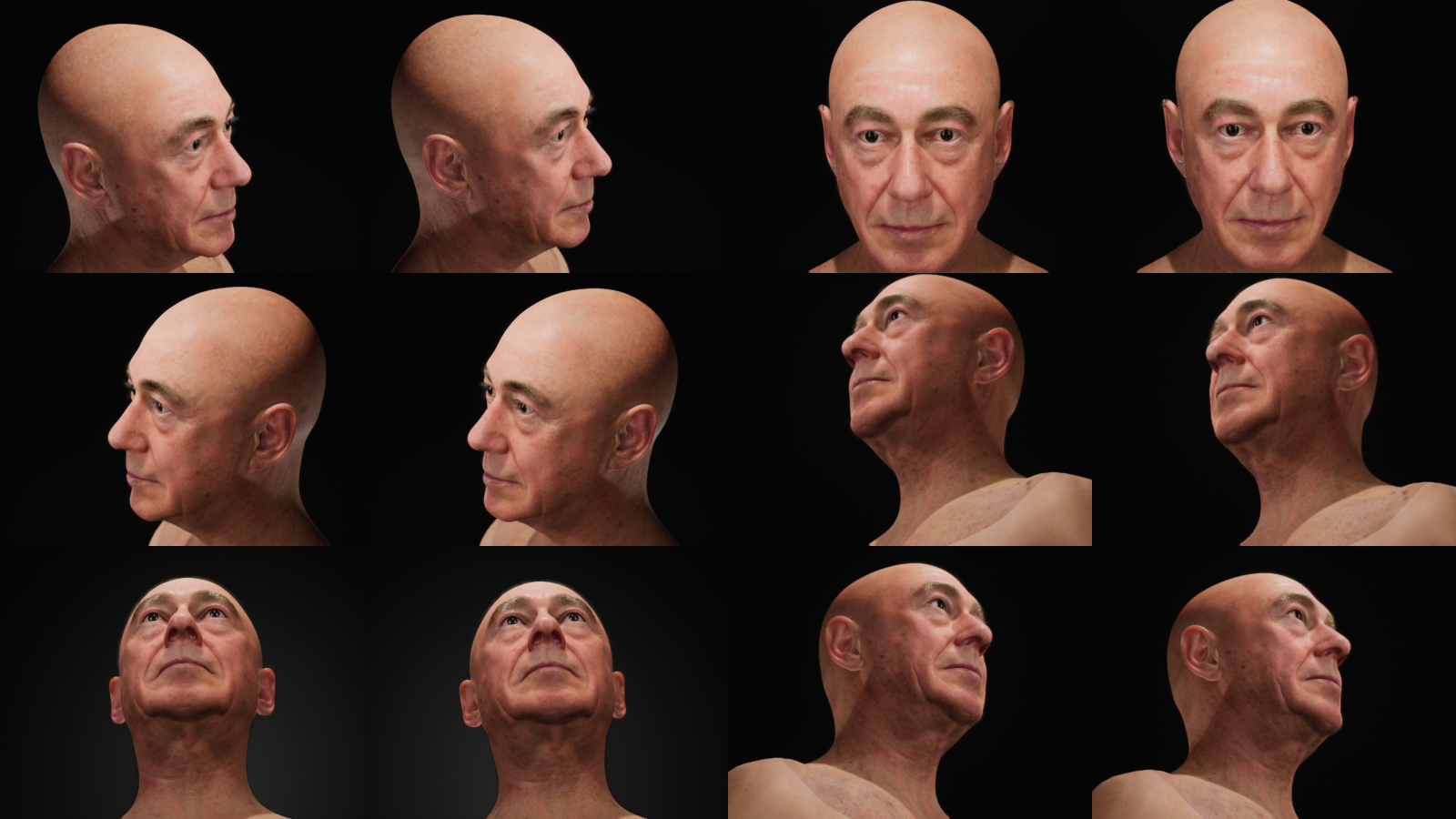}
    \caption{Metahuman in Unreal Engine with same camera setups}
    \label{fig:ue}
\end{figure}

The goal of this study is not only to evaluate the proposed model on synthetic data but also to demonstrate its practical effectiveness in reconstructing meshes from images of real human faces. One possible method for achieving this involves capturing photographs of actual subjects using camera configurations identical to those used in Visage Craft. Alternatively, to accelerate the development process, an approach based on generating highly realistic facial images using Metahuman in Unreal Engine was explored. Specifically, the 3D Scan Store provides a recent collection of facial scans designed explicitly for compatibility with Metahuman, including albedo, normal, and cavity maps. A subset of these facial meshes, distinct from those included in the Visage Craft dataset, was carefully selected for testing. Following a workflow outlined in a tutorial\footnote{https://www.3dscanstore.com/blog/Tutorials/Metahuman-Identity-Tutorial\label{link}}, the selected Scan Store meshes were imported into Metahuman Creator, where additional facial details such as hair and eyebrows were applied to enhance realism. For rendering purposes, the meshes were imported into the Metahuman Lighting environment. Test images were rendered using the albedo, normal, and cavity maps provided by the Scan Store. To further enhance realism, normal wrinkle maps were manually refined in Photoshop. Comprehensive instructions detailing these procedures can be found in the referenced tutorial video\ref{link}. Importantly, for consistency between training and testing phases, the Metahuman models were positioned similarly to the synthetic faces used in Visage Craft, identical camera settings—including both intrinsic and extrinsic camera parameters were also used. Figure \ref{fig:ue} presents an illustrative example of a facial rendering generated using Metahuman in Unreal Engine. 

\begin{figure}[htb]
    \centering
    \includegraphics[width=0.6\linewidth]{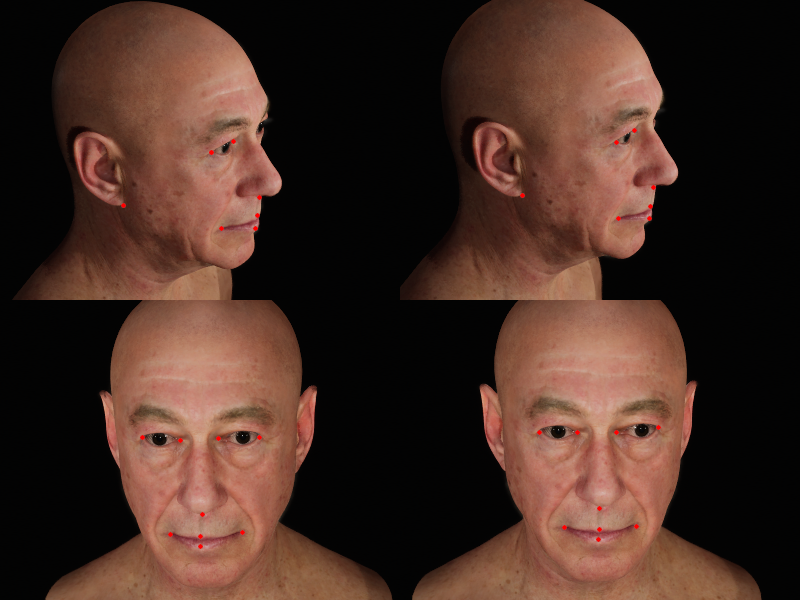}
    \caption{Manually Labeled Visible Landmarks}
    \label{fig:svd}
\end{figure}

Furthermore, the global state 2 requires ground truth landmarks to compute the rigid transformation. However, in the testing dataset, we assume that meshes with the same retopology are unavailable, needing the neural network to predict a mesh with an identical topology. Inspired by the method proposed by \cite{dias}, which uses facial landmarks detected from 2D images to refine the mesh predictions post-processing, we similarly incorporate landmarks into our approach. However, as previously discussed in Section \ref{sec:ldmks}, variations in focal length and viewing angles significantly affect the accuracy of both traditional and deep learning-based facial landmark detection methods. To solve this issue, particularly due to the limited number of landmarks (11 points), we manually annotated landmarks on each face across all 12 views. For each view, only the landmarks clearly visible from that particular viewpoint were recorded. To reconstruct a single 3D point in homogeneous coordinates, at least two corresponding viewpoints are required. In our experimental setup, each view forms part of at least one stereo pair, thus ensuring sufficient information to accurately determine all landmark positions through Singular Value Decomposition (SVD). Figure \ref{fig:svd} illustrates an example of manually annotated landmarks projected onto the images. The derived 3D landmark coordinates can subsequently be used in Global Stage 2 to calculate the rigid transformation for mesh alignment.

\section{Implementation Details}

{\bfseries Visage Craft: } We utilized 48 face meshes along with their corresponding diffuse, normal, and gloss maps to construct an Appearance Morphable Model. Using this model, we generated a dataset comprising 10,000 synthetic faces. Half of these faces were sampled following a normal distribution within a range of $\pm$3 standard deviations, while the other half were sampled uniformly to increase dataset diversity. Thereafter, the dataset was partitioned into 7,500 faces for training and 2,500 faces reserved for validation. All the meshes contain 12466 points.

{\bfseries Test Data: } The test dataset consists of Metahuman Identities acquired from the 3D Scan Store. Following the workflow described in a tutorial provided on their website, all identities were converted into Metahuman models. Subsequently, these Metahuman models were rendered using Metahuman lighting, maintaining consistency with the camera parameters and lighting configurations employed in Visage Craft.

{\bfseries Hyper Parameters: }  We adopted most of the neural network architectures from TEMPEH, with specific modifications to accommodate our task. Twelve grayscale stereo images were provided as input to the network. Training was conducted sequentially: Global Stage 1 was trained for 150,000 iterations using a learning rate of 1\textit{e}-3, followed by Global Stage 2 for an additional 150,000 iterations at a reduced learning rate of 1\textit{e}-4. Finally, the Refinement Stage underwent training for another 150,000 iterations, also with a learning rate of 1\textit{e}-4. All training stages were executed on two NVIDIA A100 GPUs.

\section{Evaluation}
\begin{figure}[htb] \centering \includegraphics[width=0.6\linewidth]{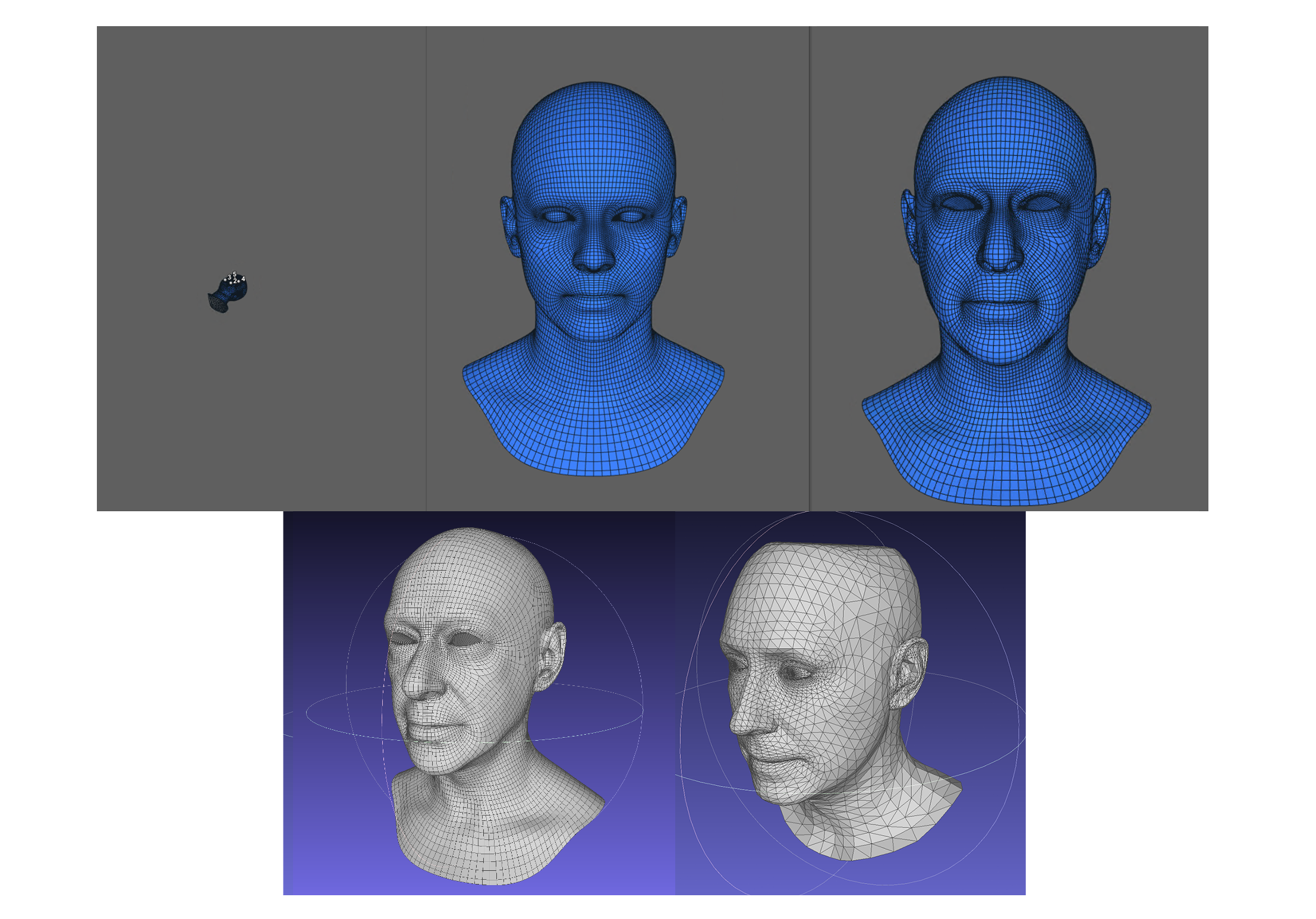} \caption{Subjective Compare} \label{fig:compare} \end{figure}
{\bfseries Subject Visualization and Post Processing: } In movie production, face meshes are retopologized into quad meshes to ensure smooth deformation during animation. A clean quad topology facilitates facial rigging and blendshape creation, making it more efficient to implement expressive movements. TEMPEH, on the other hand, supervises learning using raw scans, resulting in triangulated final meshes that lack one-to-one correspondences. Consequently, additional human labor or extra processing steps are required to clean up the meshes for production use. In contrast, our training dataset is generated using a 3D Morphable Model (3DMM) with a standardized mesh topology. As a result, all predicted meshes maintain a consistent structure with strict one-to-one correspondences, eliminating the need for manual cleanup. Furthermore, while TEMPEH relies on a separate neural network to predict rigid transformations for grid alignment, its final output often exhibits variations in scale and translation. In our approach, however, the sample grid is computed based on the bounding box derived from the training dataset, and the rigid transformations in the second stage are directly estimated from the landmarks predicted in Global Stage 1. This ensures that the predicted meshes are naturally centered within world coordinates and accurately reflect real-world head dimensions. Figure \ref{fig:compare} visually demonstrates this comparison. Top Row: TEMPEH requires predicting rotation, translation, and scale, resulting in meshes that are not scaled accurately to actual head sizes. The left image illustrates TEMPEH's predicted meshes shifted significantly within the coordinate system compared to the mean face (middle). In contrast, our method (right) produces meshes consistent with real human head dimensions, as our training data is derived from accurately sized head models. Bottom Row: Meshes predicted by TEMPEH are triangulated due to supervision from raw scans, which lack consistent topologies. Our approach directly generates meshes with uniform, rigid topology, as illustrated.

\begin{figure}[!p]
    \centering
    \includegraphics[width=0.7\linewidth]{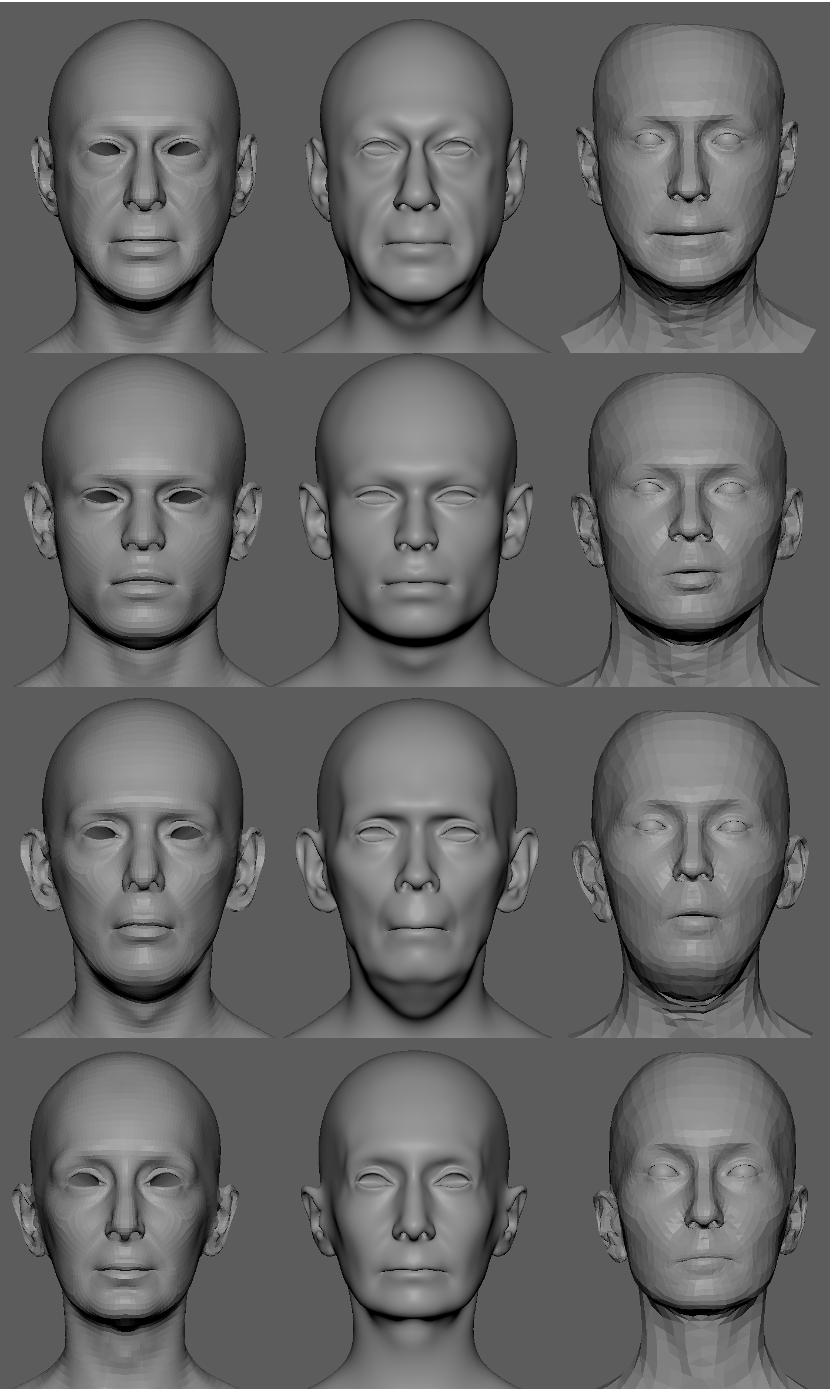}
    \caption{Visual Comparisons: Our method’s output (left), the ground-truth mesh (center), and the result from TEMPEH (right)}
    \label{fig:predcompare}
\end{figure}

To evaluate the visual accuracy of the reconstructed facial meshes, Figure \ref{fig:predcompare} shows a side-by-side comparison of three versions: our method’s output (left), the ground-truth mesh (center), and the result from TEMPEH (right). The mesh produced by our approach closely mirrors the subject’s facial structure seen in the test data, demonstrating the model’s ability to generate detailed, identity-preserving geometry directly from synthetic supervision. The strong visual agreement between our predictions and the ground-truth meshes highlights synthetic data's effectiveness in driving accurate and subject-specific facial synthesis.

\clearpage
\clearpage

\begin{figure}[!p]
    \centering
    \includegraphics[width=0.65\linewidth]{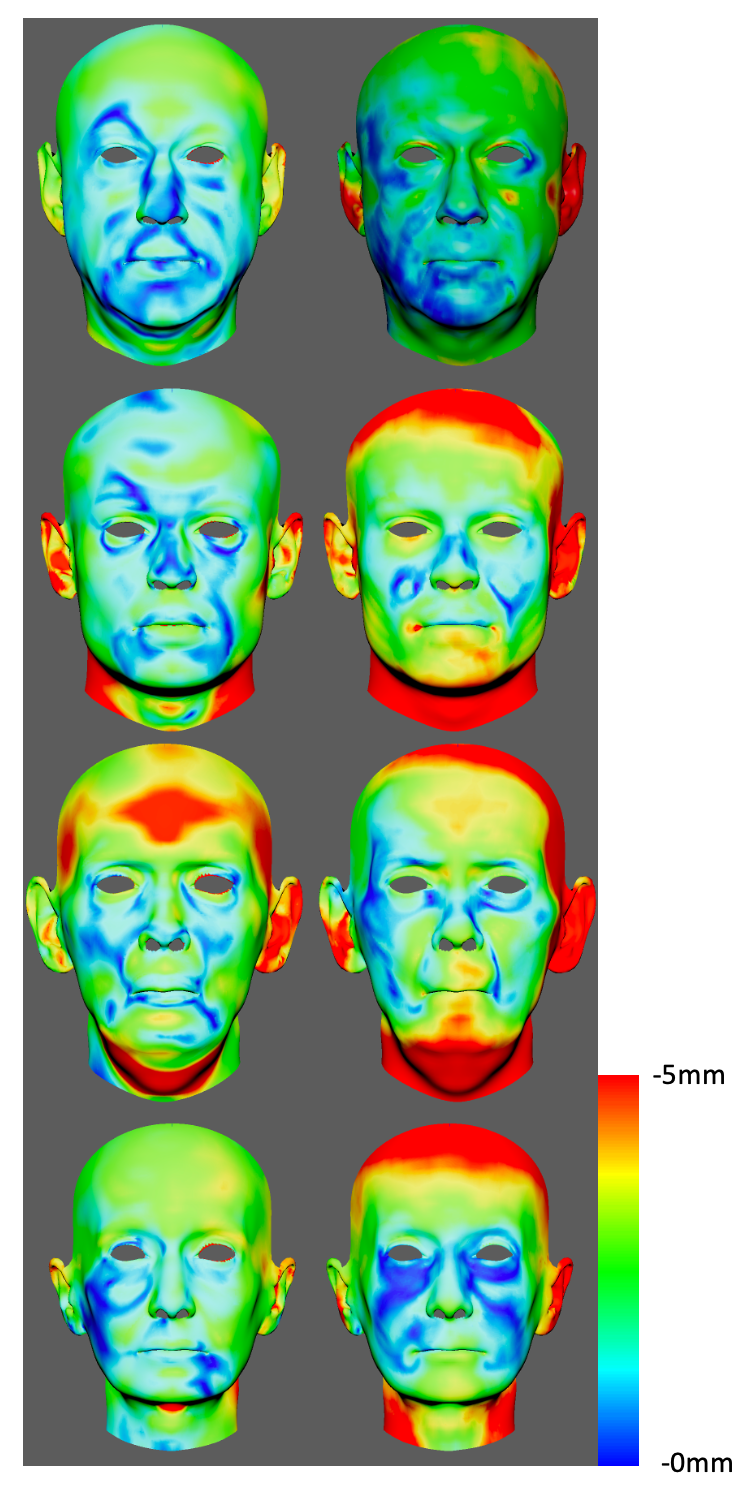}
    \caption{Heatmap comparison}
    \label{fig:heatmap}
\end{figure}

\newpage
{\bfseries Quantitative evaluation: } We have experimented with many published neural network architectures, particularly those designed to reconstruct facial parameters from multi-view images. Among these architectures, TEMPEH has demonstrated the most stable performance and yielded the best overall results. While other architectures may also successfully reconstruct facial meshes, they often come with significant restrictions or requirements, such as precise lighting conditions, big training datasets for 3D Morphable Models (3DMM), specialized datasets, or extensive hyperparameter tuning. The primary objective of this research is not to compete directly with existing facial synthesis neural networks but rather to investigate the feasibility of obtaining production-ready retopologized meshes solely through synthetic training data. However, we still want to perform a comparison with TEMPEH to assess the visual quality and accuracy achievable with our synthetic dataset. To get a meaningful comparison, the triangulated meshes generated by TEMPEH were manually wrapped onto our standardized mesh topology and then subjected to rigid transformations (translation, rotation, and scaling) to match the ground truth. As the focus of this study is limited to the facial region, only the corresponding portion of the mesh has been extracted for comparison. As depicted in Figure \ref{fig:heatmap}, our proposed method successfully produces retopologized meshes that closely resemble the subject shown in the input images, particularly in the frontal facial region. Reconstruction errors within the facial region remain under 5mm. However, TEMPEH demonstrates higher accuracy in high-frequency areas, such as wrinkles and the ear contours. Additionally, discrepancies observed in the lower head region stem primarily from differences in the distributions between the ground truth meshes obtained from Metahuman and the training dataset used to construct our Appearance 3D Morphable Model. To reiterate, the purpose of this research is not to benchmark or surpass other neural network architectures, but rather to explore and validate the potential of leveraging synthetic data to reliably generate consistent, retopologized facial meshes suitable for production pipelines. Although TEMPEH achieves better results, it is heavily based on professional capture equipment, costly lighting setups, and substantial human participation for dataset acquisition. In contrast, our approach simplifies this requirement, needing only the acquisition of retopologized meshes to construct the Appearance 3D Morphable Model.
\clearpage
\clearpage

\begin{figure}[!htb]
    \centering
    \includegraphics[width=1\linewidth]{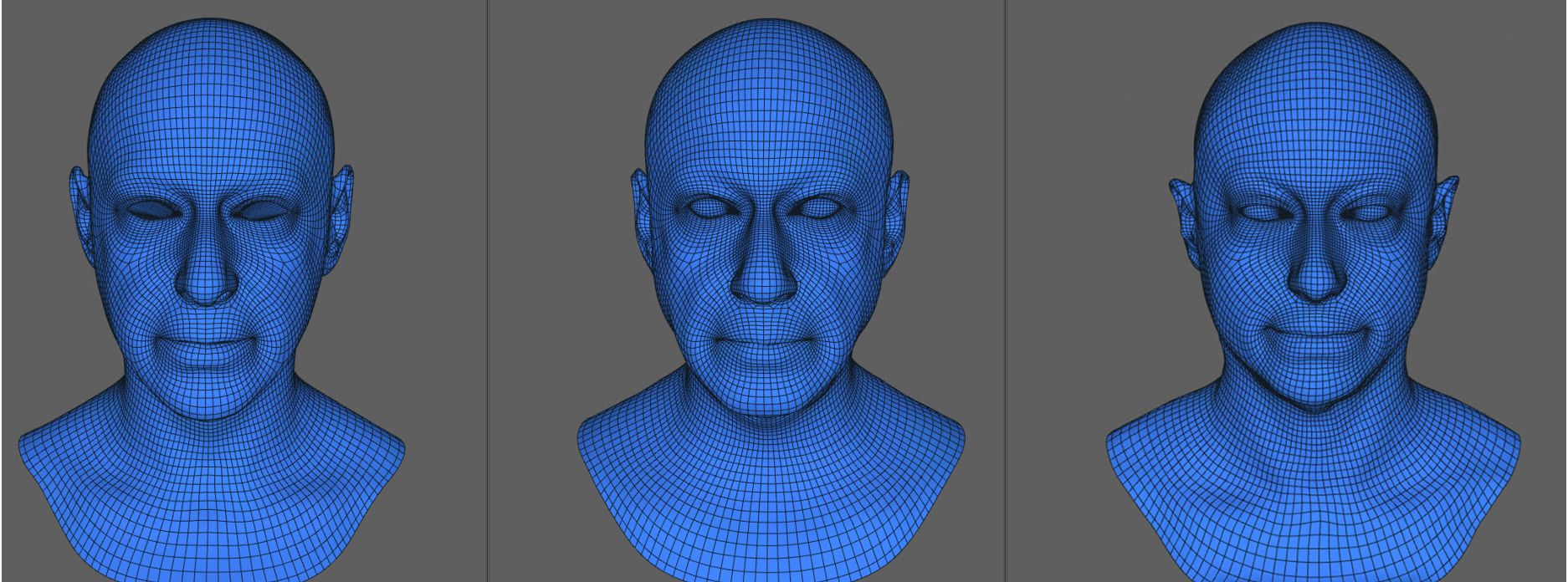}
    \caption{Effect of sampling grid shape on mesh synthesis. Left: Bounding box–based grid; Middle: Ground truth mesh; Right: Cubic grid. The bounding box grid produces a more accurate and realistic facial geometry.}
    \label{fig:ablationGrid}
\end{figure}

{\bfseries Ablation experiments:} To quantify the contribution of specific design decisions, we conducted a series of controlled comparisons:

1) Sampling Grid Shape: While TEMPEH employs a cube-shaped sampling grid, our approach utilizes a bounding box–based grid derived from the dimensions of the training data. To evaluate the impact of grid geometry, we trained identical models at the Coarse 1 stage using both the cubic grid and our bounding box–based alternative. As shown in Figure \ref{fig:ablationGrid}, the use of a cubic grid results in visibly distorted facial geometry, producing a "chubby" appearance. In contrast, the bounding box grid yields a mesh that more closely conforms to the subject's facial structure and better matches the ground truth.

2) Landmark Guidance: Unlike TEMPEH, which applies a neural network to estimate the rigid transformation of the sampling grid for improved computational efficiency, our method employs a landmark-based rigid alignment. Specifically, rigid transformations are computed using landmarks extracted during the Global Stage 1 and aligned with corresponding ground-truth landmarks.

\begin{figure}[htb]
    \centering
    \includegraphics[width=0.5\linewidth]{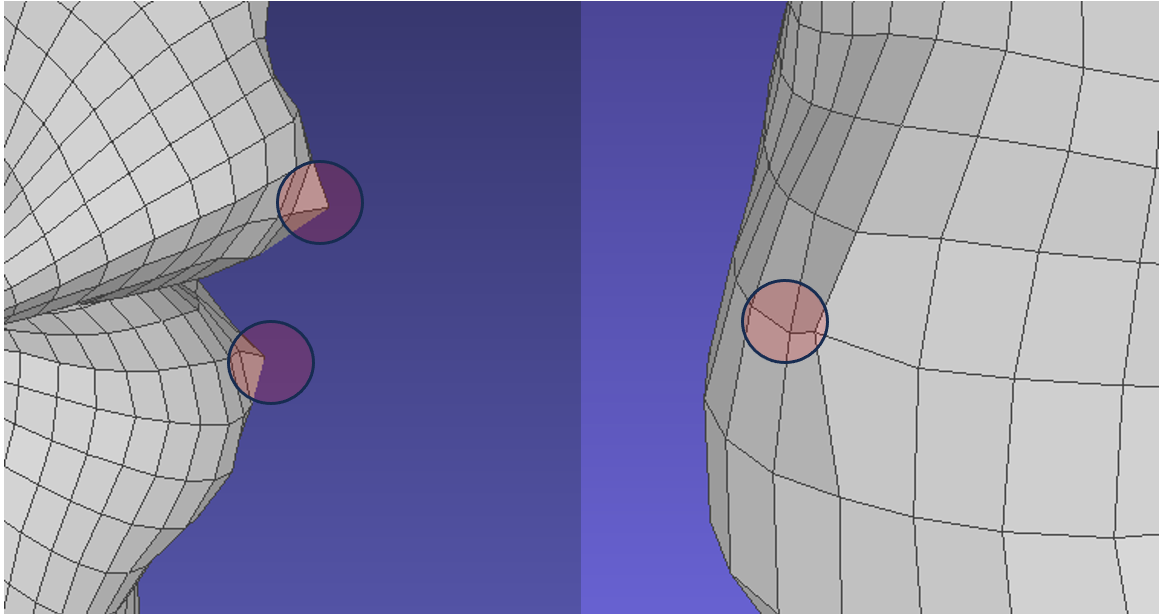}
    \caption{Weighted Points on Landmarks Shows Artifaces}
    \label{fig:lmkartifaces}
\end{figure}

There are two principal strategies for incorporating landmark information during training: 1. assigning greater weights to the landmark points within the point-based synthesis loss, as expressed in Equation \ref{eq:lmk_points}; or 2. introducing a separate landmark loss term to regularize the overall objective, as shown in Equation \ref{eq:lmk_term}. As demonstrated in Figure \ref{fig:lmkartifaces}, the first approach—using Equation \ref{eq:lmk_points}—can produce undesirable surface artifacts, such as visible dents and spikes around landmark regions, likely resulting from localized overfitting. To solve this issue, the formulation in Equation \ref{eq:lmk_term} is adopted in our final loss function.

\begin{align}  
   L_p &= \frac{1}{n} \sum_{j=1}^{n} \omega_i\left( p_j - \hat{p}_j \right)^2 \label{eq:lmk_points} \\
   L &= w_p L_p +  w_{lmk}L_{lmk} \label{eq:lmk_term}
\end{align}

\begin{figure}[htb]
    \centering
    \includegraphics[width=0.35\linewidth]{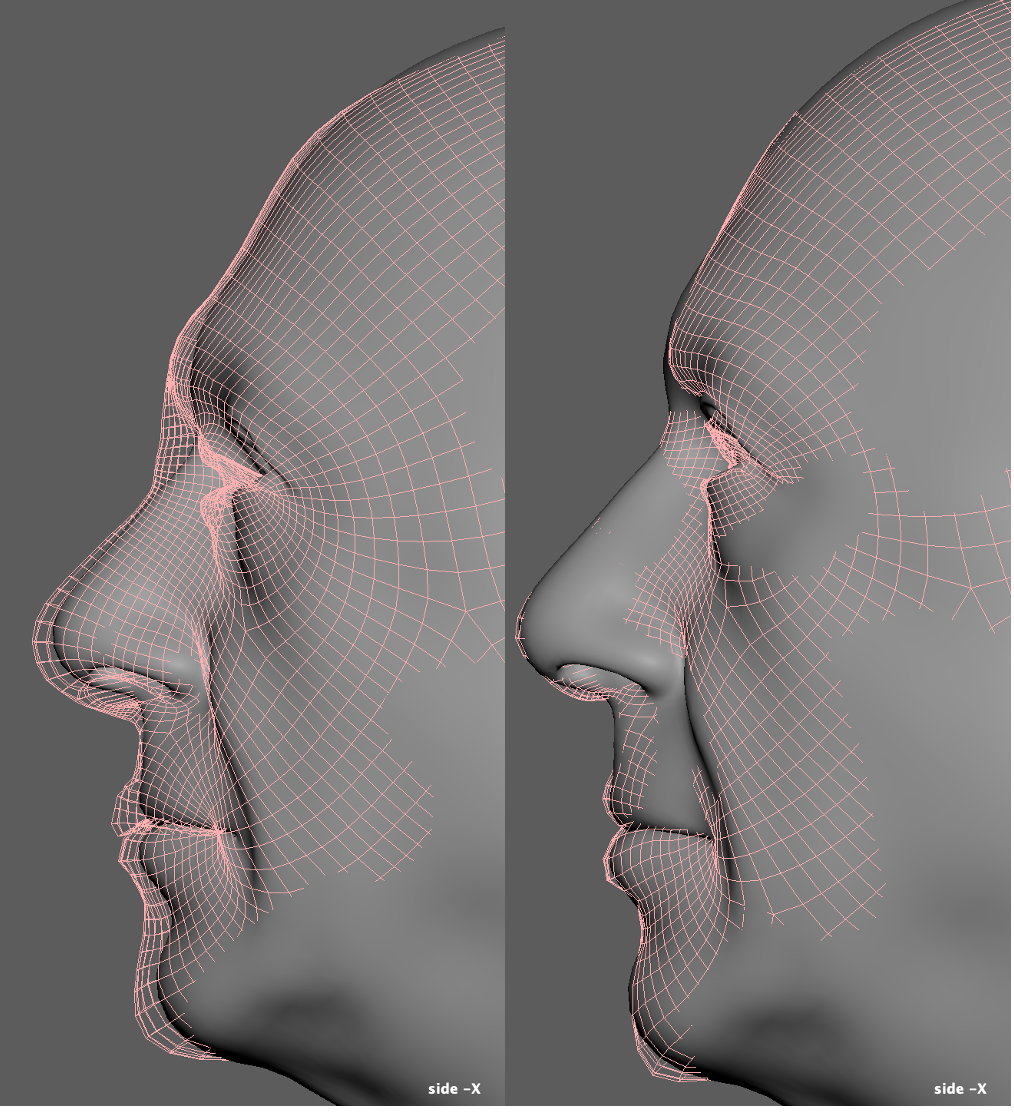}
    \caption{Effect of landmark-based rigid alignment. Left: Prediction without landmark alignment. Right: Prediction with landmark alignment.}
    \label{fig:lmkCompare}
\end{figure}

Figure \ref{fig:lmkCompare} illustrates the effectiveness of incorporating rigid alignment based on landmarks compared to a model trained without this alignment step. In the visualization, the red mesh represents the predicted output, while the gray mesh corresponds to the ground truth. When the landmark-based transformation is applied, the predicted meshes show improved positional correspondence with the ground-truth meshes, indicating enhanced spatial consistency.

\newpage
\section{Limitations}
The main limitation of this work lies in the small and narrow training dataset. Only 48 faces were used, each with identical mesh topology (12,466 vertices) and 1024 $\times$ 1024 texture resolution. This limited sample size makes it difficult for the model to learn the full range of variation found on real human faces. As a result, high-frequency details—such as fine wrinkles and skin texture—are often lost, producing overly smooth reconstructions. With so few examples, the network tends to regress toward the mean.

\begin{figure}[htb]
    \centering
    \includegraphics[width=0.5\linewidth]{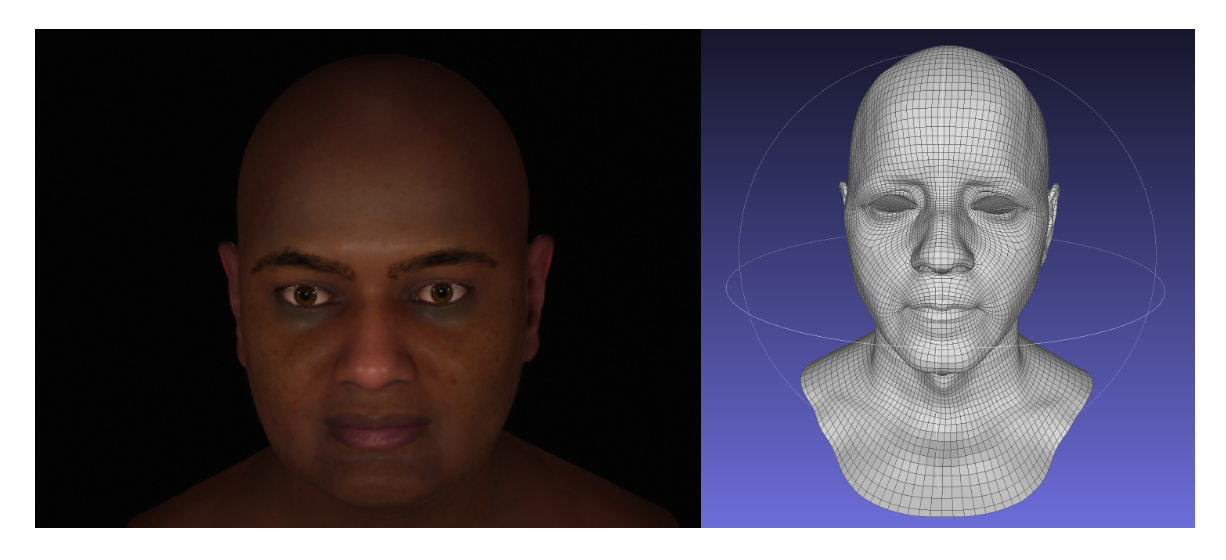}
    \caption{Example of poor mesh prediction for a dark-skinned face. The limited number of eigenvectors in the training set prevents the network from accurately reconstructing underrepresented skin tones.}
    \label{fig:limits}
\end{figure}

Texture synthesis is also affected. Of the 48 textures, only one or two represent darker skin tones. Since texture generation relies on linear combinations of this small set, the network struggles to reconstruct faces that fall outside the dominant appearance range. Faces with very light or very dark tones are often poorly synthesized, as the training data lack enough variation to support them. Figure \ref{fig:limits} shows an example where the network fails to generalize a facial mesh for an individual with a darker skin tone.

All of these issues point to the same cause: a lack of diversity in the training data. To improve, the data set needs to grow, not just in size, but in variety. It should include more faces, more textures, and a better spread of skin tones and features.

\section{Conclusion and Discussion} This chapter introduced a pipeline for predicting retopologized 3D head meshes using a fully synthetic dataset. Traditional deep learning pipelines for facial synthesis often require complex and expensive capture equipment, extensive human involvement, and significant data acquisition costs. By contrast, our approach leverages a relatively small set of publicly available retopologized meshes to construct an Appearance 3D Morphable Model, thereby enabling the efficient generation of large-scale synthetic training data. This strategy significantly reduces budget requirements and provides an accessible alternative for smaller studios or research laboratories, allowing them to train effective neural networks without the need for substantial investments in equipment or participant recruitment.

Future improvements of the proposed pipeline may focus on enhancing the Visage Craft system, specifically by building a more comprehensive 3DMM and achieving more realistic facial renderings. Currently, Visage Craft incorporates only 48 faces, limiting its ability to represent the full variability in human facial structures. Expanding this data set with additional facial scans could help cover a broader range of head shapes and appearances. Furthermore, the present implementation utilizes the Cook-Torrance Bidirectional Reflectance Distribution Function (BRDF) model for shading and rendering, which may be improved with more sophisticated and realistic rendering techniques. In particular, Unreal Engine’s Metahuman framework offers highly realistic rendering capabilities combined with detailed facial expression controls, which could further enhance synthetic dataset realism and improve the quality of model predictions.

%% file: conclusions.tex
\chapter{Conclusions and Discussion}
This dissertation examined the convergence of traditional computer graphics pipelines and contemporary deep learning methodologies for 3D facial synthesis. It introduced a novel end-to-end framework for producing high-fidelity, animation-ready facial meshes through multi-view neural regression. The study began by establishing a theoretical foundation in facial analysis, geometric modeling, and photorealistic rendering—contextualizing the technical and perceptual challenges inherent in digital avatar creation. A detailed review of prior work underscored the constraints of classical 3D Morphable Models (3DMMs), the advancements made possible through Light Stage systems, and the growing potential of neural networks in capturing complex facial geometry with improved realism and efficiency.

\section{Discussion of Dissertation Work}
The technical contributions of this dissertation are organized into four principal components. First, traditional 3D mesh acquisition workflows—particularly those involving retopologized meshes and light stage-based capture—were revisited to identify persistent bottlenecks in geometry acquisition and retopology. The Variable Illumination Sphere (VarIS), developed in our lab, leverages variable lighting and stereo reconstruction techniques to produce high-quality 3D facial meshes following conventional methods. Although effective, traditional workflows demand substantial manual effort, time, and financial resources. In contrast, recent advances in deep learning offer a promising alternative—enabling the direct synthesis of detailed facial meshes from images and significantly reducing both development time and associated costs. However, training such models requires large-scale datasets with fully retopologized meshes. Existing publicly available 3D Morphable Models (3DMMs) often lack physically accurate rendering capabilities and offer limited extensibility. To bridge this gap, the Visage Craft system was developed—an Appearance 3DMM platform with physically based rendering support. This tool not only enables the generation of high-fidelity synthetic datasets but can also be extended with additional features such as known camera intrinsics and extrinsics, lighting parameters, and mesh correspondences, providing a robust foundation for training and validating neural networks.

Second, a focused investigation was conducted on the role of camera intrinsic and extrinsic parameters in facial landmark localization and geometric alignment. Experimental results revealed that many existing neural network architectures neglect the impact of camera calibration during training, leading to significant errors in landmark prediction across multi-view input. This study underscores the importance of incorporating accurate camera modeling into learning-based synthesis pipelines.

The third and central component of this research introduces a multi-view neural regression framework for generating retopologized facial meshes directly from synthetic imagery. A three-stage neural architecture was implemented and evaluated, demonstrating the capability to produce topologically consistent, animation-ready meshes. These results support the feasibility of replacing traditional mesh creation pipelines with real-time inference-driven solutions, particularly when supported by synthetic training datasets.

In conclusion, this dissertation bridges the divide between traditional, handcrafted modeling workflows and modern, automated approaches powered by deep learning. The proposed methods improve production efficiency while preserving geometric fidelity and photorealism. The tools, methodologies, and findings presented here offer practical applications in digital human creation, game development, and visual effects, while also opening new directions for future research in face modeling and real-time avatar systems.

\section{Suggested Directions of Research}

While this dissertation has demonstrated the conceptual viability of using synthetic datasets derived from a 3D Morphable Model (3DMM) to train neural networks for facial mesh synthesis, several promising directions remain for extending this research.

First, a key next step involves the integration of the Variable Illumination Sphere (VarIS) with the Visage Craft system. Since VarIS provides precise knowledge of camera and lighting configurations, this data can be imported into Visage Craft to generate synthetic facial meshes that are fully aligned with the physical capture setup. Such a pipeline would enable controlled real-world image acquisition using VarIS and subsequent synthesis using models trained on its synthetic analogs. This would provide a highly effective closed-loop system, bridging the gap between synthetic training and real-world deployment.

Second, the current version of Visage Craft is limited to generating static face meshes. While this sufficed for the scope of this dissertation, more realistic training data that includes dynamic facial expressions is essential for advancing toward fully animatable avatars. In this regard, Unreal Engine’s Metahuman framework presents a compelling opportunity. It offers the ability to generate high-fidelity facial renderings with a wide range of expressions and photorealistic textures. Future research could leverage this capability to generate multi-view training datasets featuring dynamic facial expressions, enhancing the expressiveness and generalization of trained neural networks.

These future directions aim to further narrow the divide between simulated data and real-world applicability, pushing forward the development of robust, production-ready facial synthesis systems that are both efficient and expressive.